\documentclass[letterpaper, 10 pt]{IEEEtran}
\IEEEoverridecommandlockouts

\usepackage{url}
\usepackage[x11names]{xcolor}
\usepackage[colorlinks, bookmarksopen, bookmarksnumbered, citecolor=Red4, urlcolor=Turquoise4, linkcolor=Green4]{hyperref}

\usepackage{natbib}
\usepackage{amsmath}
\usepackage{bm}
\usepackage{amssymb}
\usepackage{graphicx}
\usepackage{multirow}
\usepackage{makecell}
\usepackage{siunitx}
\usepackage{tikz}
\usetikzlibrary{positioning}
\usepackage[caption=false]{subfig}
\usepackage{booktabs}
\usepackage{siunitx}
\usepackage{balance}

\DeclareMathOperator{\sign}{sign}

\title{\Large \bf A real-time RGB-D perception pipeline for autonomous impact hammers in mining: self-filtering, rock segmentation and rock-breaking poses generation}

\author{Martín Gallegos$^{1}$, Francisco Leiva$^{1}$, Patricio Loncomilla$^{1}$, Michelle Cortés$^{1}$ and Javier Ruiz-del-Solar$^{1}$%
\thanks{This work was supported by FONDECYT project 1251823, and ANID-PIA project CIA250010.}%
\thanks{$^1$Advanced Mining Technology Center (AMTC) and Department of Electrical Engineering, Universidad de Chile, Tupper 2007, Santiago, Chile.}%
\thanks{\tt\small{martin.gallegos@amtc.uchile.cl, francisco.leiva@ing.uchile.cl, ploncomi@ing.uchile.cl, \\michelle.valenzuela@amtc.uchile.cl, jruizd@ing.uchile.cl}}%
}

\begin{document}

\maketitle
\begin{abstract}

Impact hammers, also known as rock-breakers, are essential machines in mining operations, where they perform secondary reduction. In underground mining, these machines are typically teleoperated, limiting operational efficiency. This paper presents a real-time RGB-D perception pipeline as a step towards automating the operation of hydraulic impact hammers used in mining. The proposed system simultaneously generates operationally feasible rock-breaking poses and a robot-free 3D representation of the workspace. The proposed approach combines image-based instance segmentation with geometric point cloud processing, and operates on embedded hardware at approximately 10~Hz with a total latency of around 675~ms, enabling responsive closed-loop behavior when integrated with a control system. Experimental results in a representative scaled scenario demonstrate that the proposed system is suitable for real-time autonomous impact hammer operation.

\end{abstract}

\begin{IEEEkeywords}
    Automation in mining, hydraulic impact hammers, real-time perception.
\end{IEEEkeywords}

\section{Introduction}
\label{sec:introduction}

In underground mining operations, Load-Haul-Dump (LHD) machines move material from drawpoints, or blasting zones, to vertical  tunnels named ore passes. The transported material may contain boulders, whose size must not be excessively large for it to be processed in following productive stages. On top of ore passes there are steel grates (also known as grizzlies) that only allow sufficiently fragmented material to pass through; therefore, as large boulders accumulate on top of steel grates, they need to be further fragmented. This task, known as secondary reduction, is performed by impact hammers.

Impact hammers (also known as rock-breakers) are heavy-duty machines that typically consist of a hydraulic arm with an impact hammer attached as end-effector. These machines are widely used in mining operations for rock breaking and material fragmentation. Currently, they are typically remote-controlled by human operators working in safe control rooms who rely mainly on visual feedback provided by CCTV cameras~\citep{correa2022haptic} to conduct the teleoperation. In large mines, many ore passes may be active during production, so a given operator may be in charge of several impact hammers at once (each near a different ore pass). While the above prevents the exposure of operators to in-site hazards, it also limits operational efficiency due to, e.g., poor visibility of the environment and operators relying on control interfaces at the joint-level. More importantly, since each operator is typically responsible of several machines at the same time, operator unavailability introduces bottlenecks in the production pipeline, a problem that automation aims to alleviate.

\begin{figure}
    \centering
    \includegraphics[width=\linewidth, trim=0 40 0 0, clip]{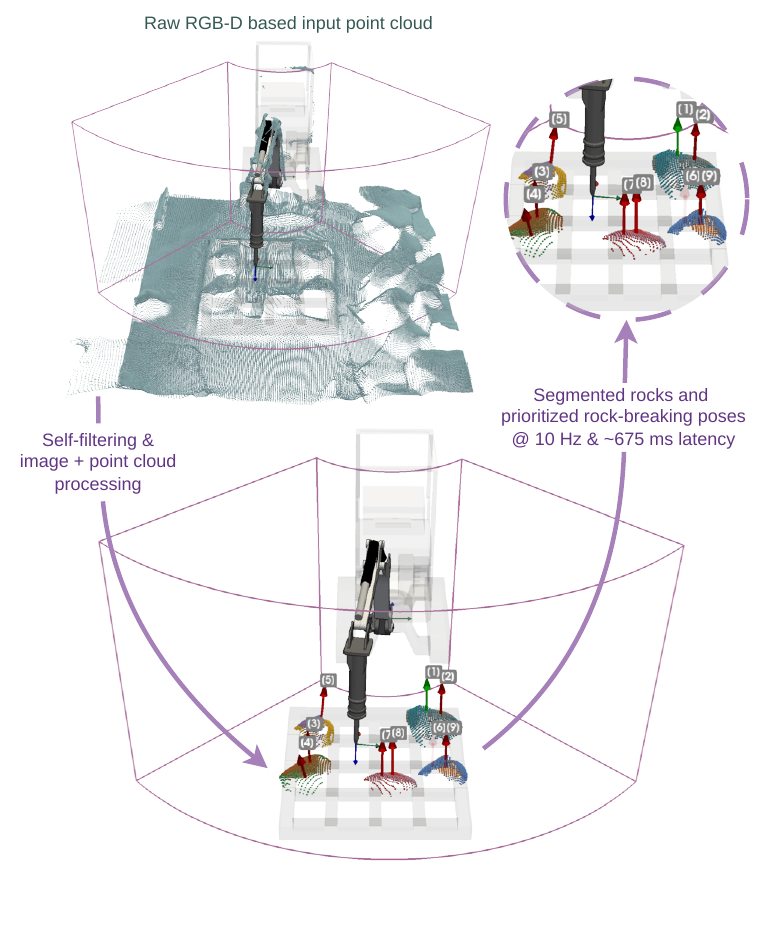}
    \caption{The perception pipeline uses RGB-D data and the impact hammer's kinematic model as inputs, and outputs segmented rocks that should be subjected to secondary reduction, alongside prioritized rock-breaking target poses to carry out that task.}
    \label{fig:hero}
\end{figure}

In this context, this paper presents a real-time RGB-D perception pipeline with the outlook of enabling the autonomous operation of impact hammers. The proposed pipeline characterizes relevant elements of the working environment of the impact hammer from sensor data (e.g., rocks above the grizzly), and generates target rock-breaking poses accordingly (see Fig.~\ref{fig:hero}). The proposed perception pipeline is equipped with a method for filtering the robot's geometry, which can be applied to either point clouds or depth maps; furthermore, occlusions due to the impact hammer moving over the grizzly are managed by means of a background model over depth maps (based on~\citet{yang2004real,yang2005real}), which is updated in non-occluded areas. Rocks over the grizzly are segmented via geometric clustering over filtered point clouds and by using instance segmentation over RGB images. Lastly, a geometric analysis on the segmented rocks is performed to generate and prioritize rock-breaking target poses given various criteria, e.g., the estimated size of the rock associated to a given pose, and the pose's distance to the impact hammer's end-effector. 

The proposed system is evaluated in a scaled environment that emulates the operating conditions of impact hammers used in mining. It runs at approximately $10$~Hz on an NVIDIA Jetson AGX Orin platform, supporting reactive closed-loop behaviors when paired with a control system for the impact hammer.

The main contributions of this work are the following:

\begin{itemize}

\item A computationally efficient RGB-D perception pipeline that enables the simultaneous real-time generation of rock-breaking target poses and a characterization of the impact hammer's workspace at approximately $10$~Hz on embedded hardware, making it suitable for integration into a full automation stack.

\item A surface-normal-based rock-breaking pose generation method that generates operationally feasible target poses to attempt rock fragmentation by explicitly integrating the local rock geometry along with the kinematic and operational constraints of hydraulic impact hammers.

\item A comprehensive experimental evaluation of the system's performance through quantitative and qualitative assessments, including a closed-loop validation with an impact hammer control system in the real world, and a computational profiling.

\end{itemize}

\section{Related work}

The secondary reduction task performed by hydraulic impact hammers is crucial in mining operations. As wheel loaders and LHDs deposit material on the steel grates on top of ore passes, large rocks and boulders may accumulate on them and potentially block a given production line. To avoid this issue, the static hydraulic impact hammers placed besides the ore passes have the task of breaking the rocks that cannot pass through the steel grate. The current manner in which these machines are remotely operated, however, may still induce bottlenecks in the production chain. Hence, several prior works have attempted to address the various challenges that arise from their automation, with the aim of improving efficiency in mining.

Documented efforts to automate this process date back roughly two decades~\citep{takahashi1998automatic,takahashi1999automatic,corke2005robotics}. While some recent works solely focus on the problem of controlling the impact hammers (e.g.,~\citet{samtani2023learning,leiva2026data}), our emphasis is on those that address the perception challenges associated to characterizing the environment in which these machines operate.

\citet{niu2018clustering} proposed a perception system that individualizes the rocks on top of the steel grate by applying clustering over point clouds obtained by a ToF camera. In~\citet{niu2019efficient}, the same research group later proposed a perception pipeline that relies on \mbox{RGB-D} data produced by a stereo camera to carry out this task; the system processes RGB images via YOLOv3~\citep{redmon2018yolov3} to detect rocks over the steel grate, and projects the resulting 2D bounding boxes to 3D space to get the point cloud associated to each detected rock. The positions of target rock-breaking poses are set given the centroid of each rock's bounding box, and their orientation is set by means of a surface-normal analysis around the centroid.

\citet{lampinen2021autonomous} proposed a full automation stack for impact hammers used in mining, encompassing perception and control. The perception pipeline is quite similar to that proposed by~\citet{niu2019efficient}, however, the target rock-breaking positions are generated by selecting the highest point in the neighborhood of the centroid of each rock's bounding box, and their orientation is set to be orthogonal to the grizzly. The authors also document a method for surface-normal analysis and more nuanced ways in which target poses can be generated, given that, empirically, they found that their strategy had flaws that resulted in unsuccessful rock breaking attempts when pairing the perception pipeline with their control system.

Similarly to~\citet{niu2018clustering}, \citet{lampinen2021robust} proposed a perception pipeline for the individualization of rocks over steel grates based on clustering analysis over point clouds. In this system, target rock-breaking poses are generated by analyzing the surface properties of each individualized rock; instead of selecting the rock's centroid as a candidate position to attempt rock breaking, the authors proposed searching for the largest cluster of points that corresponds to a flat surface (with respect to the steel grate), selecting the centroid of this cluster as the position of the target rock-breaking pose, and setting its orientation to be orthogonal to the steel grate. The authors also assign a score to each generated pose that approximately quantifies (via geometric features) how difficult a rock-breaking attempt on it would be.

\citet{cardenas2022automatic} proposed a perception system that individualizes the rocks over the steel grate and then generates target rock-breaking poses. This system characterizes individual rocks by performing a segmentation over a point cloud of the environment. This segmentation is performed via a watershed-like algorithm that can take image-based rock detections (using the RockyCenterNet model~\citep{loncomilla2022detecting}) as priors. The authors perform a heuristics-based geometric analysis on the segmented rocks' point clouds to generate target rock-breaking poses, and also prioritize them according to criteria based on operation manuals and feedback provided by human operators that control impact hammers in underground mining.

The review of prior literature reveals several unaddressed challenges that motivate this work. First, most previous approaches do not account for the dynamic nature of impact hammer operation in mining. Although the working environment is largely quasi-static, the impact hammer must continuously move from rock to rock to fragment them, which introduces challenges that go beyond characterizing how rocks are distributed over the steel grate. To account for this, we introduce an efficient self-filter that operates on either depth maps or point clouds while also accounting for occlusions produced by the impact hammer through a depth-based background model. This enables the generation of a 3D representation of the workspace that can subsequently be used, for instance, by collision avoidance modules. Furthermore, the proposed perception pipeline is explicitly designed for its integration into a full automation stack; it runs on embedded hardware in real time, and takes into account the need for temporal consistency in the generation of target rock-breaking poses.

A second limitation of some of the previous approaches is that they do not jointly consider the kinematic feasibility and operational constraints associated with hydraulic impact hammers when producing target rock-breaking poses. Instead, target poses are generated using simple geometric criteria that do not always account for either the local characteristics of the rock surface or the physical limitations of the machine. As a result, these methods may generate poses that lead to end-effector slippage during operation (which subsequently may produce blank firing), or that simply cannot be reached by the impact hammer's end-effector. Such situations are operationally undesirable, as they increase the risk of damaging either the impact hammer or the surrounding infrastructure. The proposed method explicitly incorporates these constraints into the pose generation process (see Sect.~\ref{subsubsec:breaking_poses_generator} and App.~\ref{app:breaking_poses}), ensuring that only feasible and operationally compliant rock-breaking poses are generated.

\section{Problem description}

Although the automation of impact hammers broadly entails control and perception, this work focuses on the latter while taking into consideration technical aspects that would allow pairing the perception pipeline with a control system. In this context, the problem consist of characterizing the rocks that remain on the grizzly (e.g., due to their size), and generating target rock-breaking poses so that the impact hammer can attempt to break them. 

\begin{figure}
    \centering
    \includegraphics[width=\linewidth]{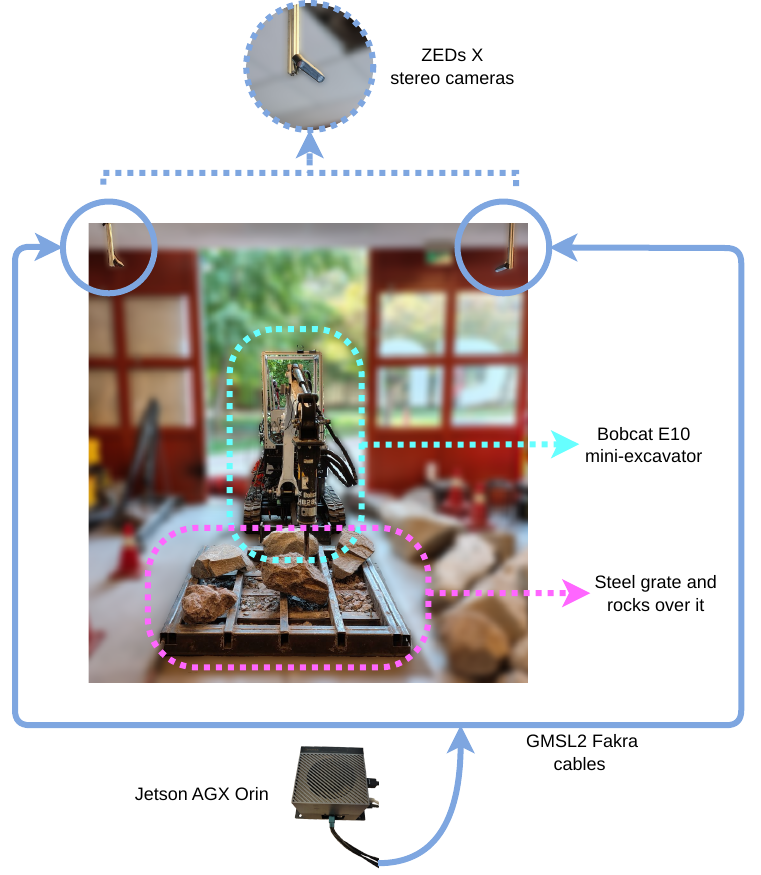}
    \caption{Real-world experimental setup.}
    \label{fig:experimental_setup}
\end{figure}

The experimental setup consists of a scaled environment designed to mimic underground mining ore passes. It comprises a Bobcat E10 mini-excavator equipped with a hydraulic impact hammer as its end-effector, and a $1.5\times1.5$~m steel grate. The mini-excavator is mechanically fixed to the floor adjacent to the grate (see Fig.~\ref{fig:experimental_setup}). Together, they form a laboratory-scale experimental setting that is representative of some of the real setups in which impact hammers operate in mining.

The Bobcat E10 mini-excavator is electro-hydraulically intervened so their actuators are controllable via electrical signals (currents), and the configuration of its hydraulic arm (four degrees of freedom) is measured via rotational encoders collocated to each joint. A kinematic model of this machine is also available via Unified Robot Description Format (URDF) files, although this model does not account for the arm's hydraulic cylinders due the inability of standard URDF to encode kinematic loops. Note that the system proposed in this work requires a kinematic model of the machine to support several stages of the perception pipeline.

To characterize the environment, two Stereolabs ZED~X\footnote{\url{https://www.stereolabs.com/en-cl/products/zed-x}} cameras are utilized. In our experimental setup, these stereo cameras are mounted on the roof over the steel grate and the mini-excavator to provide a direct view of the overall scene from two different point of views. Both cameras are connected to an NVIDIA Jetson AGX Orin\footnote{\url{https://www.nvidia.com/es-la/autonomous-machines/embedded-systems/jetson-orin/}} via GMSL2 Fakra cables. Fig.~\ref{fig:experimental_setup} illustrates the described experimental setup.

\begin{figure*}
    \centering
    \includegraphics[width=\linewidth]{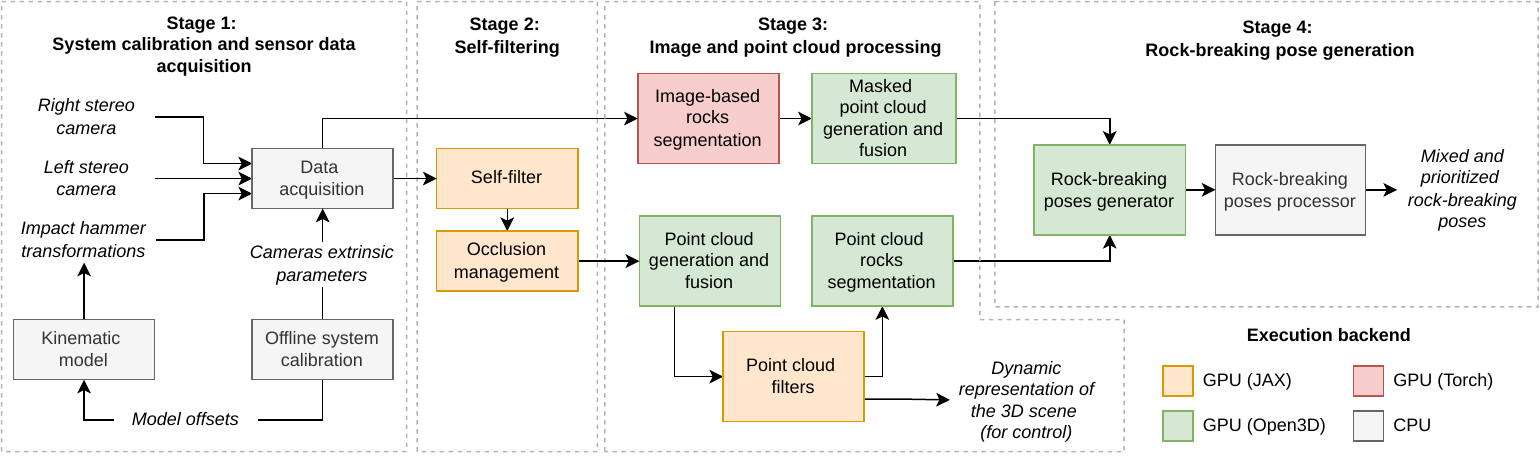}
    \caption{Complete perception pipeline diagram, highlighting the primary execution backend utilized for each sub-module.}
    \label{fig:streamlined_perception_pipeline}
\end{figure*}

Formally, the addressed problem can be described as follows: at every discrete time step $t$, the perception pipeline receives a pair of RGB-D frames from two stereo cameras, the intrinsic and extrinsic parameters of said cameras, and a characterization of the current impact hammer's configuration. With this data, the system has to simultaneously generate (i) a robot-free three-dimensional representation of the workspace via a point cloud and (ii) a set of kinematically (and operationally) feasible rock-breaking target poses for the impact hammer's end-effector, $\mathcal{R}=\{(\bm{p}_i, \bm{R}_i)\}_{i=1...N}$, where each pose $(\bm{p}, \bm{R})$ is characterized by a 3D position, $\bm{p}\in\mathbb{R}^3$ and an orientation $\bm{R}\in \text{SO}(3)$. These target poses are deemed kinematically feasible if they are within a constrained region of the impact hammers' workspace, and are reachable by its end-effector. Moreover, the poses have to be prioritized given several criteria based on operational constraints. The generated robot-free, 3D workspace representation of the environment, is intended to be consumed by downstream control modules, such as collision avoidance or motion planning sub-systems, so it should account for the robot's geometry and the occlusions that its movement induces in the scene. The rock-breaking poses, on the other hand, provide the desired impact targets for the hammer. Finally, both outputs must be generated at a frequency compatible with responsive closed-loop operation (i.e., that would allow pairing the perception pipeline with a control system), which we target at $10$~Hz or above.

\section{Proposed perception pipeline}

The proposed perception pipeline encompasses several steps, from data acquisition, up until the segmentation of rocks, the generation of target poses for the impact hammer, and their subsequent prioritization. Fig.~\ref{fig:streamlined_perception_pipeline} shows the sequence of operations performed by the proposed pipeline. Note that these operations are grouped into four main stages: system calibration and sensor data acquisition, self-filtering, image and point cloud processing, and rock-breaking pose generation. In what follows, each of these stages and their sub-modules are described.

\subsection{Stage 1: System calibration and sensor data acquisition}
\label{subsec:system_data_acquisition}

This stage involves calibrating the perception system and initializing and setting up the exteroceptive sensors so that the \mbox{RGB-D} data they produce can be correctly fused and interfaced with the rest of the pipeline.

Since the extrinsic parameters of the stereo cameras are not readily available (whereas their intrinsic parameters are provided by the vendor), and may change between experimental setups, an offline calibration procedure is required to estimate them. Moreover, a procedure to get the kinematic model's joint offsets (assuming calibrated encoders) is also carried out, so as to match the projection of each link's mesh with what is sensed by the cameras.

After calibration, the impact hammer's kinematic model and sensor static transformations are set as inputs to the perception system. Once this data is available, the stereo cameras are initialized, configured and queried through the SDK provided by their vendor.\footnote{\url{https://github.com/stereolabs/zed-python-api}} 

In what follows, both the system's offline calibration and the data acquisition procedures are described in further detail.

\subsubsection{Offline system calibration}
\label{subsec:calibration}

To fuse the data provided by the stereo cameras, their extrinsic parameters need to be estimated. This is achieved by registering the point clouds produced by the cameras (from the depth maps they provide) using the Iterative Closest Point (ICP) algorithm~\citep{besl1992method}. In addition, since the impact hammer's kinematic model is used for self-filtering (see Sect.~\ref{subsubsec:self_filter}), the projection of its geometry into the images needs to match the real robot configuration. This is achieved by performing a reprojection error minimization procedure. Both calibration steps are described in what follows.

\paragraph{ICP calibration}
\label{subsubsec:icp_calibration}

To perform this calibration procedure, one of the stereo cameras is selected as fixed, and the poses of the other sensors are estimated relative to it via point-to-point ICP. This requires obtaining the point clouds associated to the depth maps provided by each stereo camera, which is done via the per-pixel projection 
\begin{equation}
    \bm{p} = d\bm{K}^{-1}\tilde{\bm{u}}
    \label{eq:projected_2d_points}
\end{equation}
where $\bm{K}\in\mathbb{R}^{3\times 3}$ is the camera's intrinsic matrix, $\tilde{\bm{u}}\in{\mathbb{R}^3}$ is a depth map (homogeneous) coordinate, $d\in\mathbb{R}$ is the depth value for that coordinate, and $\bm{p}\in\mathbb{R}^3$ is the resulting 3D point.

Given the resulting point clouds, ICP estimates a static transformation between the stereo cameras that produced them, which enables fusing their data into a unified point cloud, which is later used by other modules of the perception pipeline. 

Also as part of this calibration step, an approximate pose of the sensor set as fixed for ICP with respect to the center of the grizzly (or other reference) must be provided; we will denote this pose as $\bm{T}_\text{aux}\in \text{SE}(3)$. This will be required for the reprojection error minimization procedure described in Sect.~\ref{subsubsec:image_robot_registration}, and can be set by inspection.

Given the above, the pose of each stereo camera can be written as $[\bm{R}|\bm{t}]\in\mathbb{R}^{3\times 4}$, where $\bm{R}\in \text{SO}(3)$ and $\bm{t}\in \mathbb{R}^3$ represent their rotation matrix and translation expressed in terms of the introduced auxiliary frame. Recalling that the intrinsic matrix of a given sensor is denoted by $\bm{K}\in \mathbb{R}^{3\times 3}$, their camera matrix will be given by $\bm{C}=\bm{K}\cdot[\bm{R}^{\top}|-\bm{R}^{\top}\bm{t}]\in\mathbb{R}^{3\times 4}$.

Since our experimental instantiation of the perception pipeline has two stereo cameras placed besides the mini-excavator (see Fig.~\ref{fig:experimental_setup}), the above produces two camera matrices, which will be denoted as $\bm{C}_\text{left}$ and $\bm{C}_\text{right}$ for the camera placed at the left and right side of the machine, respectively. 

\paragraph{Reprojection error minimization}
\label{subsubsec:image_robot_registration}

This subsystem estimates the position $\bm{p}_\text{aux}$ of the auxiliary frame introduced in the ICP calibration stage, $\bm{T}_{\text{aux}}$, with respect to the impact hammer's base link frame, and estimates the joints' offsets for the impact hammer's kinematic model, so that its projection into the images provided by the stereo cameras is accurate. 

\begin{figure}
    \centering
    \includegraphics[width=\linewidth, trim=0 30 0 0, clip]{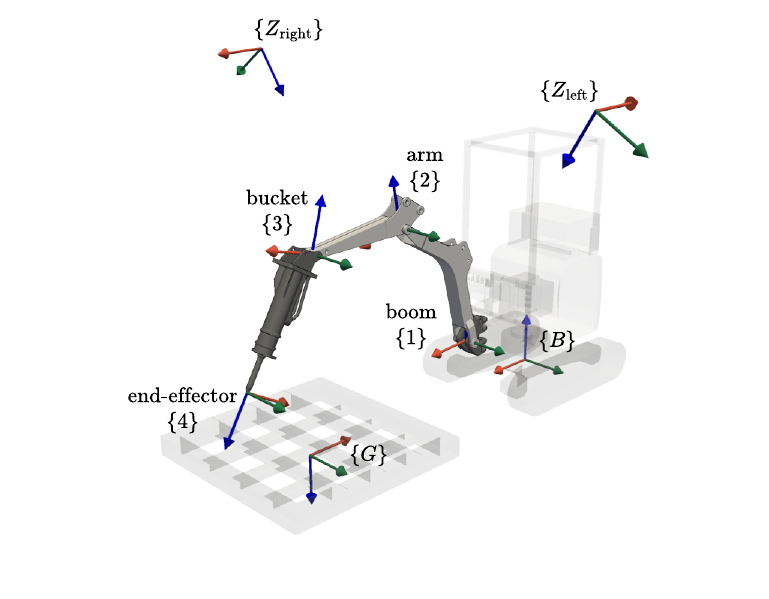}
    \caption{Calibration layout of the Bobcat E10 mini-excavator. The fixed frames $\{B\}$ and $\{G\}$ are attached to the mini-excavator base link and to the center of the grizzly, respectively. The fixed frames $\{1\}$, $\{2\}$, $\{3\}$, and $\{4\}$ are attached to each link of the hammer arm. The fixed frames $\{Z_\text{left}\}$ and $\{Z_\text{right}\}$ correspond to the left and right cameras respectively.}
    \label{fig:calibration_frames}
\end{figure}

The kinematic model of the impact hammer allows mapping instantaneous configurations $\bm{q}_t\in\mathbb{R}^4$ (i.e., the four angular positions that define the spatial pose of the hydraulic arm for a given time step) into poses for its end-effector and the fixed frames attached to every link, which are aligned with each joint axis. Formally, these poses are computed via forward kinematics and (by default) are represented with respect to the base link frame of the impact hammer. As a reference, Fig.~\ref{fig:calibration_frames} shows the fixed frames associated to three out of the four joints of the Bobcat~E10's arm (the \emph{``swing''} joint is omitted), the fixed frame for its end-effector and base link, and those associated to the stereo cameras, and the steel grate.

If we assume that the encoders measuring the configuration of the impact hammer are calibrated, a given set of angular positions may still not produce good estimates for the poses of each of its links' fixed frames if the kinematic model zero-pose does not accurately match the zero-pose of the real machine. This can happen due to modeling errors, specially because kinematic models for impact hammers may not be readily provided by vendors and also due to slight physical modifications that these machines may undergo during maintenance sessions. To account for this potential mismatch, we can estimate a set of angular offsets $\Delta\bm{q}\in\mathbb{R}^4$ so that the corrected configuration fed to the kinematic model is $\bm{q}_t+\Delta\bm{q}$.

Setting a fixed rotation matrix for the auxiliary frame $(\bm{p}_\text{aux}, \bm{R}_\text{aux})=\bm{T}_\text{aux}$ (assuming that the rotation of the fixed sensor with respect to the auxiliary frame is accurate, and considering the difficulty of minimizing rotational error from 2D projections), we want to estimate its position $\bm{p}_\text{aux}$ and the joint's angular offsets $\Delta \bm{q}$. We aim at accomplishing this by means of a reprojection error minimization procedure between 2D projections (in image coordinates) of the positions of certain fixed frames as predicted by the machine's kinematic model, and manually labeled keypoints (also in image coordinates) associated to the same fixed frames.

Given our experimental setup, we select per-link fixed frames that are aligned with each joint's rotational axis, but discard the first joint (which controls the hydraulic arm's \emph{``swing''} motion) given that it is not easy to label keypoints on it in the images provided by the stereo cameras. We also use a fixed frame with origin in the tip of the impact hammer's chisel, as can be seen in Fig.~\ref{fig:calibration_frames}.

Denoting the mapping that computes the poses of each \mbox{$j$-th} fixed frame with respect to the machine's base link as $\text{FK}(\cdot, j)$, we can get their poses with respect to $\bm{T}_\text{aux}$ via 
\begin{equation}
   \bm{T}_j= (\bm{p}_j, \bm{R}_j) = \bm{T}_{\text{aux}}^{-1}  \cdot \text{FK}(\bm{q}+\Delta\bm{q} , j).
    \label{eq:fk_computation}
\end{equation}

Since all joints have physical widths, but the fixed frames associated to their corresponding links are centered in their rotational axes, we must take into account whether their image projection will be located on the right side or on the left side of the hammer. This can be easily done by considering two offsets on the fixed frame axes that coincides with the axis of rotation of each joint.

By offsetting these positions along the rotational axis of each joint, we can get positions that would be on the right or left surface of the hydraulic arm:
\begin{align}
     \bm{p}^{\text{right}}_j &= \bm{p}_j+ \bm{R}_j \cdot \bm{p}^{+\text{right}}_{j},\\
     \bm{p}^{\text{left}}_j &= \bm{p}_j + \bm{R}_j \cdot \bm{p}^{+\text{left}}_j.
     \label{eq:forward_kinematics}
\end{align}
Note that for the fixed frame attached to the end-effector tip, both offsets are set to zero given the geometry of the chisel.

These 3D points can then be expressed in homogeneous coordinates and projected to their corresponding left or right camera's image coordinates via
\begin{equation}
    \hat{\bm{u}}^{\text{right}}_j = \mathrm{proj} \left(\bm{C}_{\text{right}}\cdot \tilde{\bm{p}}_j^{\text{right}}\right) \quad \text{or} \quad
    \hat{\bm{u}}^{\text{left}}_j = \mathrm{proj} \left(\bm{C}_{\text{left}}\cdot \tilde{\bm{p}}_j^{\text{left}}\right),
    \label{eq:pinhole_model}
\end{equation}
such that 
\begin{equation}
   \mathrm{proj}(\bm{p})=(p_x/p_z, p_y/p_z)^{\top}. 
   \label{eq:2d_proj}
\end{equation}

To get labeled, ground-truth data to compare the projected points described above, we capture and store synchronized RGB images $\bm{I}_t$, and impact hammer's configurations, $\bm{q}_t\in\mathbb{R}^4$, where the former are provided by the stereo cameras, and the latter are measured by per-joint encoders. Then, using a simple GUI, the positions of the visible joint axes are clicked on the stored images.

For the two stereo cameras utilized in our experimental setup, the above procedure allows the generation of a dataset $\mathcal{D} = \{\bm{u}^{c, (k)}_j, \bm{q}_t^{(k)}\}_{k=1...N}$, where $\bm{u}^c_j$ is the position of the pixel that was clicked on the image (the keypoint), $c \in \{\texttt{left},\texttt{right}\}$ indexes the camera in which the pixel was clicked, $j$ indexes the fixed frame that was clicked, and $\bm{q}_t$ is the arm configuration that was measured at that time. 

With the above, the calibration of the robot is performed in two steps. In the first step, the position of its auxiliary frame, $\bm{p}_\text{aux}$ is estimated by only considering the fixed frame associated to the \emph{``boom''} joint (i.e., $j=1$). The position of this fixed frame can be affected by a poorly calibrated offset for the swing joint, however, given its physical closeness to said joint's axis, this potential issue has minimal impact in practice. 

The estimation of $\bm{p}_\text{aux}$ is performed by the minimization of \eqref{eq:loss_base} via stochastic gradient descent, where $\mathcal{D}_1\subset\mathcal{D}$ only contains samples for $j=1$ (as previously stated), and we have highlighted, due to Eq.~\eqref{eq:fk_computation}, the dependence of the reprojections $\hat{\bm{u}}$ on $\bm{q}$ and $\bm{p}_\text{aux}$ (since the rotation of the auxiliary frame is fixed), and that this loss is computed by setting $\Delta\bm{q}=\bm{0}$ (i.e., without corrections over potential model errors).
\begin{equation}
\mathcal{L}_{{\text{aux}}}(\bm{p}_{\text{aux}}) = \sum_{(\bm{u},\bm{q}) \in \mathcal{D}_1}{\Vert\hat{\bm{u}}(\bm{q}, \bm{p}_{\text{aux}})|_{\Delta\bm{q}=\bm{0}} - \bm{u}\rVert_2^2 }
\label{eq:loss_base}
\end{equation}

In the second calibration step, the kinematic model offsets $\Delta \bm{q}$ are estimated. Analogous to the estimation of $\bm{p}_{\text{aux}}$, this is done by minimizing~\eqref{eq:loss_offsets}, where $\mathcal{D}_{1...3}\subset\mathcal{D}$ contains samples for indexed fixed frames with $j\in\{1,2,3\}$, $\mathcal{D}_{4}$ only contains samples associated to the end-effector fixed frame, and we have highlighted the dependency of $\hat{\bm{u}}$ on $\bm{q}$ and $\Delta\bm{q}$. Moreover, note that this loss is computed by fixing the auxiliary frame position to that obtained in the previous calibration step, which we have denoted as $\bm{p}^*_\text{aux}$. 

To encourage a precise end-effector projection over the stereo camera images, note that~\eqref{eq:loss_offsets} weights all the other fixed frames with a scalar factor $\lambda=0.2$.

\begin{equation}
\begin{split}
    \mathcal{L}_\text{offsets}(\Delta \bm{q}) &= \lambda \sum_{(\bm{u}, \bm{q})\in\mathcal{D}_{1...3}}{\|\hat{\bm{u}}(\bm{q}, \Delta\bm{q})|_{\bm{p}_\text{aux}=\bm{p}^*_\text{aux}} - \bm{u}\|_2^2}\\ & \qquad + \sum_{(\bm{u},\bm{q})\in\mathcal{D}_4}\|\hat{\bm{u}}(\bm{q}, \Delta\bm{q})|_{\bm{p}_\text{aux}=\bm{p}^*_\text{aux}} - \bm{u}\|_2^2
\end{split}
\label{eq:loss_offsets}
\end{equation}

The resulting offsets $\Delta\bm{q}^*$ are utilized afterwards to modify the zero-joints pose of the mini-excavator's kinematic model. The optimization of both $\mathcal{L}_\text{aux}$ and $\mathcal{L}_\text{offsets}$ rely on the instantiation of the impact hammer's kinematic model in Brax~\citep{freeman2021brax}, which eases the computation of gradients.

\subsubsection{Data acquisition}
\label{subsec:data_acquisition}

We use the vendor API to query both stereo cameras for data simultaneously, which ensures that the RGB images and depth maps provided by the sensors are synchronized given their capture card. To interface this data with the rest of the pipeline, we use ROS~\citep{quigley2009ros}, and a custom ROS message which we call \texttt{Images}, that contains pairs of RGB images, depth maps, intrinsic parameters and a time stamp, and that is constructed and sent every time the cameras are queried for RGB-D data.

The initialization parameters of the stereo cameras, including resolution and depth mode (a machine learning-based post-processing stage provided by the vendor), can be found in App.~\ref{app:sensor_data_acquisition}. The transformations that account for the instantaneous configuration of the impact hammer, and those that are static and associated to the stereo cameras, are also streamed separately using ROS (see Fig.~\ref{fig:streamlined_perception_pipeline}).

\subsection{Stage 2: Self-filtering}
\label{subsec:self_filtering}

During operation, the hammer motion inherently causes occlusions in the camera views, leaving portions of the environment partially unobserved, regardless of whether the system is teleoperated or autonomous. As a consequence, the quality of the reconstructed 3D point cloud degrades, leading to incomplete regions and holes in the scene. These artifacts hinder reliable rock segmentation and may negatively affect the quality of the information that could be provided for other downstream tasks (e.g., control sub-systems).

To address these issues, the second stage of the pipeline focuses on removing the impact hammer from the scene representation, via self-filtering, and on handling the occlusions caused by it during operation. The approaches adopted to implement an ad-hoc self-filter and an occlusion management strategy are described in detail in the following sub-sections.

\subsubsection{Self-filter}
\label{subsubsec:self_filter}

To filter the robot's geometry, we adopt a strategy akin to the legacy ROS package \texttt{rgbd\_self\_filter}\footnote{\url{https://wiki.ros.org/rgbd_self_filter}}, which, in turn, aligns with the \texttt{camera\_self\_filter}\footnote{\url{https://wiki.ros.org/camera_self_filter}} implementation. Similarly to the underlying ideas of these methods, we perform self-filtering over depth maps by relying on the robot's kinematic model, but we perform approximations over the robot's geometry, therefore, we do not rely on rendering the (exact) URDF on the camera's viewpoint to perform the filtering. In addition, given our experimental setup, where cameras are on top of the hammer and the grizzly, we also assume that no relevant information is worth characterizing from the cameras up to the robot's geometry. Since this completely limits the perception of objects that would intersect the per-pixel camera rays passing through the robot's body, if the previous assumption does not hold for certain settings (e.g., different exteroceptive sensors placement), the optional point-cloud based self-filtering we describe in Sect.~\ref{subsubsubsec:3d_hammer_self_filter} should be used instead.

To address the impact hammer self-filtering problem, a three-stages method is proposed. The first stage consists of processing the robot's links as geometric bodies to compute the values of several variables of interest that will be required to filter the hammer in 2D depth maps. The second stage, similarly to the first one, focuses on modeling the hammer's hydraulic cylinders, which are not readily available as meshes in the machine's URDF due to the inability of this format to handle closed-loop kinematic chains (see Fig.~\ref{fig:calibration_frames} for reference). Finally, the third stage describes the filtering procedure itself, which, as previously stated, is carried out in 2D depth maps, after which the results are projected into 3D space.

As a preliminary step, a JAX-based implementation of the bilateral filter is applied to the depth maps provided by the stereo cameras to remove spurious depth values. Details regarding this filter are described in App.~\ref{app:bilateral_filter}. 

\paragraph{Robot meshes subdivision} 
\label{subsubsec:robot_mesh_subdivision}
This stage consists of subdividing each of the impact hammer's link meshes into ``slices''. This is done by first fitting an oriented bounding box to a given mesh, and slicing the mesh's vertices along the major axis of its bounding box. Then, for every slice, oriented bounding boxes and enclosing spheres are computed, and their geometric parameters (origin and extents for bounding boxes and origin and radii for spheres) are stored in a configuration file for latter reuse. It is possible to apply offsets to the extents of the bounding boxes and to the radii of the spheres in order to enlarge their dimensions and thereby extend the range of the self-filtering process.

\begin{figure}
    \centering
    \subfloat[]{\includegraphics[width=\linewidth, trim={0, 2.2cm, 0, 4.6cm},clip]{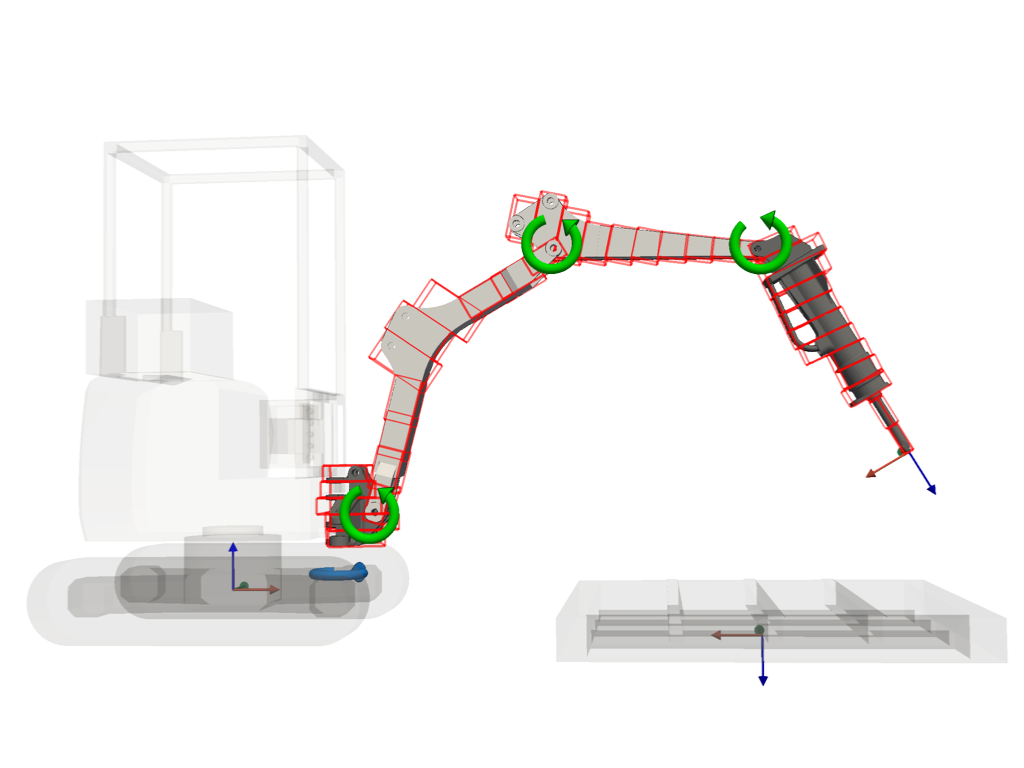}\label{fig:boxes_hammer_divisions}} \\
    \subfloat[]{\includegraphics[width=\linewidth, trim={0, 2.2cm, 0, 4.6cm},clip]{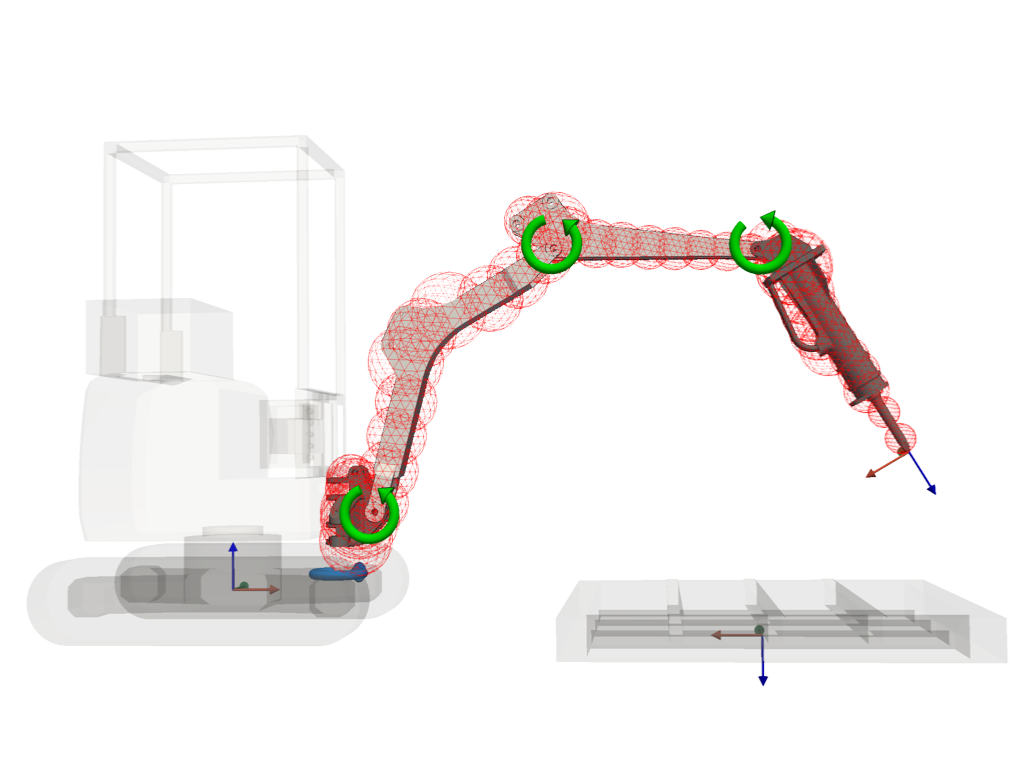}\label{fig:spheres_hammer_divisions}}
    \caption{Example of the subdivision procedure applied on the meshes of the Bobcat E10 mini-excavator. Here we only show the results associated to the slices on its hydraulic arm and end-effector. (a) Meshes subdivided into oriented bounding boxes. (b) Meshes subdivided into enclosing spheres.}
    \label{fig:hammer_divisions}
\end{figure}

From the origins and extents stored for the bounding boxes, we compute their vertices; analogously, from the origins and radii for the enclosing spheres, we get points on their surfaces following a prescribed routine that iterates over spherical coordinates. Finally, the transformations required to express these computed vertices with respect to their corresponding mesh's origin are also computed and stored. An example of the result obtained following the described procedure is illustrated in Fig.~\ref{fig:hammer_divisions}, where both the mesh slicing using oriented bounding boxes and enclosing spheres is shown.

All mesh processing and geometric computations are carried out using the PyVista framework \citep{sullivan2019pyvista}. Note that all the calculations required for performing the meshes subdivisions are performed only once and their results are saved in a configuration file. This allows initializing the self-filter without repeating all these pre-processing steps, and enables quickly switching between boxes-based and spheres-based representations for subsequent processing steps (see Sect.~\ref{subsubsec:filtering_process}). Moreover, several configuration files may be generated for different numbers of slices, allowing tailored parameter sets to be employed in different self-filter instantiations.

\paragraph{Modeling the impact hammer cylinders} 
\label{subsec:modeling_hammer_cylinders}
The hydraulic cylinder meshes of the impact hammer arm are not part of its kinematic model, as their inclusion would imply the existence of kinematic loops, which are not supported in vanilla URDF. For this reason, to consider them for self-filtering, the cylinders are abstracted by representing their geometry as oriented bounding boxes.

To do so, each cylinder is assigned a radius $r>0$ and two end points, $\bm{p}_{\text{local-start}}$ and $\bm{p}_{\text{local-end}} \in\mathbb{R}^3$ (referenced with respect to its corresponding link). At each time step, these points are transformed to the impact hammers' base link frame, which results in ${\bm{p}}_\text{start}$ and $\bm{p}_\text{end}$. These new points are then utilized to compute a pose with origin in the cylinder's midpoint, $\bm{p}_\text{cylinder}=(\bm{p}_\text{end} + \bm{p}_\text{start})/2$, and rotation matrix $\bm{R}_\text{cylinder}=\begin{bmatrix}
    \hat{\bm{x}}_r& \hat{\bm{y}}_r& \hat{\bm{z}}_r
\end{bmatrix}\in \text{SO}(3)$, such that 
\begin{equation}
    \hat{\bm{z}}_r=\frac{\bm{p}_\text{end}-\bm{p}_\text{start}}{\|\bm{p}_\text{end}-\bm{p}_\text{start}\|},
    \hat{\bm{x}}_r=\frac{\hat{\bm{z}}_r\times \hat{\bm{w}}_r}{\|\hat{\bm{z}}_r\times \hat{\bm{w}}_r\|}, \text{and}~\hat{\bm{y}}_r=\hat{\bm{z}}_r \times \hat{\bm{x}}_r,
\end{equation}

where $\bm{w}_r:=\hat{\bm{x}}$ if $|z_{r,z}|>0.9$ and $\bm{w}_r:=\hat{\bm{z}}$ otherwise.\footnote{$|z_{r,z}|>0.9$ is required (arbitrarily) to avoid $\hat{\bm{z}}_r\times \hat{\bm{w}}_r / \|\hat{\bm{z}}_r\times \hat{\bm{w}}_r\|$ to be numerically unstable.}

After performing these calculations, the vertices of the oriented bounding box associated to the cylinder, $\bm{v}_j\in\mathbb{R}^3$, $j \in \{1, ..., 8\}$, are computed according to
\begin{equation}
\bm{v}_j = \alpha^{(1)}_j {\bm{p}}_\text{start} + 
           \alpha^{(2)}_j {\bm{p}}_\text{end}   + 
          r\alpha^{(3)}_j  \hat{\bm{x}}_r       + 
          r\alpha^{(4)}_j  \hat{\bm{y}}_r, 
\label{eq:cylinders_vertices}
\end{equation}
where $\alpha^{(1)}_j = \left\lfloor \frac{j-1}{4} \right\rfloor$, $\alpha^{(2)}_j =   1 - \left\lfloor\frac{j-1}{4} \right\rfloor$, $\alpha^{(3)}_j = (-1)^{\left\lfloor \frac{j-1}{2} \right\rfloor}$, and $\alpha^{(4)}_j = (-1)^{j-1}$.

\begin{figure}
    \centering
    \includegraphics[width=\linewidth, trim={0, 2.2cm, 0, 2cm},clip]{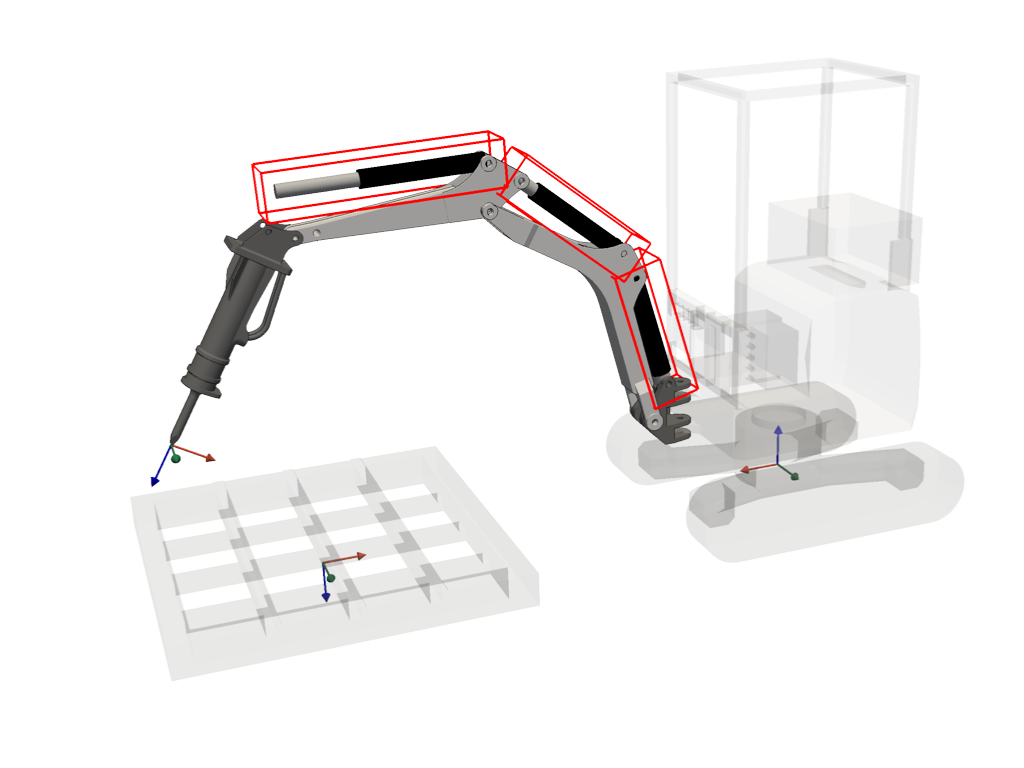}\label{fig:hammer_cylinders_raw}
    \caption{Side view of the Bobcat E10 mini-excavator, highlighting simplified markers for the hydraulic cylinders of its arm, and the computed oriented bounding boxes for each of them. Note that here an offset of $0.12$~m is applied over the nominal radii of each cylinder (see Eq.~\eqref{eq:cylinders_vertices}), which allows partially controlling the volumes of the bounding boxes outlined in red.}
    \label{fig:hammer_cylinders}
\end{figure}

The result of applying this procedure to characterize the Bobcat E10 hydraulic arm's cylinders is shown in Fig.~\ref{fig:hammer_cylinders}, where both simplified cylinder markers and their corresponding oriented bounding boxes are illustrated. Note that the hydraulic cylinder associated to the \emph{``swing''} joint is not modeled via bounding boxes, since, given its placement, does not occlude the scene.

\paragraph{Filtering process} \label{subsubsec:filtering_process} This stage consists of filtering the impact hammer from the depth maps provided by the stereo cameras by leveraging the previously computed oriented bounding boxes or enclosing spheres.

To do so, at each time step, all the vertices obtained in the first and second stages (via bounding boxes or enclosing spheres), are projected onto both the left and right depth images (since the cylinder’s vertices are already expressed in the hammer’s base-link coordinate frame, the link vertices must likewise be transformed into this frame so that all vertices are represented within the same reference system). Given that these vertices are associated to a given impact hammer link or hydraulic cylinder, we denote them as $\bm{v}_{i,j}\in\mathbb{R}^3$, where $i\in\{1,..., k_j\}$ indexes slices, and $j$ indexes the links and cylinders. Then, as in Eq.~\eqref{eq:pinhole_model}, these vertices are represented in homogeneous coordinates, and then projected onto the depth maps using
\begin{equation}
    \hat{\bm{u}}_{i,j}^c=\mathrm{proj}\left(\bm{C}_c\cdot\tilde{\bm{v}}_{i,j}\right),
\end{equation}

where $\mathrm{proj}(\cdot)$ is given by \eqref{eq:2d_proj}, $c\in\{\texttt{left}, \texttt{right}\}$ indexes the cameras, $\bm{C}_c$ is a camera matrix, and $\hat{\bm{u}}_{i,j}^c\in\mathbb{R}^2$ is the resulting projection. 

Subsequently, the 2D convex hulls of the projections associated to each indexed links and cylinders are computed.
This is done independently for both depth maps (left and right), and the resulting polygons are grouped for each depth map so as to get two (binary) masks in image coordinates. The resulting masks are then applied over their depth map, where corresponding pixels are assigned \texttt{NaN} values. 

\begin{figure}
    \centering
    \subfloat[]{\includegraphics[width=0.495\linewidth]{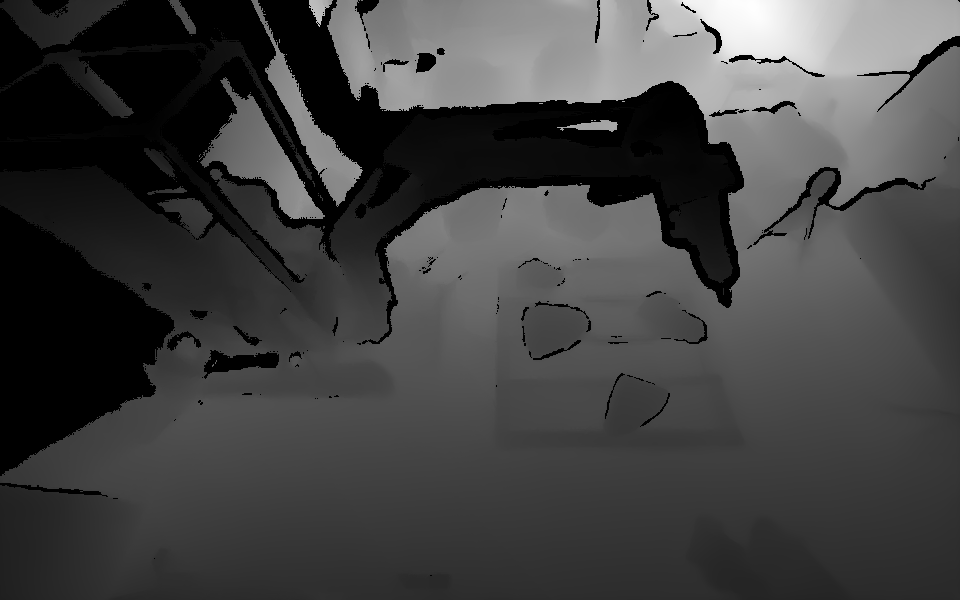}\label{fig:projected_self_filter_a}} \hfill
    \subfloat[]{\includegraphics[width=0.495\linewidth]{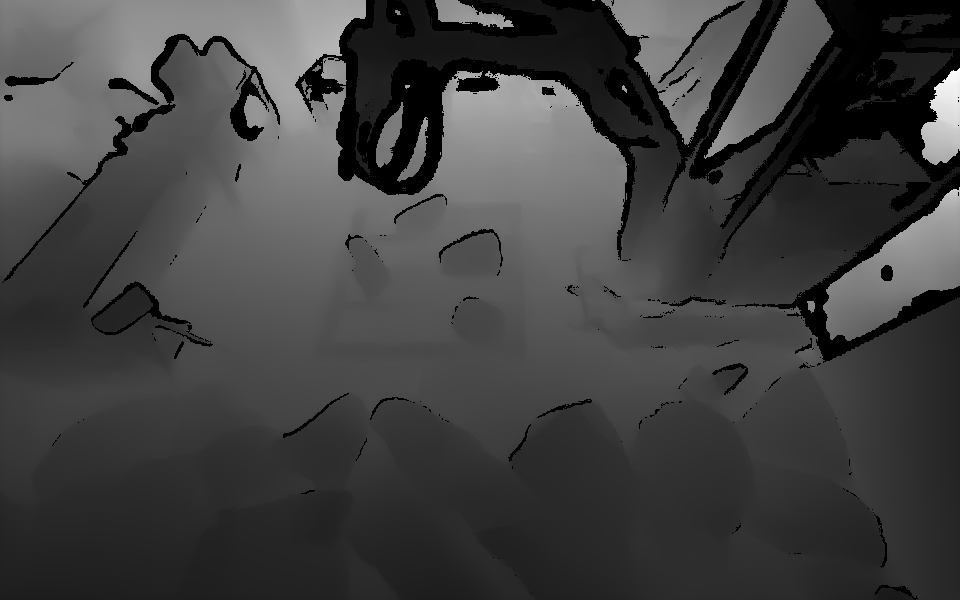}\label{fig:projected_self_filter_b}} \\
    \subfloat[]{\includegraphics[width=0.495\linewidth]{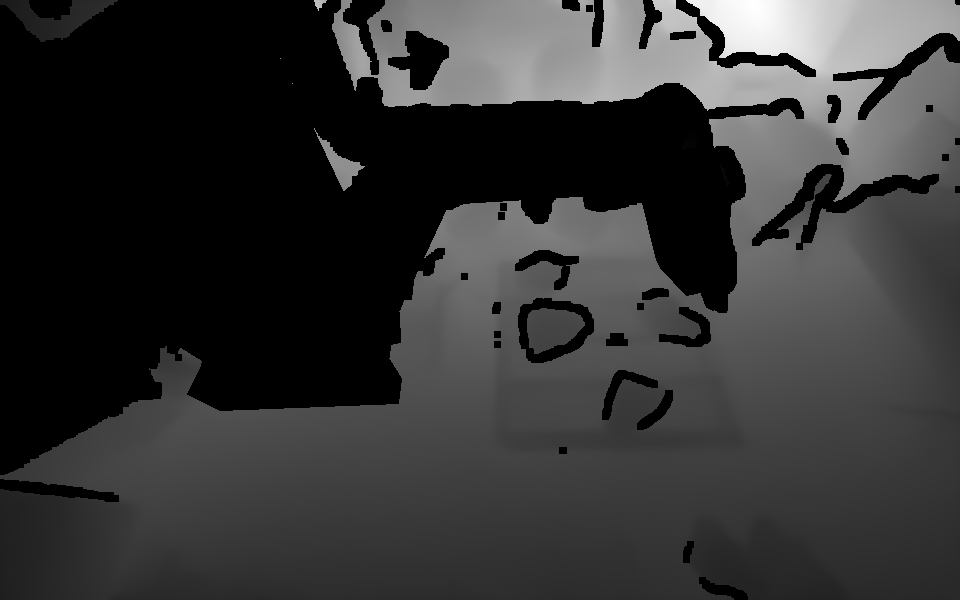}\label{fig:projected_self_filter_c}}\hfill
    \subfloat[]{\includegraphics[width=0.495\linewidth]{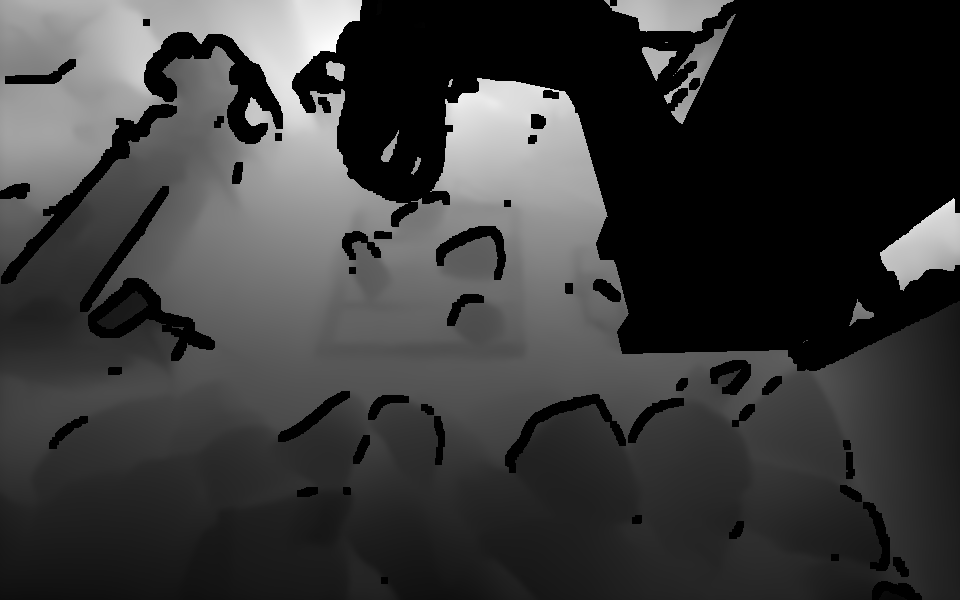}\label{fig:projected_self_filter_d}} \\
     \subfloat[]{\includegraphics[width=0.495\linewidth]{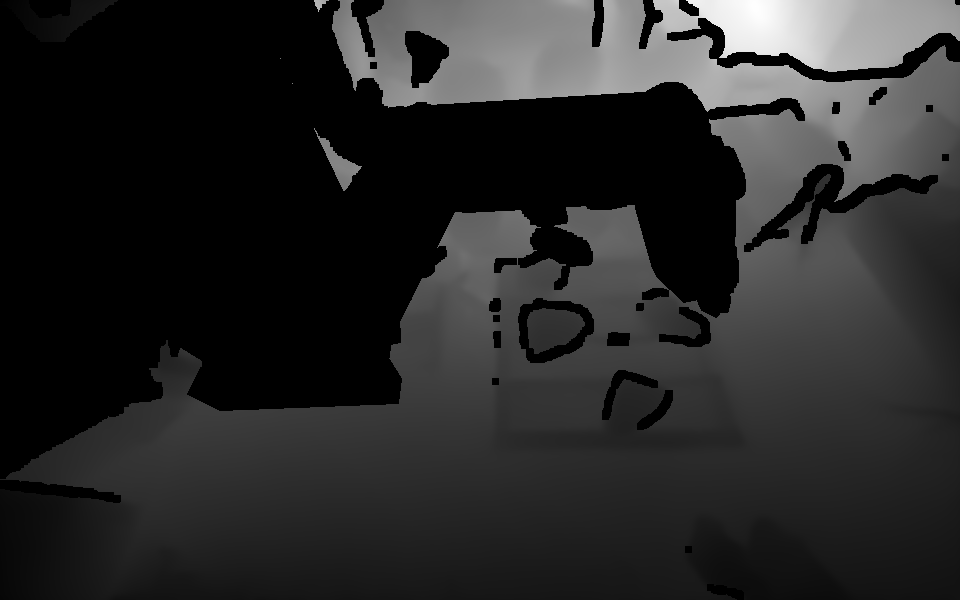}\label{fig:projected_self_filter_e}} \hfill
    \subfloat[]{\includegraphics[width=0.495\linewidth]{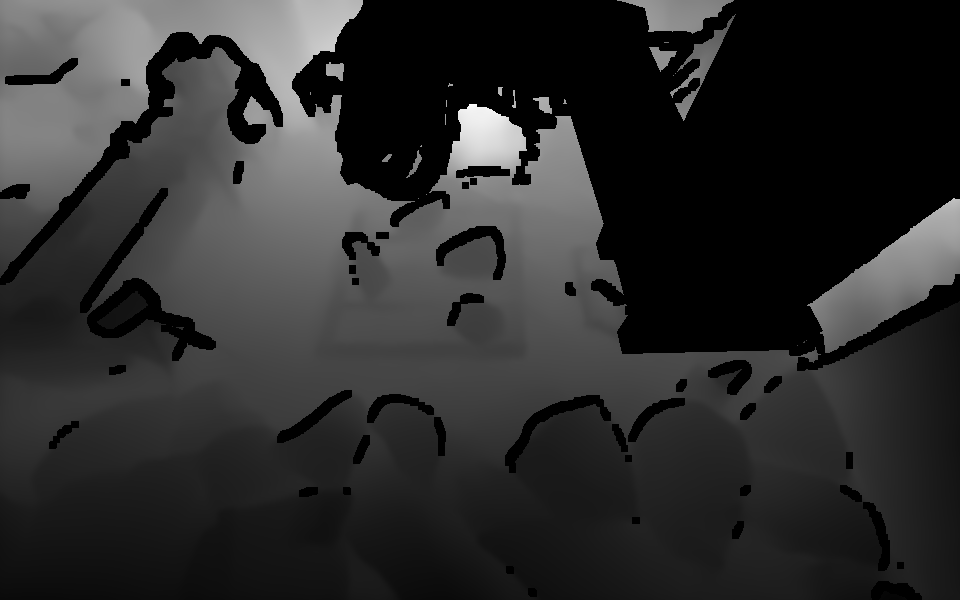}\label{fig:projected_self_filter_f}} \\
     \subfloat[]{\includegraphics[width=0.495\linewidth]{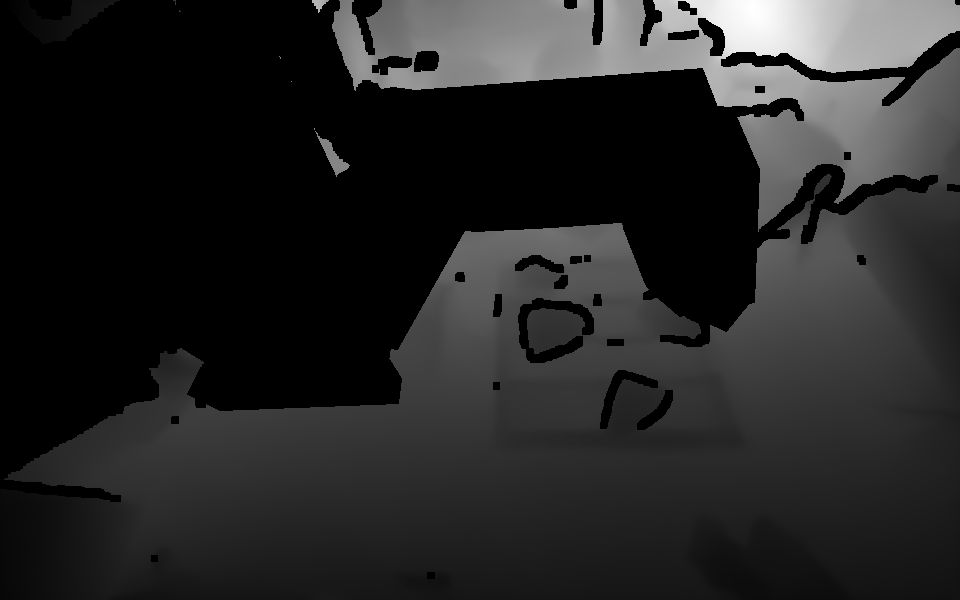}\label{fig:projected_self_filter_g}} \hfill
    \subfloat[]{\includegraphics[width=0.495\linewidth]{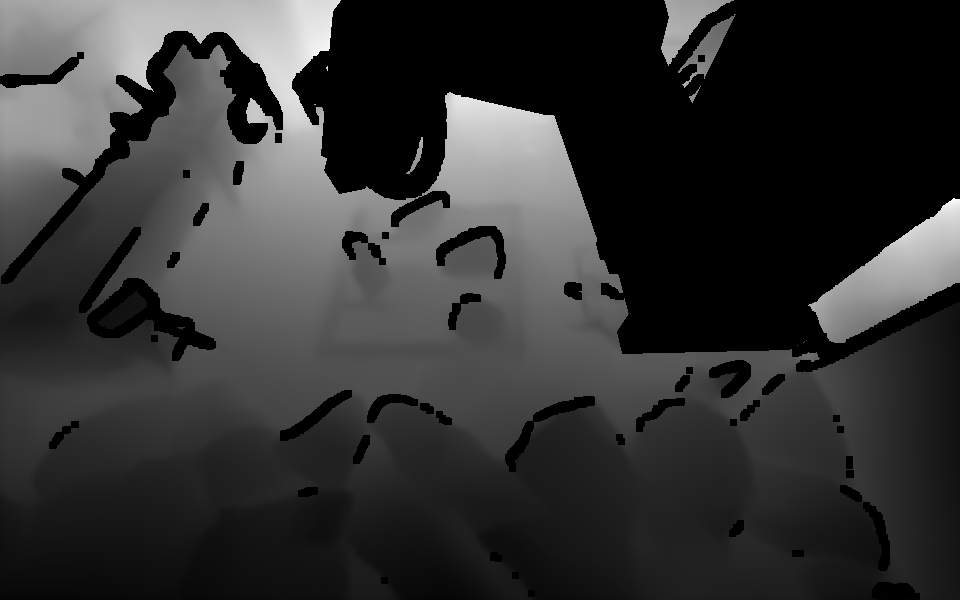}\label{fig:projected_self_filter_h}}
    \caption{Self-filter masks obtained via per-link bounding boxes, projected into the stereo cameras' provided depth images. (a)--(b) Raw depth images; (c)--(d) masked depth images without box offsets; (e)--(f) masked depth images with box offsets set to $0.05$~m; (g)--(h) masked depth images with box offsets set to $0.12$~m.}
    \label{fig:projected_self_filter}
\end{figure}

The result of this procedure when utilizing only oriented bounding boxes is illustrated in Fig.~\ref{fig:projected_self_filter}, where different box offsets are set to highlight their effect on the resulting masks. These offsets are useful, for instance, to exclude the hydraulic hoses, which are not part of the impact hammer model.

\subsubsection{Occlusion management}
\label{subsubsec:occlusion_management}

As the impact hammer moves, it continuously occludes parts of the scene captured by exteroceptive sensors, limiting the amount of information available for perception. To mitigate this effect, we maintain a background model of the static environment using the self-filtered (i.e., masked) depth maps. The key idea is that regions that are temporarily occluded by the hammer become visible again as the arm moves, therefore, by continuously integrating these observations over time, a reference depth map of the static environment can be constructed. During operation, this model is used to fill regions that are temporarily occluded by the hammer, thus, producing a more complete depth map for the subsequent stages of the perception pipeline. The construction of the background model and the corresponding depth filling procedure are described in what follows.

\begin{figure*}[ht]
    \centering
    \subfloat[]{\includegraphics[width=0.23\linewidth]{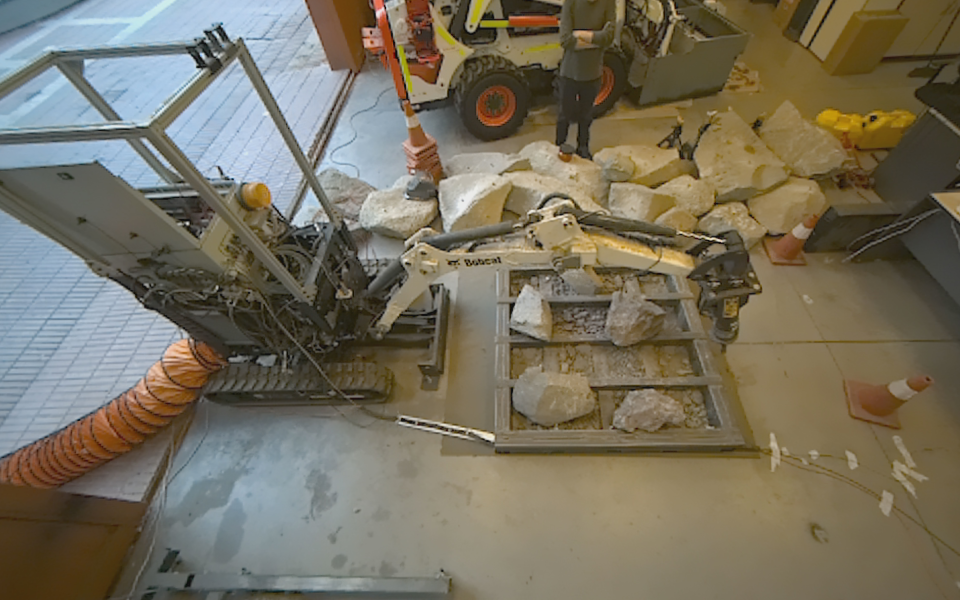}\label{fig:occlusion_management_a}} \quad
    \subfloat[]{\includegraphics[width=0.23\linewidth]{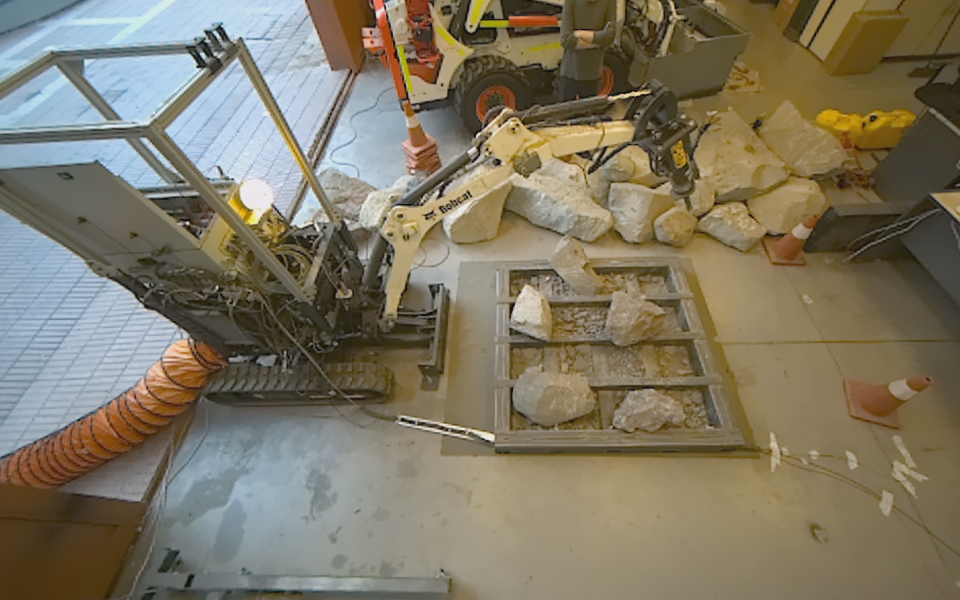}\label{fig:occlusion_management_b}} \quad
    \subfloat[]{\includegraphics[width=0.23\linewidth]{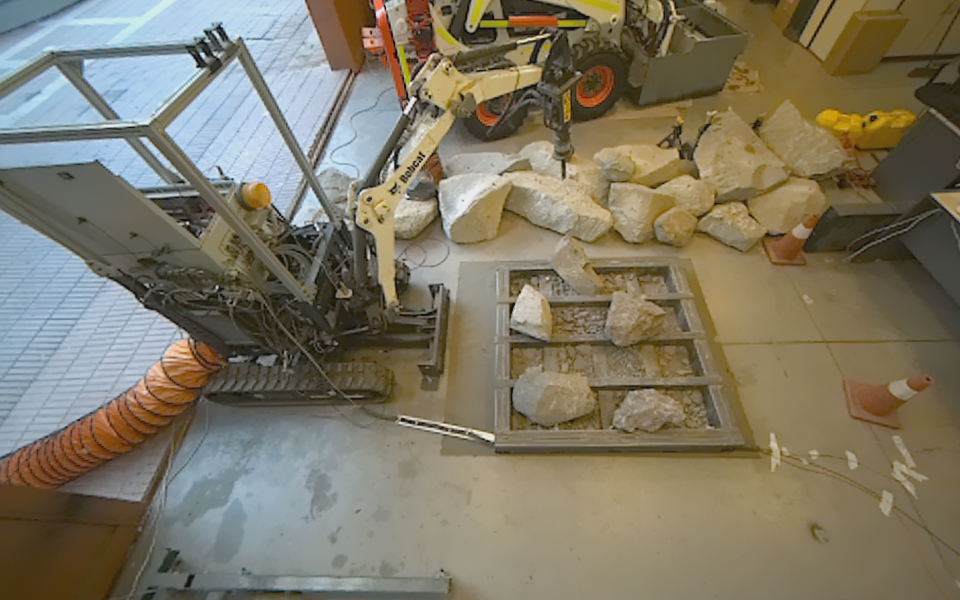}\label{fig:occlusion_management_c}} \quad
    \subfloat[]{\includegraphics[width=0.23\linewidth]{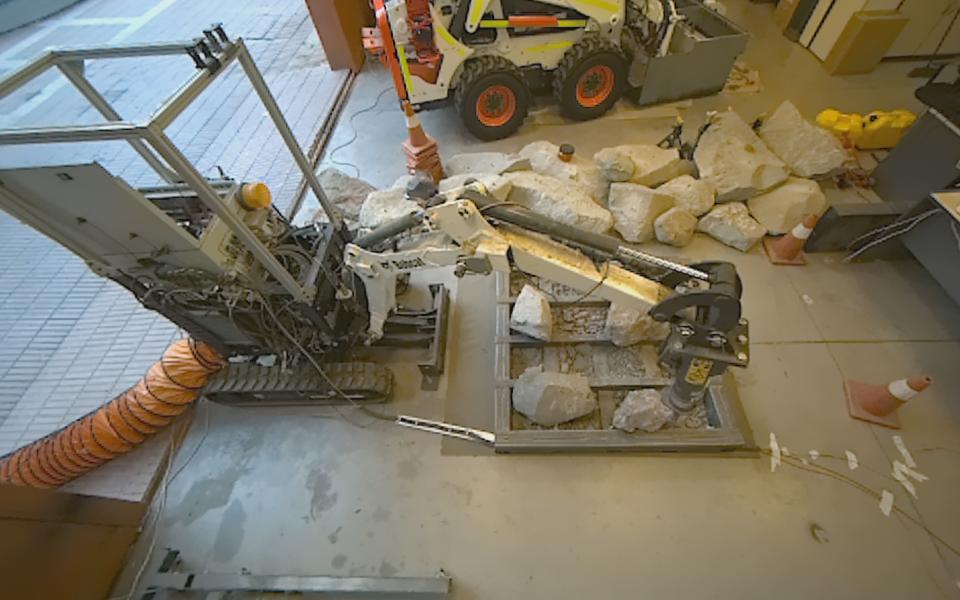}\label{fig:occlusion_management_d}} \\
    \subfloat[]{\includegraphics[width=0.23\linewidth]{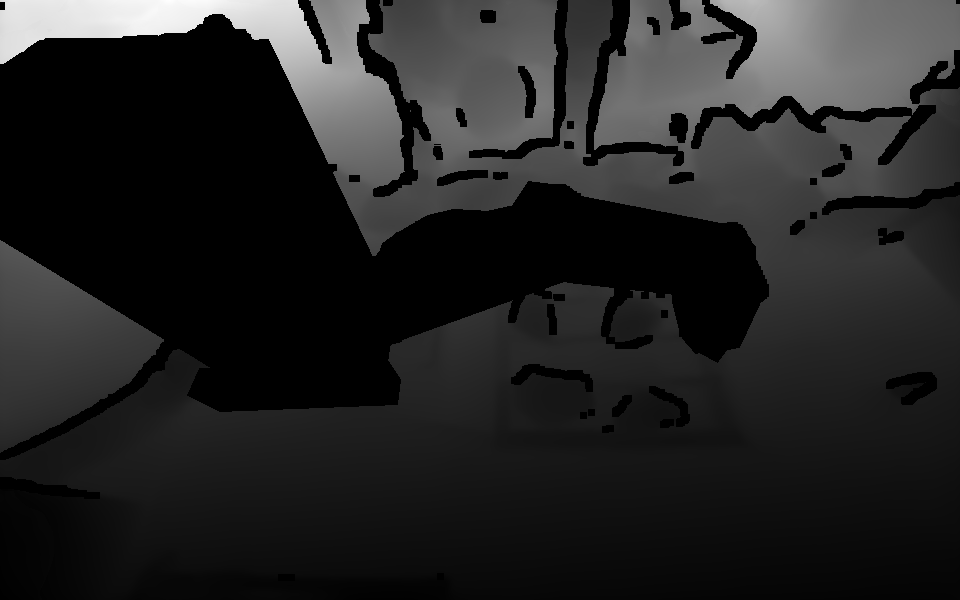}\label{fig:occlusion_management_e}} \quad
    \subfloat[]{\includegraphics[width=0.23\linewidth]{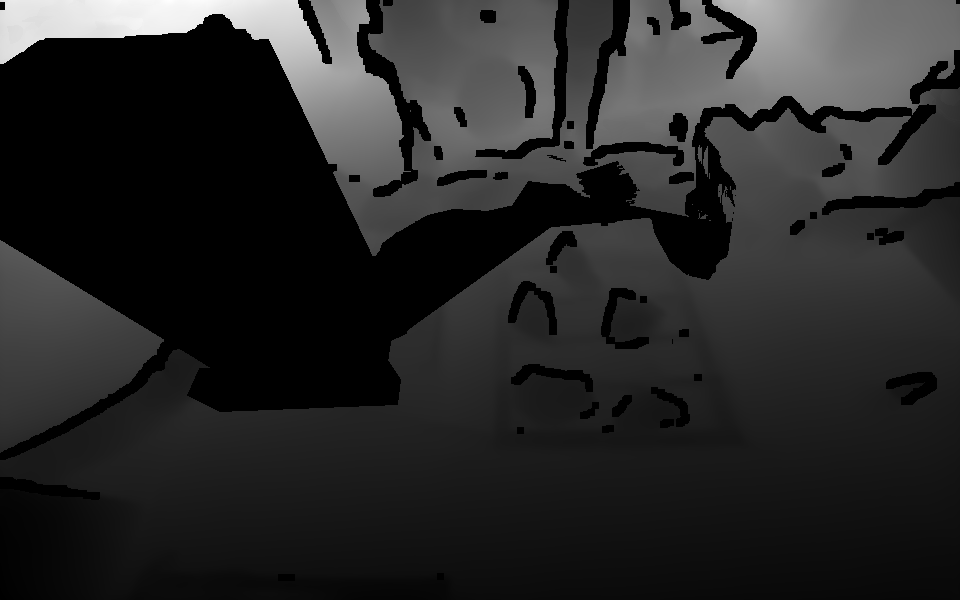}\label{fig:occlusion_management_f}} \quad
    \subfloat[]{\includegraphics[width=0.23\linewidth]{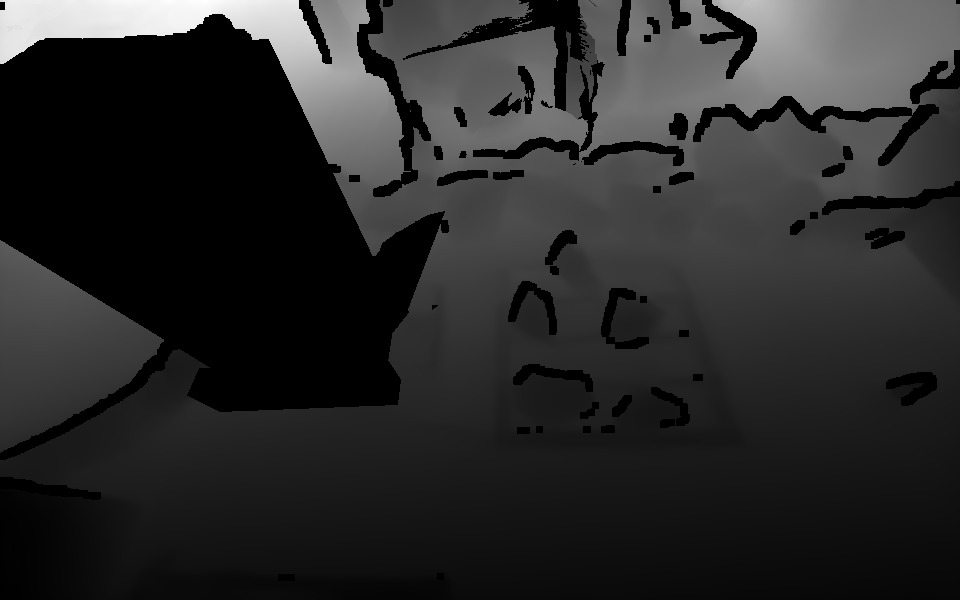}\label{fig:occlusion_management_g}} \quad
    \subfloat[]{\includegraphics[width=0.23\linewidth]{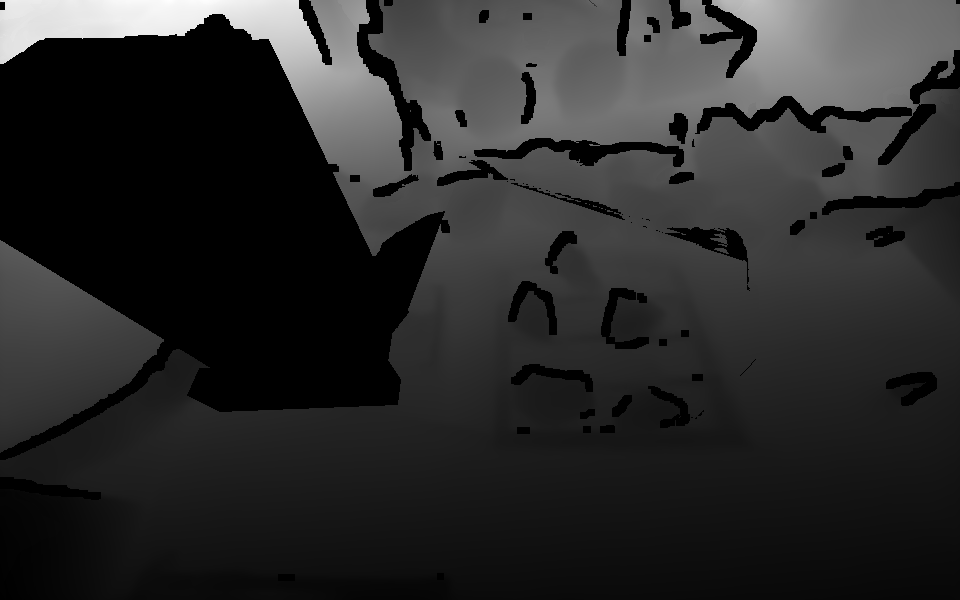}\label{fig:occlusion_management_h}} \\
    \subfloat[]{\includegraphics[width=0.23\linewidth]{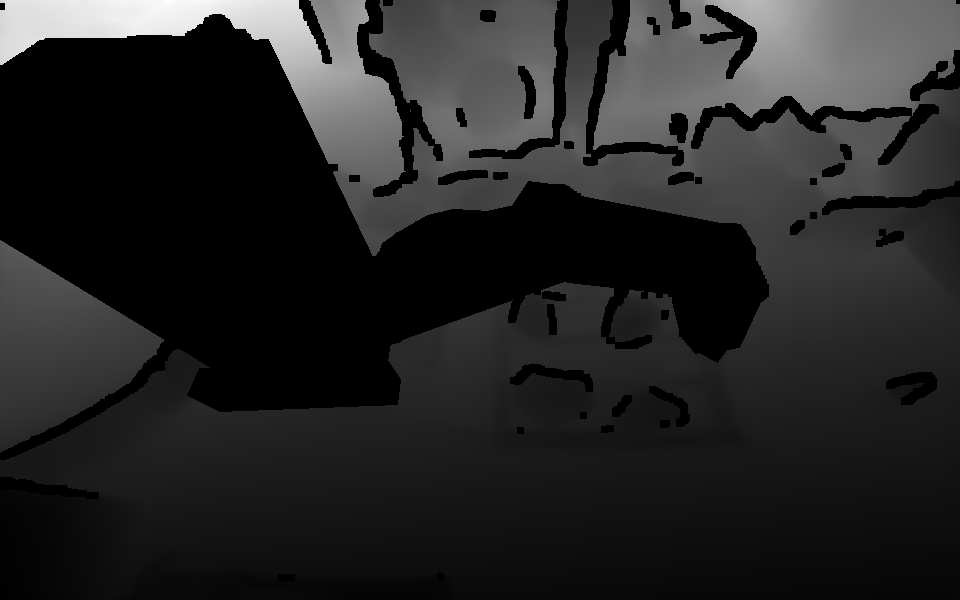}\label{fig:occlusion_management_i}} \quad
    \subfloat[]{\includegraphics[width=0.23\linewidth]{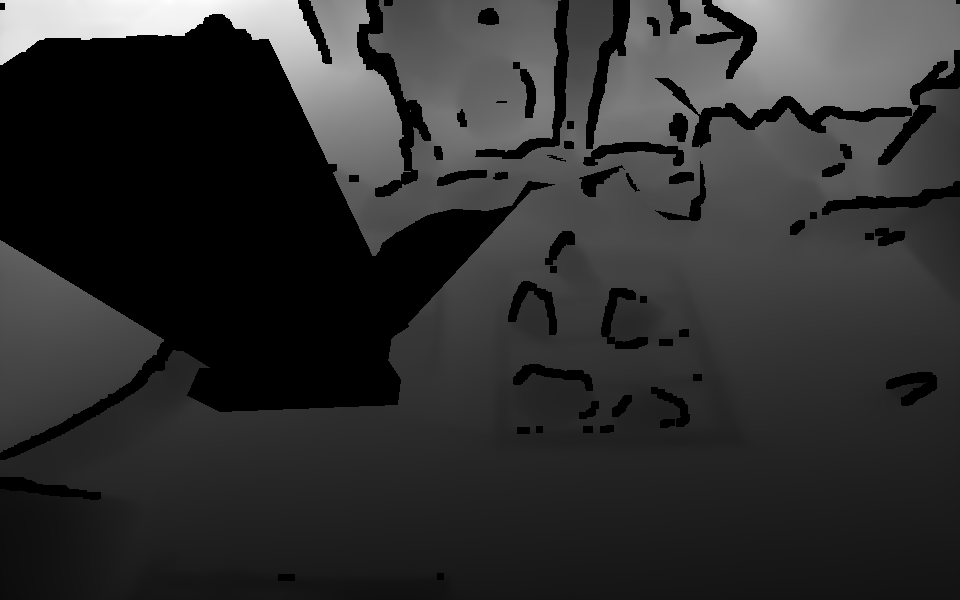}\label{fig:occlusion_management_j}} \quad
    \subfloat[]{\includegraphics[width=0.23\linewidth]{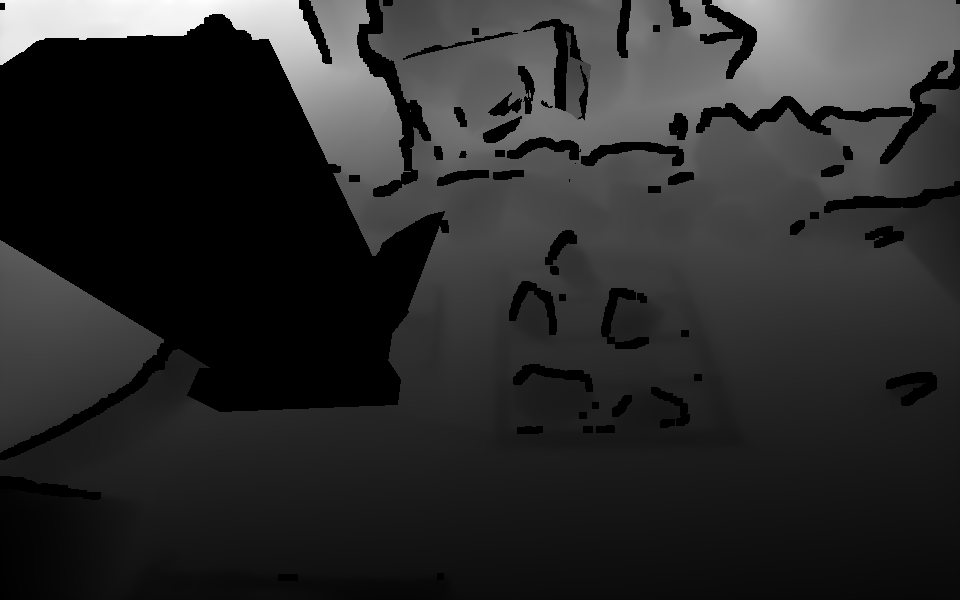}\label{fig:occlusion_management_k}} \quad
    \subfloat[]{\includegraphics[width=0.23\linewidth]{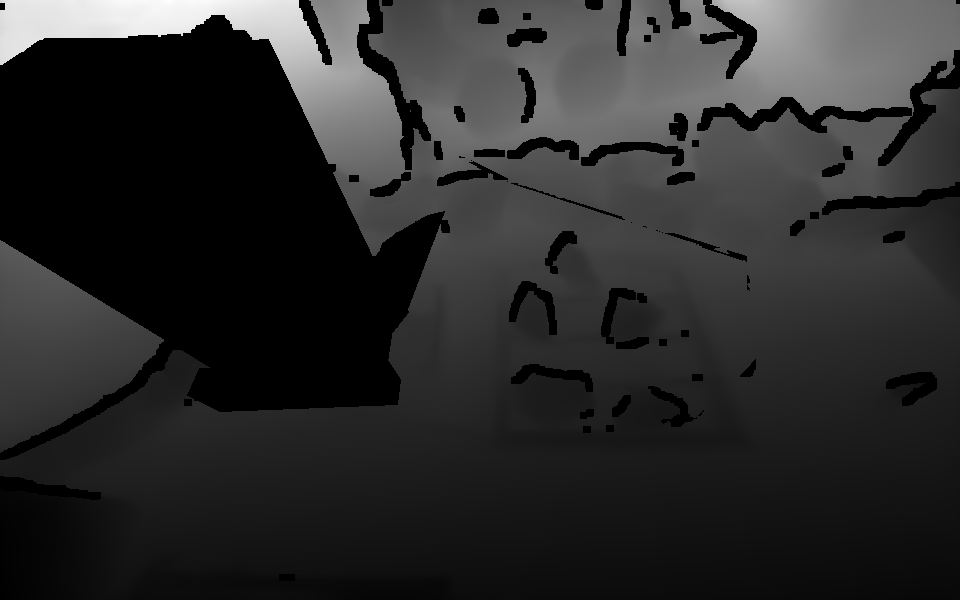}\label{fig:occlusion_management_l}}
    \caption{Time evolution of the background models and filled depth images provided by the occlusion management module for the right stereo camera at indexed frames $t\in\{0, 120, 250, 500\}$. (a)--(d) RGB images, (e)--(h) background models, and (i)--(l) filled depth images.}
    \label{fig:occlusion_management}
\end{figure*}

\paragraph{Background model}
\label{subsubsec:background_model}

We use a background model based on that proposed by~\citet{yang2004real,yang2005real}, in which a two-level background maintenance strategy (at both the pixel and frame levels) was presented. The model uses a dynamic matrix to analyze frame-to-frame pixel variations and determine whether each pixel belongs to the background (i.e., its static or quasi-static) or to the foreground.

Given a stream of time-indexed depth images, $\bm{I}_\text{depth}(t)$, first we define a binary frame-to-frame difference image, $\bm{F}(t)$, via
\begin{equation}
\bm{F}(t)[i,j] = 
\begin{cases}
0  & \text{if } \left|\bm{I}_{\text{depth}}(t)[i,j] - \bm{I}_{\text{depth}}(t - k)[i,j] \right| \leq d_\text{th}, \\
1  & \text{otherwise},
\end{cases}
\label{eq:frame_to_frame}
\end{equation}

where $\bm{I}[i,j]$ denotes the pixel value at image coordinates $(i,j)$, $k$ is the interval between the frames used for comparison, and $d_\text{th}\in\mathbb{R}$ is a scalar threshold.

With the above definition, and for $\lambda \in \mathbb{N}_{>0}$, a dynamic matrix $\bm{D}(t)$ is computed via
\begin{equation}
\bm{D}(t)[i,j] = 
\begin{cases}
\lambda           & \text{if } \bm{F}(t)[i,j] \neq 0, \\
\bm{D}(t-1)[i,j] - 1  & \text{if } (\bm{F}(t)[i,j] = 0) \land \\& \quad (\bm{D}(t - 1)[i,j] \neq 0),\\
\bm{D}(t-1)[i,j]  & \text{otherwise}.
\end{cases}
\label{eq:dynamic_matrix}
\end{equation}

Finally, a binary image $\bm{M}(t)$ is also defined using the result provided by the self-filter (for a given camera), such that $\bm{M}(t)[i,j]:= 1$ in all the image regions where the impact hammer is, and $\bm{M}(t)[i,j]:= 0$ everywhere else.

Thus, denoting the difference image domain as $\Omega$, and for a scalar threshold $v_\text{th}$, the background model $\bm{B}(t)$ is given by
\begin{equation}
\bm{B}(t)[i,j] =
\begin{cases}
\bm{I}_\text{depth}(t)[i,j]   & \text{if }(\bm{D}(t)[i,j] = 0 \, \lor \\
& \frac{1}{|\Omega|}\sum_{(i,j)\in\Omega}\bm{F}(t)[i,j] < v_\text{th} )\land \\
                              & \quad \bm{M}(t)[i,j] = 0, \\
\bm{B}(t-1)[i,j] & \text{otherwise}. \\

\end{cases}
\label{eq:background_model}
\end{equation}
 
This update rule differs from the background model in~\cite{yang2005real}, which uses linear interpolations, as this strategy is not suitable when processing depth maps. Moreover, they do not use a binary mask, which we require to know where the impact hammer is currently occluding the scene.

Finally, since the perception pipeline has two distinct depth images streams (one per stereo camera), $\bm{I}^c_\text{depth}(t)$, with $c\in\{\texttt{left}, \texttt{right}\}$, we maintain two separate background models, each denoted as $\bm{B}^c(t)$, which are constructed using a distinct frame-to-frame difference image $\bm{F}^c(t)$ a dynamic matrix $\bm{D}^c(t)$, and corresponding binary image $\bm{M}^c(t)$.

\paragraph{Filling occluded regions}
\label{subsubsec:filling_occluded_regions}

At every time step, we use the background models described in the previous section, alongside their corresponding depth images, to construct and update a \emph{``filled''} depth image using
\begin{equation}
\bm{I}_{\text{filled}}(t)[i,j] =
\begin{cases}
\bm{I}_\text{depth}(t)[i,j]   & \text{if }\bm{M}(t)[i,j] = 0, \\
\bm{B}(t)[i,j]   & \text{if }\bm{M}(t)[i,j] = 1.
\end{cases}
\label{eq:final_depth_image}
\end{equation}

As for the background models, this is done per camera, resulting in $\bm{I}^c_\text{filled}(t)$, with $c\in\{\texttt{left}, \texttt{right}\}$.

An example of the results provided by this module is illustrated in Fig.~\ref{fig:occlusion_management}. Although certain regions of the scene cannot be reconstructed due to limitations of our impact hammer model (e.g., unmodeled hydraulic hoses), the background models capture the motion of the impact hammer and enable the filtered depth images to be updated using the most recent measurements available in the occluded areas.

\subsection{Stage 3: Image and point cloud processing}
\label{subsec:image_and_point_cloud_processing}

The third stage of the pipeline comprises all filtering and processing operations, in both images and point clouds, that aim at characterizing the environment to provide relevant information to perform rock fragmentation. This stage consists of two parallel branches that address rock segmentation following complementary approaches: (i) via instance segmentation of rocks in RGB images, followed by the projection of the resulting masks onto their corresponding depth maps, and (ii) via filtering point clouds so that only those points above the grizzly and that are not associated to the impact hammer itself remain. Each approach induces distinct processing paths in the pipeline, which include dedicated sub-modules, as illustrated in Fig.~\ref{fig:streamlined_perception_pipeline}. We describe both approaches in detail in the following sub-sections.

\subsubsection{Image rocks segmentation}
\label{subsubsubsec:image_rocks_segmentation}

\begin{figure*}
        \centering
        \includegraphics[width=0.98\linewidth]{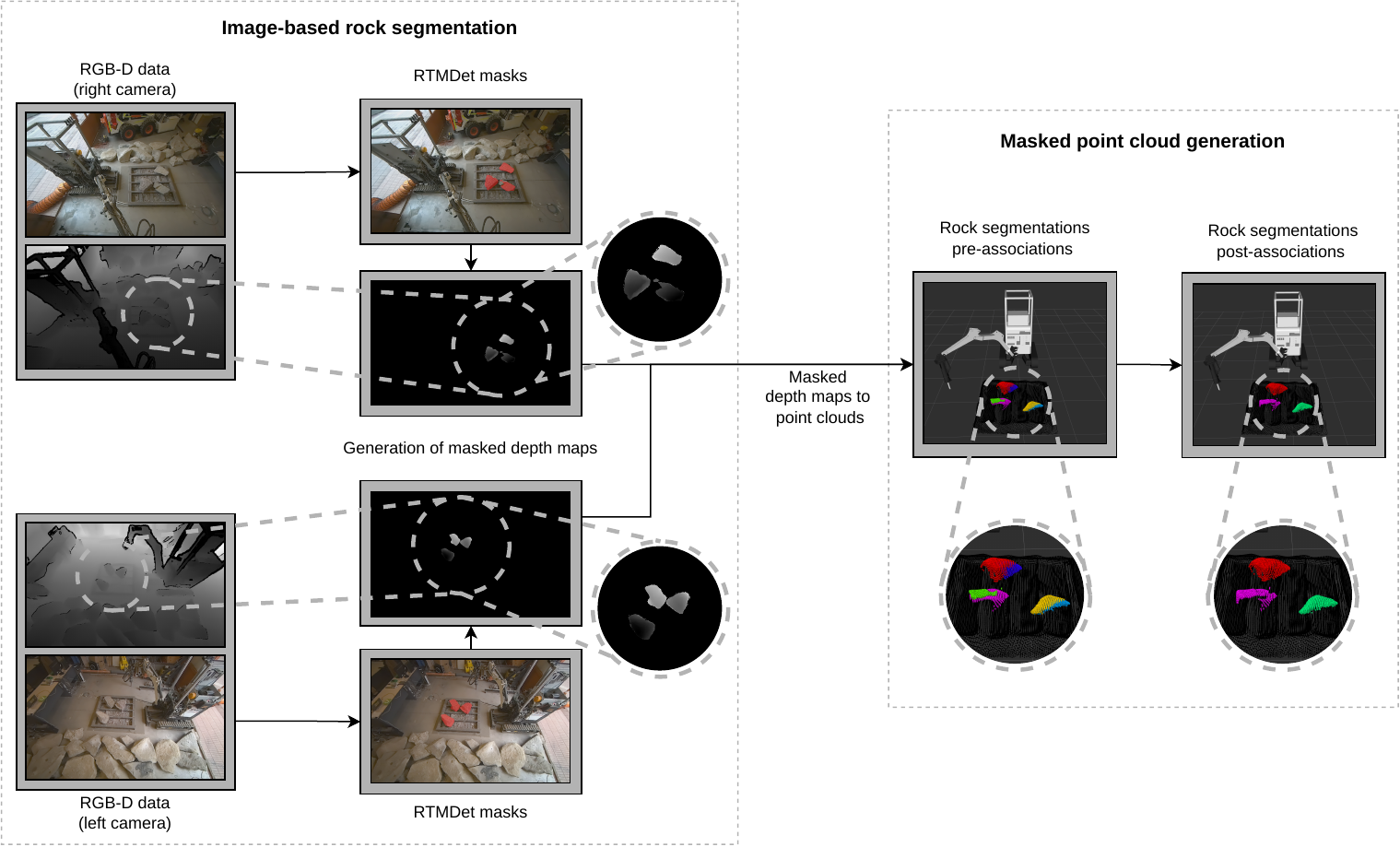}
        \caption{Visual segmentation process diagram.}
    \label{fig:visual_segmentation}
\end{figure*}

This module employs the RTMDet model~\citep{lyu2022rtmdet} to perform instance segmentation of rocks in RGB images. Similarly to \citet{ruiz2025system}, the inputs to the RTMDet model are generated by pre-processing the original images to extract the regions (in pixel coordinates) that contain the steel grate. These regions of interest (ROIs) are fixed, camera-dependent and calibrated offline. Moreover, ROIs are rescaled to a resolution of $480\times480$ pixels before being fed to the segmentation model. Once the segmentation masks are obtained, they are mapped back to the original RGB image resolution ($960\times600$ pixels). Subsequently, their new coordinates are used to extract the segmented rocks from the depth maps associated to each RGB image, while preserving their corresponding labels.

The above procedure results in two types of image per processed RGB-D data instance: a ``labeled'' image and a masked depth map. The labeled images encode each segmented rock with a distinct color. The masked depth maps, on the other hand, only preserve depth values in the same pixel coordinates in which segmented rocks are present in the labeled images, while all the other pixels are masked with a distinct invalid value (using \texttt{NaNs}). The described process is illustrated in the left-side block shown in Fig.~\ref{fig:visual_segmentation}.

\subsubsection{Masked point cloud generation and fusion}
\label{subsubsubsec:masked_pointcloud_generation_and_fusion}

The masked depth maps and the ``labeled'' images obtained from the \emph{``image rocks segmentation''} module described above are transformed into 3D point clouds (using Eq.~\eqref{eq:projected_2d_points}) and then referenced with respect to $\bm{T}_\text{aux}$ (see Sect.~\ref{subsubsec:icp_calibration}), with a prior application of a bilateral filter to the depth maps for noise reduction. The result of this process is a labeled point cloud in which each point inherits the color of the corresponding ``labeled'' image and represents a specific rock. Note that from App.~\ref{app:software_architecture} (see the ``\texttt{pcl\_vis\_rock\_segmentator}'' module in Fig.~\ref{fig:software_architecture}), here an Open3D-based implementation of the bilateral filter is used; further details can be found in App.~\ref{app:bilateral_filter}. 

The final step consists of identifying the corresponding detections between the left and right cameras that belong to the same physical rock, in order to generate a merged point cloud with a single label assigned to each rock. To achieve this, the 3D centroid of each segmented rock's point cloud is computed. The Euclidean distances between each centroid from the left camera and all centroids from the right camera are then calculated (whenever detections are available; otherwise, this step is skipped). These distances are arranged into a distance matrix (which is not necessarily a square matrix) and each detection from the left camera is associated with the closest detection from the right camera without allowing repeated assignments. In other words, once a detection from one camera has been matched, it cannot be assigned to another detection from the opposite camera. 

If the associated distance between left and right detection is smaller than a predefined threshold, then the label of the left detection is propagated to the right detection; otherwise, the segmentations are considered different and each detection maintain its distinct label. In the case left or right instances do not have a right or left match (due to the presence of more detections in one of the two cameras) the label of those detections remains, as it means that a given rock is just segmented from one perspective. The described process is illustrated in the right-side block shown in Fig.~\ref{fig:visual_segmentation}.

\subsubsection{Point cloud generation and fusion}
\label{subsubsubsec:pointcloud_generation_and_fusion}

This step of the pipeline fuses the point clouds generated using the pre-processed depth maps provided by each sensor after applying the self-filtering (Sect.~\ref{subsubsec:filtering_process}) and occlusion management (Sect.~\ref{subsubsec:occlusion_management}) procedures over them. For each depth map, we first generate their corresponding point clouds using Eq.~\eqref{eq:projected_2d_points}. This produces the local point clouds $\mathcal{P}_{\text{left}}$ and $\mathcal{P}_\text{right}$.

Using the calibration procedure outlined in Sect.~\ref{subsubsec:icp_calibration}, the poses of each stereo camera are determined with respect to an auxiliary frame $\bm{T}_\text{aux}\in\text{SE}(3)$. Denoting these poses as $\bm{T}_\text{left}$ and $\bm{T}_\text{right}$, we can obtain a fused point cloud via
\begin{equation}
    \mathcal{P}_{\text{fused}} = 
    \left\{ \bm{T}_{\text{left}}\, \bm{p} \;\middle|\; \bm{p} \in \mathcal{P}_\text{left} \right\} 
    \cup
    \left\{ \bm{T}_{\text{right}}\, \bm{p} \;\middle|\; \bm{p} \in \mathcal{P}_{\text{right}} \right\}.
    \label{eq:combined_pointcloud}
\end{equation}

Finally, a box filter is applied over this merged point cloud. This filter reduces the number of points and removes those that do not provide relevant information about the environment, specifically, points located far from the steel grate.

\subsubsection{Point cloud filters}
\label{subsubsubsec:pointcloud_filters}

\begin{figure}
        \centering
        \subfloat[]{\includegraphics[width=0.48\linewidth]{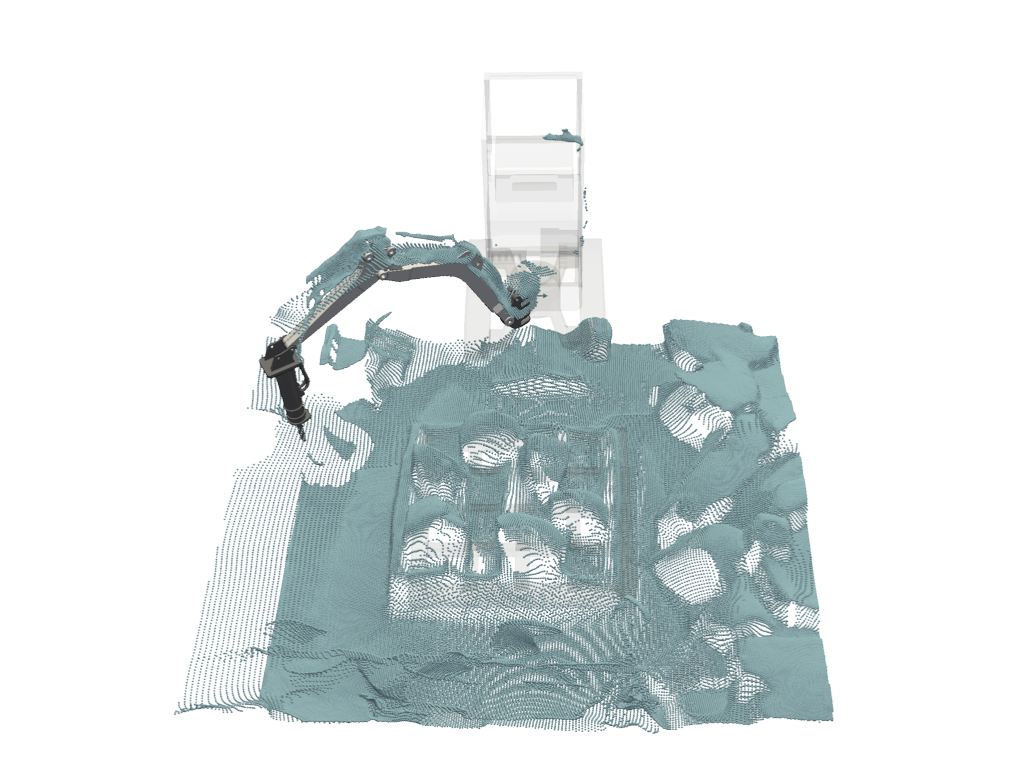}\label{fig:raw_point_cloud}} \quad
        \subfloat[]{\includegraphics[width=0.48\linewidth]{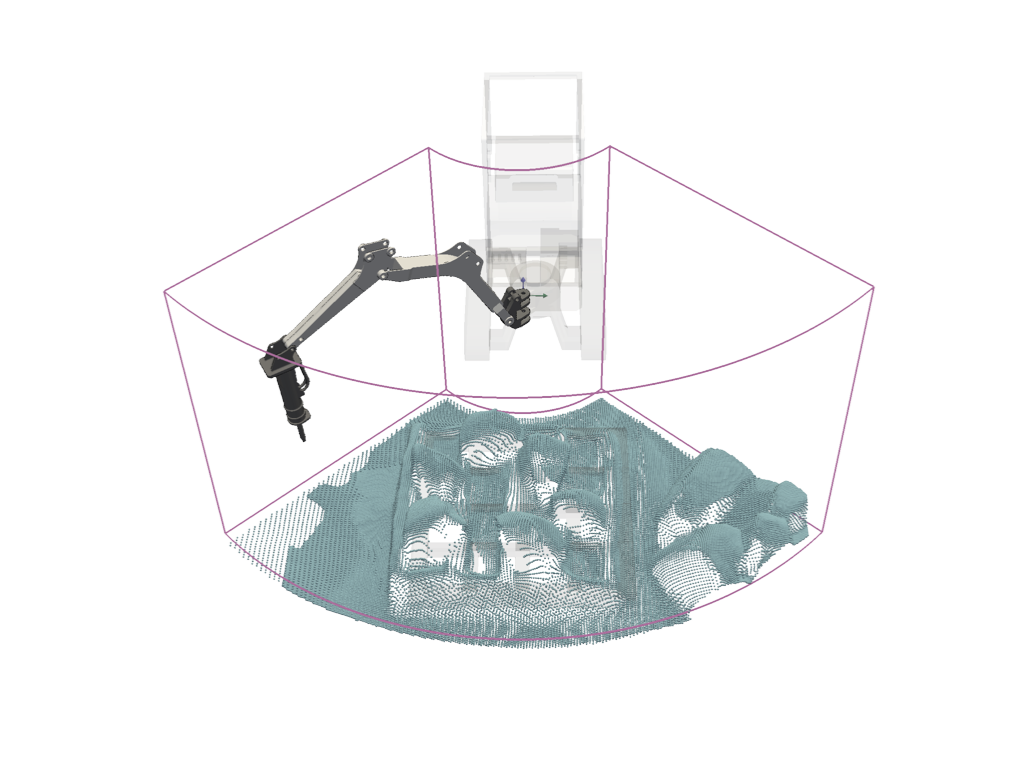}\label{fig:self_filter_plus_ws_filter}} \\
        \subfloat[]{\includegraphics[width=0.48\linewidth]{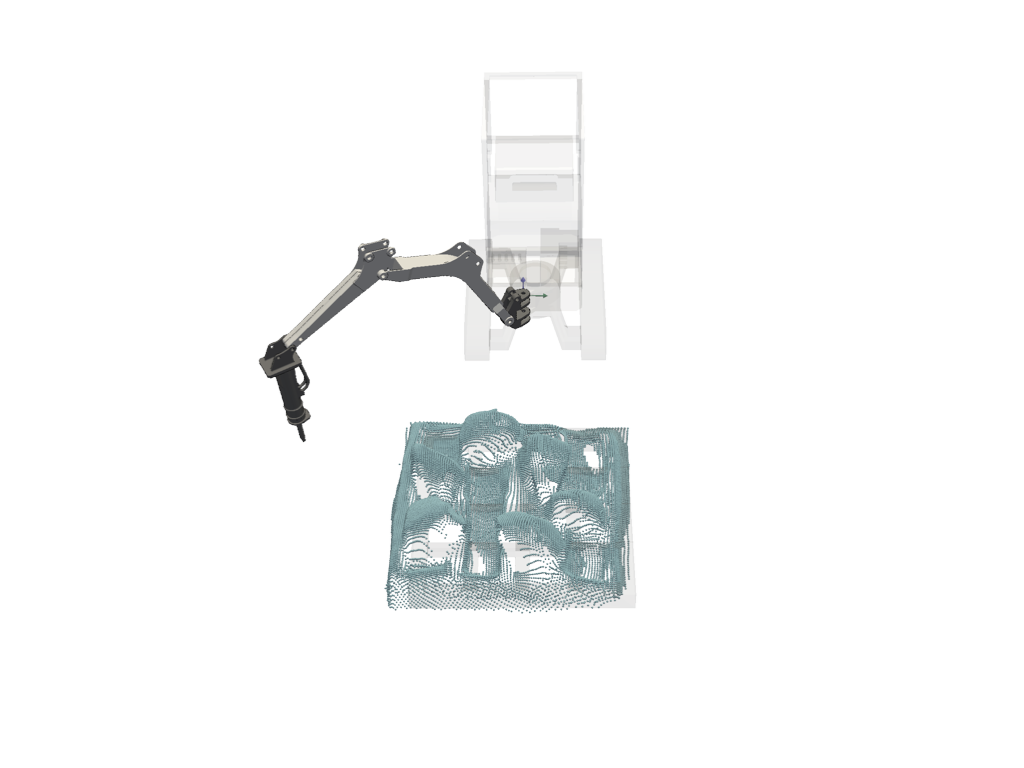}\label{fig:self_filter_plus_box_filter}} \quad
        \subfloat[]{\includegraphics[width=0.48\linewidth]{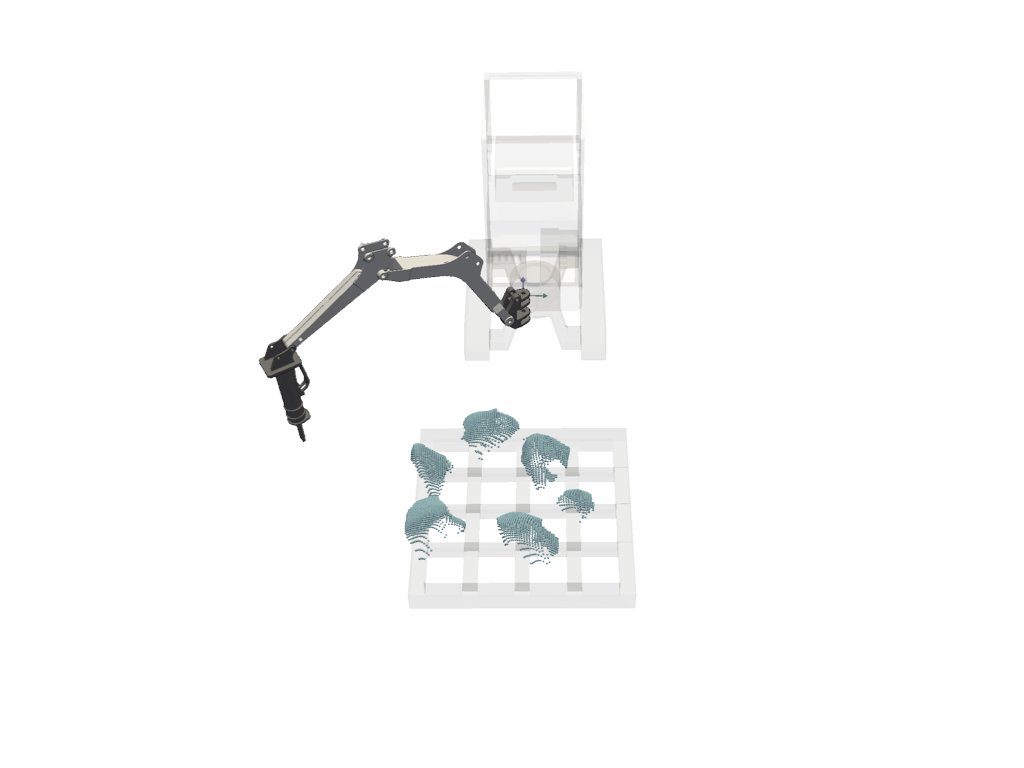}\label{fig:self_filter_plus_box_filter_plus_grill_filter}}
        \caption{Example results of the application of point cloud filters. (a) Input point cloud; (b) workspace filter followed by self-filter; (c) box filter followed by self-filter; (d) box filter followed by self-filter, and then followed by the grizzly filter (using $\mathcal{P}_{\text{grizzly}}$, see Sect.~\ref{subsubsubsec:grill_filter}).}
    \label{fig:result_workspace_filter}
    \end{figure}

These filters operate directly in $3$D space, with point clouds as both inputs and outputs. Their design is based on geometric constraints or geometric primitives that define the regions of space to be filtered. The results of applying (different compositions of) the filters described in this section are shown in Fig.~\ref{fig:result_workspace_filter}.

\paragraph{Workspace filter}
\label{subsubsubsec:workspace_filter}

The extent of the environment represented by the fused point cloud may go beyond the impact hammer's workspace. Since points that are far from this region generally do not provide useful information for control or perception tasks, they are removed.

To do so, we consider a restricted workspace modeled as a cylindrical sector around the impact hammer's ``swing'' joint rotational axis. This geometric representation captures both the minimum and maximum angular and radial reach of the impact hammer's end-effector (as interpreted in planar cylindrical coordinates), and provides upper and lower limits via planar surfaces associated to the mine's roof and floor. Moreover, these planes are allowed to have independent orientations, enabling a more realistic representation of the operational environment.

To filter the points outside the restricted workspace, we first transform every point $\bm{p}\in\mathcal{P}_\text{fused}$ to the fixed frame associated to the rotational axis of the ``swing'' joint, and project them to the lower and upper oriented planes defining the restricted workspace. Let us denote the resulting points as $\bm{p}_\text{swing}$, $\bm{p}_\text{floor}$ and $\bm{p}_\text{roof}$. Then, for radial limits $[r_\text{min}, r_\text{max}]$, angular limits $[\theta_\text{min}, \theta_\text{max}]$, and $z_\text{min}>0$, any point $\bm{p}$ for which its transformed versions fail to fulfill any of the conditions given by Eqs.~\eqref{eq:r_comp}--\eqref{eq:z_comp} is filtered out.
\begin{gather}
    r_\text{min} \leq \sqrt{p_{\text{swing}, x}^2+p_{\text{swing}, y}^2} \leq r_\text{max}\label{eq:r_comp} \\
    \theta_\text{min} \leq \mathrm{atan2}(p_{\text{swing}, y},p_{\text{swing}, x}) \leq \theta_\text{max} \label{eq:theta_comp}\\
    (p_{\text{floor},z} > z_\text{min}) \land (p_\text{roof,z}<0) \label{eq:z_comp}
\end{gather}

\paragraph{Box filter}
\label{subsubsubsec:box_filter}

This filter determines whether points lie inside or outside a given 3D box, and filters out those outside of it. This is helpful to isolate points within the steel grate bounds, where rocks that should be characterized by the perception pipeline are. This filter is implemented by using the box signed distance field (SDF) equation given by
\begin{equation}
    \mathrm{SDF}_{\text{box}}(\bm{p},\bm{s}) = \|\max(\bm{q},0)\|_2 + q,
    \label{eq:box_sdf}
\end{equation}
where $\bm{p}\in\mathbb{R}^3$ is a point referenced to the box's origin, $\bm{s}=(s_x, s_y,s_z)^T$ is the box's extent, $\bm{q} = |\bm{p}|-\bm{s}$, and $q = \min(\max(q_x, q_y, q_z), 0)$.

To apply this filter, the points that are going to be processed are first referenced with respect to the box's origin. Let us denote the resulting points as $\bm{p}_\text{box}$ and the grizzly's extent as $\bm{s}_\text{grizzly}$. Then, the filtered point cloud is given by
\begin{equation}
    \mathcal{P}_\text{box}=\left\{\bm{p}\in\mathcal{P}_{\text{fused}} \middle| \mathrm{SDF}_\text{box}(\bm{p}_\text{box}, \bm{s}_\text{grizzly})\leq0\right\}.
\end{equation}

\paragraph{Grizzly filter}
\label{subsubsubsec:grill_filter}

To get the rocks from the region of the scene characterized by $\mathcal{P}_\text{box}$, we need to filter out all points belonging to the steel grate. As in~\cite{cardenas2022automatic}, this is done by first constructing a model of the steel grate offline, and then ``subtracting'' it from $\mathcal{P}_\text{box}$. 

The steel grate model is built by capturing $n$ depth snapshots of empty scenes from the stereo cameras (i.e., scenes where there are no rocks on top of the grizzly), converting them into fused point clouds, and then adding them into a single large unified point cloud. This point cloud is box-filtered (see Sect.~\ref{subsubsubsec:box_filter}) so as to match the grizzly dimensions, and then voxelized to remove redundant points. The resulting point cloud, which we will denote as $\mathcal{P}_{\text{grizzly}}$, corresponds to the grizzly model.

To filter $\mathcal{P}_\text{box}$ using the grizzly model we consider two methods. The first method relies on the sphere SDF equation given by 
\begin{equation}
\mathrm{SDF}_{\text{sph}}(\bm{p}, r) = \|\bm{p}\|_2 - r,
\label{eq:sphere_sdf}
\end{equation}
where $\bm{p}\in\mathbb{R}^3$ is a point referenced to the sphere's origin, and $r>0$ is the sphere's radius. Denoting as $\bm{p}_\text{sph}$ to the point that results from referencing $\bm{p}\in\mathcal{P}_\text{box}$ with respect to a given point $\bm{p_\text{grizzly}}\in\mathcal{P}_\text{grizzly}$, filtering the grizzly model from $\mathcal{P}_\text{box}$ can be subsumed as constructing the set given by
\begin{equation}
    \mathcal{P}_\text{rocks} =\left\{ \bm{p}\in\mathcal{P}_\text{box} \middle| \min_{\bm{p}_\text{grizzly}\in\mathcal{P}_\text{grizzly}}\left(\mathrm{SDF}_\text{sph}(\bm{p}_{\text{sph}}, r)\right) > 0\right\}.
\end{equation}

In contrast, the second method approximates the steel grate model and subdivides it into $k$ groups using the $k$-means algorithm~\citep{lloyd1982least}. Then, it computes the bounding boxes associated with the resulting partitions. The SDF of each point in $\mathcal{P}_\text{box}$ is subsequently evaluated with respect to these boxes using~\eqref{eq:box_sdf}. Denoting the set of boxes' origins and their extents as pairs $(\bm{p}_{k\text{-box}}, \bm{s}_\text{box})\in \mathcal{B}_{\text{box}}$, applying the grizzly model filter in this case consists of constructing the set
\begin{equation}
    \mathcal{P}_\text{rocks} =\left\{ \bm{p}\in\mathcal{P}_\text{box} \middle| \min_{(\bm{p}_{k\text{-box}}, \bm{s}_\text{box})\in\mathcal{B}_{\text{box}}}\left(\mathrm{SDF}_\text{box}(\bm{p}_{\text{box}}, \bm{s}_\text{box})\right) > 0\right\}
\end{equation}
where $\bm{p}_\text{box}$ results from representing a given point $\bm{p}\in \mathcal{P}_\text{box}$ with respect to a given point $\bm{p}_{k\text{-box}}$.

\begin{figure}
        \centering
        \subfloat[]{\includegraphics[width=0.48\linewidth]{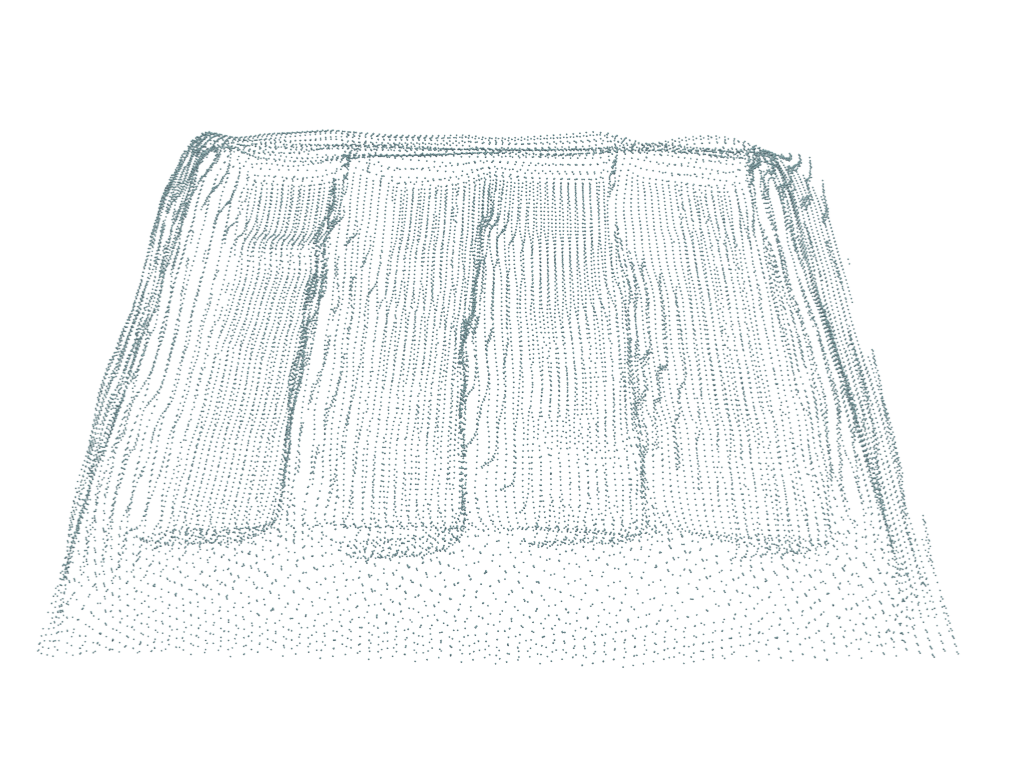}\label{fig:grill_models_exact}} \quad
        \subfloat[]{\includegraphics[width=0.48\linewidth]{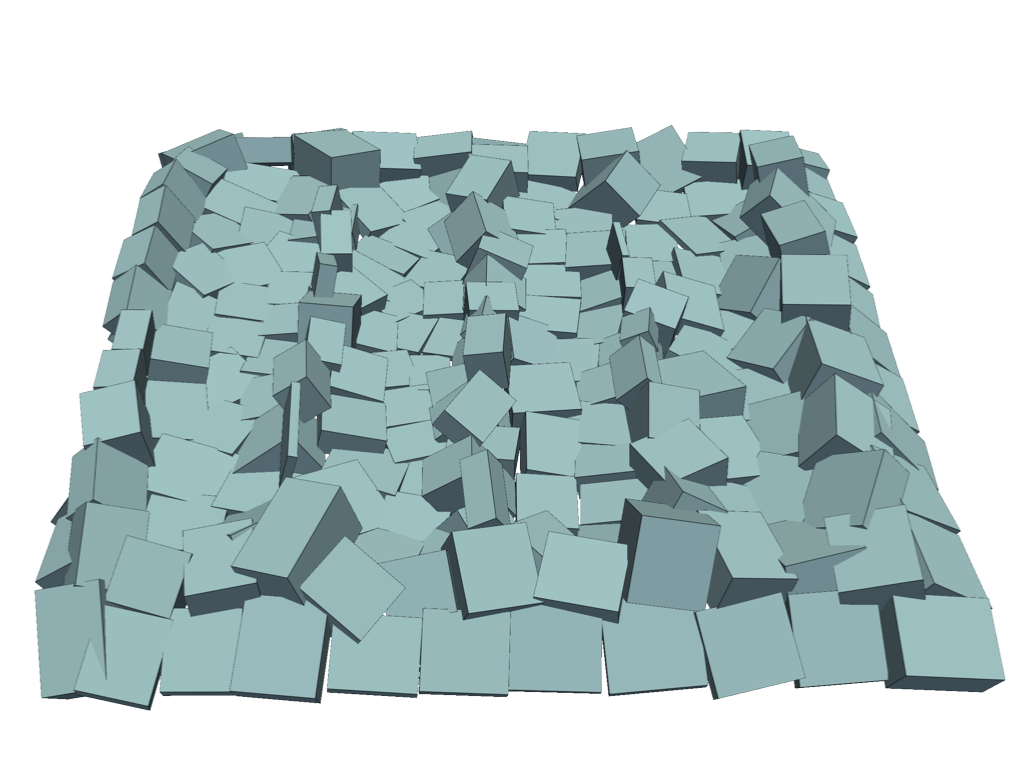}\label{fig:grill_models_approximate}}
        \caption{Steel grate models. (a) Point cloud model, $\mathcal{P}_{\text{grizzly}}$. (b) Boxes-based model, $\mathcal{B}_{\text{box}}$.}
    \label{fig:grill_models}
\end{figure}

The representation of both the grizzly model, $\mathcal{P}_\text{grizzly}$, and its approximate version, $\mathcal{B}_\text{box}$, are shown in Fig.~\ref{fig:grill_models}. The experimental evaluation presented in Sec.~\ref{sec:experiments} uses the grizzly filtering method that uses $\mathcal{P}_\text{grizzly}$ and the sphere SDF equation defined previously.

\subsubsection{Point cloud rocks segmentation}
\label{subsubsec:point_cloud_rocks_segmentation}

\begin{figure}
        \centering
        \subfloat[]{\includegraphics[width=0.30\linewidth]{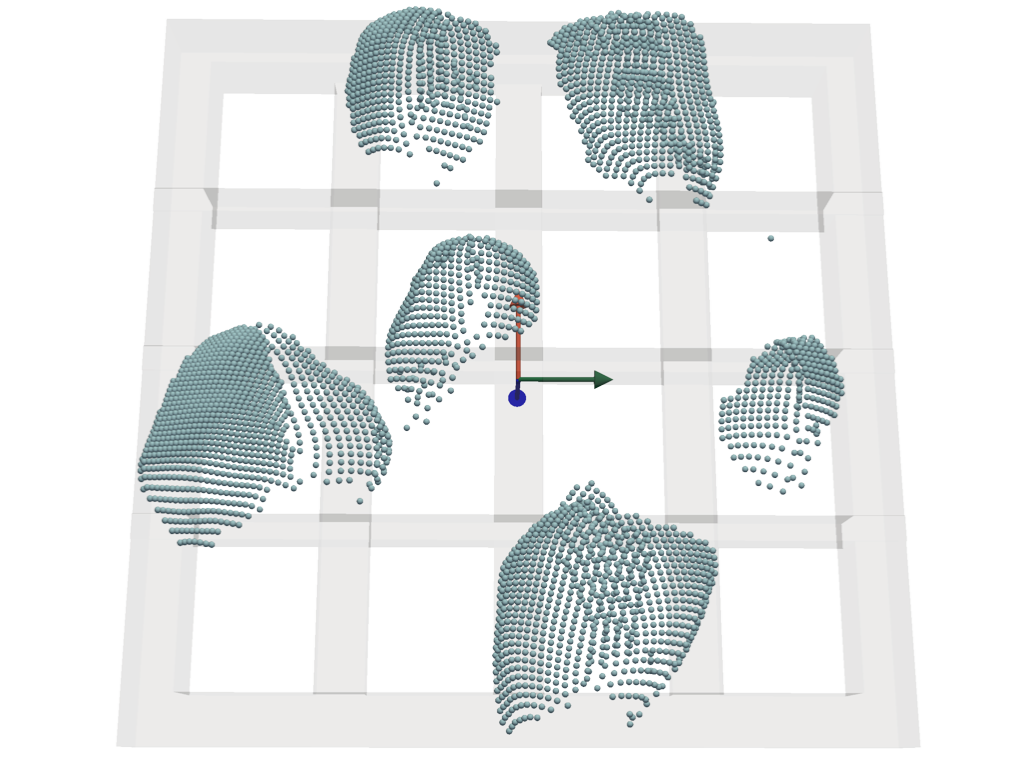}\label{fig:geometric_segmentation_input}} \quad
        \subfloat[]{\includegraphics[width=0.30\linewidth]{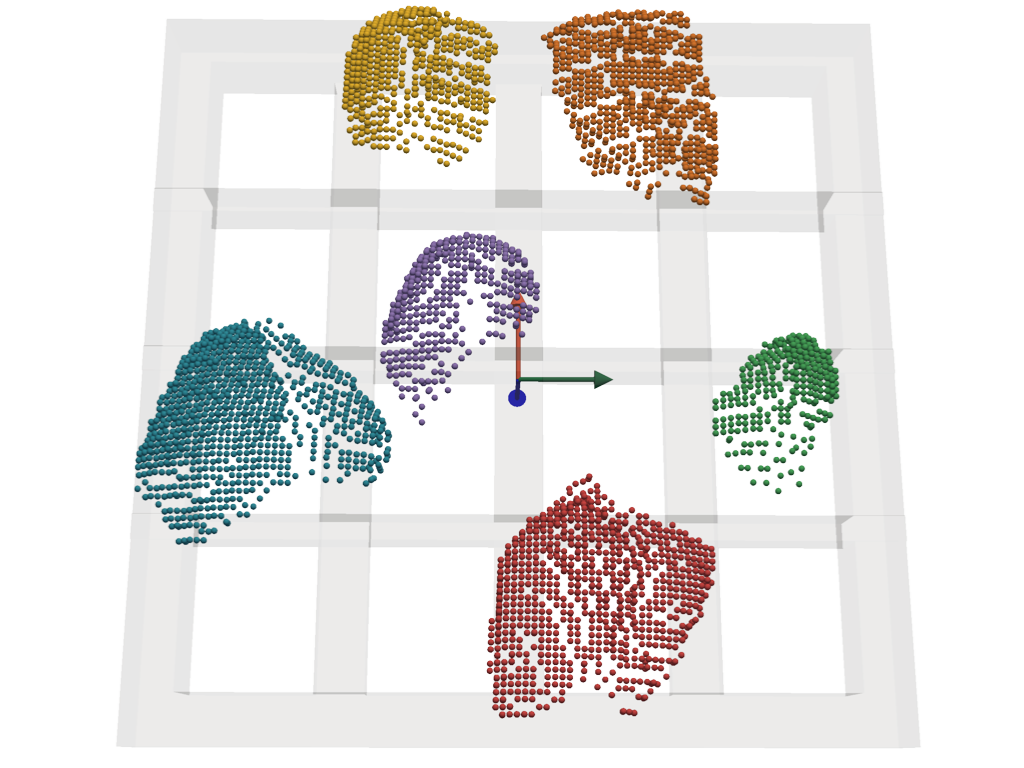}\label{fig:geometric_segmentation_1}} \quad
        \subfloat[]{\includegraphics[width=0.30\linewidth]{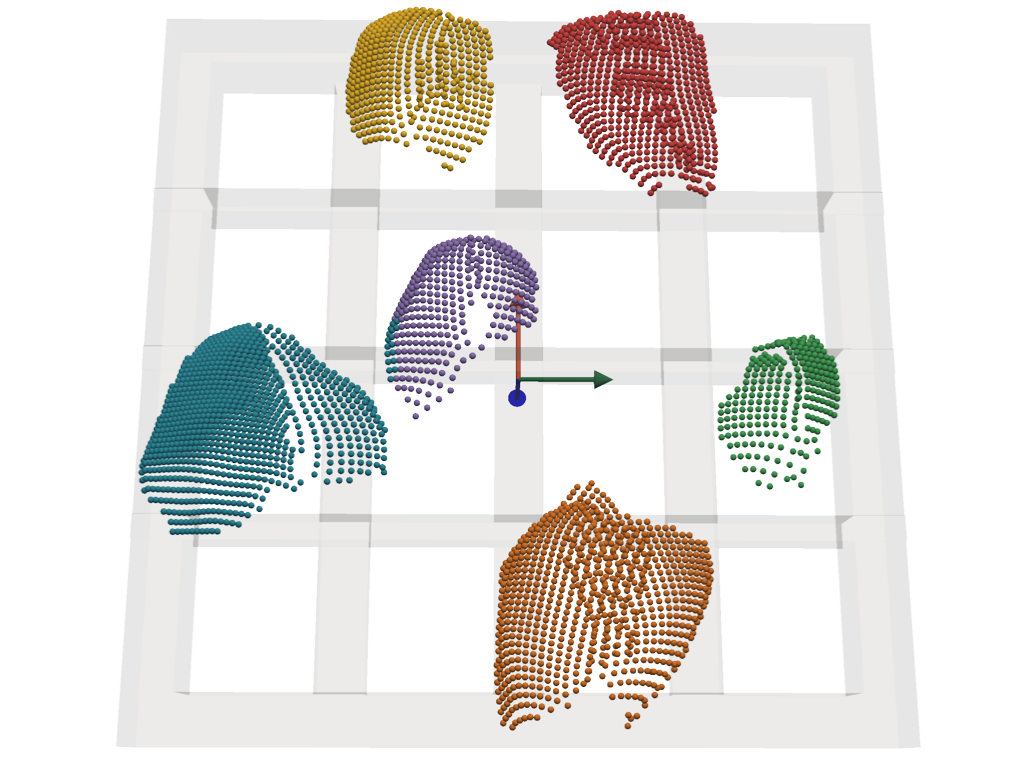}\label{fig:geometric_segmentation_2}}
        \caption{Point cloud rocks segmentation process. (a) Input point cloud; (b) voxelization, DBSCAN clustering and clusters filtering process; (c) re-projection to the original, denser point cloud.}
    \label{fig:geometric_segmentation}
\end{figure}

The instance segmentation of point clouds associated to rocks consists of four steps, which are summarized in what follows; the result of their application is illustrated in Fig.~\ref{fig:geometric_segmentation}.

\begin{itemize}
    \item Voxelization: The filtered input point cloud is voxelized to reduce its size so as to achieve a faster computation in the following processing steps.
    \item Clustering: In this step the DBSCAN algorithm~\citep{ester1996density} is applied to the voxelized input point cloud to group points that are spatially close to one another, and to mark as outliers points that lie in low-density regions.
    \item Cluster filtering: After obtaining the clusters, the number of points and the dimensions of their bounding boxes are evaluated to determine whether each cluster corresponds to a rock or to unfiltered noise.
    \item Re-projection to original point cloud: To preserve as much information as possible, for each remaining cluster, its points are defined as all points from the input point cloud (pre-voxelization) that lie within its bounding box.
\end{itemize}

\subsubsection{3D Hammer self-filter}
\label{subsubsubsec:3d_hammer_self_filter}

This final filter can be optionally applied in a given input point cloud instead of the self-filtering method described in Sect.~\ref{subsec:self_filtering}. This filter is useful in cases where the stereo cameras are installed in places such that disregarding information between them and the robot may not be suitable, or when the utilized sensors simply do not provide depth maps (e.g., if using 3D LiDARs).

To apply this filter, each point is transformed into the coordinate system of every subdivision of each mesh of the impact hammer (see Sect~\ref{subsubsec:robot_mesh_subdivision}), and the signed distance field (SDF) of a box (Eq.~\eqref{eq:box_sdf}) or sphere (Eq.~\eqref{eq:sphere_sdf}) is computed, depending on the shape of the subdivisions. For the hydraulic cylinders (see Sect.~\ref{subsec:modeling_hammer_cylinders}), we take advantage of the analytical expression of the cylinder SDF; for a given point $\bm{p}\in\mathbb{R}^3$ referenced to the origin of a cylinder, points $\bm{a},\bm{b}$ corresponding to the cylinders' endpoints, and a radius $r>0$, this signed distance is computed via 
\begin{equation}
    \mathrm{SDF}_{\text{cyl}}(\bm{p}, \bm{a}, \bm{b}, r) 
= \frac{\sign(d)\sqrt{|d|}}{B},
\label{eq:cylinder_sdf}
\end{equation}
where $B=\|\bm{b}-\bm{a}\|_2^2$ is the squared length of the cylinder, and $d$ is computed by defining $P = (\bm{p}-\bm{a}) \cdot (\bm{b}-\bm{a})$, such that,
\begin{equation} 
\begin{aligned}
x &= \| B (\bm{p}-\bm{a}) - P (\bm{b}-\bm{a})\|_2 - r B, \\
y &= \bigl|P - \tfrac12 B\bigr| - \tfrac12 B,\\
x^2_{+} &= \begin{cases}
        x^2 & x > 0 \\
        0   & x \leq 0
        \end{cases}, \qquad 
y^2_{+} = \begin{cases}
        y^2 B & y > 0 \\
        0   & y \leq 0
        \end{cases},\\
d &= 
\begin{cases}
-\min(x^2, y^2 B) & \text{if} \max(x,y) < 0,\\
x^2_{+} + y^2_{+} & \text{otherwise}.
\end{cases}\\
\end{aligned}
\end{equation}

With the above, the impact hammer SDF for a given point $\bm{p}$ is computed by getting the minimum signed distance considering all the boxes or spheres representing its geometry, and the three cylinders representing the actuators on its arm. Finally, all points with an associated signed distance below or equal to a predefined threshold are removed from the point cloud.

\subsection{Stage 4: Rock-breaking pose generation}
\label{subsec:rock_breaking_pose_generation}

\subsubsection{Rock-breaking poses generator}
\label{subsubsec:breaking_poses_generator}

The rock-breaking target poses are derived through a geometric analysis that computes surface normals, and then applies several constrains over the result. This approach aims to identify flat rock faces to generate and prioritize candidate rock-breaking poses with a reduced risk of end-effector slippage. The proposed method is similar to that described in~\cite{lampinen2021robust} or in~\cite{cardenas2022automatic}, where local curvature analysis is also performed, however, here we do not enforce the impact hammer's end effector to be strictly oriented vertically relative to the grizzly.

In our approach, the generation of target rock-breaking poses is conducted independently for each segmented rock and it is applied to both the rocks segmented by the image-based and point cloud-based execution branches of the pipeline's third stage (see Sect.~\ref{subsec:image_and_point_cloud_processing}). Let us denote the set of segmented rocks (i.e., the point clouds produced by either execution branch) as $\{\mathcal{P}_{\text{rock}}^{(i)}\}_{i=1...N}$. For each point $\bm{p}\in\mathcal{P}_\text{rock}^{(i)}$ we compute a local surface normal vector, $\bm{n}$, and then apply the DBSCAN algorithm to cluster the pairs $(\bm{p}, \bm{n})$, thus, grouping surface regions with similar orientations.

Afterwards, for each cluster, a ``proto'' rock-breaking pose candidate is defined such that its position corresponds to the point closest to the cluster centroid, and its orientation is computed as the mean surface normal of the points in the cluster. Let us denote a given valid cluster as the set of point-normal pairs $\mathcal{C_\text{rock}}$, such that $|\mathcal{C_\text{rock}}|>P_\text{min}$, and its centroid as $\bm{p}_c\in\mathbb{R}^3$, then the proto rock-breaking pose has a position given by
\begin{equation}
    \bm{p}_\text{{proto}} = \arg \min_{\bm{p} \in \mathcal{C}_\text{rock}} \|\bm{p}_c - \bm{p}\|_2,
  \label{eq:projected_surface_normal}
\end{equation}
and an orientation defined by the normal vector
\begin{equation}
    \bar{\bm{n}}_\text{proto} = \frac{1}{|\mathcal{C_\text{rock}}|}\sum_{\bm{n}\in\mathcal{C}_\text{rock}} \bm{n}.
    \label{eq:average_surface_normal}
\end{equation}

After computing the ``proto'' pose, an initial soft filtering step is applied over it. If the angle between the (normalized) mean surface normal and the $z$-axis (of the impact hammer base link) exceeds a predefined threshold, i.e., 
\begin{equation}
\label{eq:soft_angle_constraint}
\arccos \left(\bar{n}_{\text{proto},z}\right) > \theta_{\text{soft}},
\end{equation}
then the rock-breaking proto pose candidate is discarded.

Since the impact hammer only possesses four degrees of freedom (so that its end-effector cannot reach arbitrary orientations, and in particular, its roll Euler angle is fully fixed locally), the proto pose normal is projected onto the $t$-plane defined by $\bm{p}_\text{proto}$, the rotation center of the hammer arm, and the $z$-axis, to determine the Euler pitch angle that the hammer should achieve for the corresponding rock-breaking pose.

Each projected normal vector is computed via
\begin{equation}
    \bm{n}_{\text{proj}} = \bar{\bm{n}}_\text{proto} - (\bar{\bm{n}}_\text{proto} \cdot \bm{n}_{t}) \bm{n}_{t},
    \label{eq:pojected_surface_normal}
\end{equation}
with $\bm{n}_t$ denoting the normal vector of the $t$-plane, and computed as
\begin{equation}
    \bm{n}_{t} = ((p_{t,x}, p_{t,y}, 0)^{\top} \times \hat{\bm{z}})/\| (p_{t,x}, p_{t,y}, 0)^{\top} \times \hat{\bm{z}} \|_2,
    \label{eq:plane_normal_vector}
\end{equation}

where $p_{t,x}$ and $p_{t,y}$ represent the $x$- and $y$-coordinates of $\bm{p}_\text{proto}$, but measured with respect to the rotation axis link. With the projected normal vectors computed, the roll, pitch, and yaw Euler angles associated to the orientation of the rock-breaking pose for each cluster $\mathcal{C}_\text{rock}$ can be computed via Eqs.~\eqref{eq:roll_comput}--\eqref{eq:yaw_comput}.
\begin{align}
    \theta_\text{roll} &= 0, \label{eq:roll_comput}\\
    \theta_\text{pitch} &= -\arccos \left(\frac{\bm{n}_{\text{proj}} \cdot (p_{t,x}, p_{t,y}, 0)^{\top}}{\| \bm{n}_{\text{proj}} \|_2 \| (p_{t,x}, p_{t,y}, 0)^{\top} \|_2}\right)- \frac{\pi}{2}, \label{eq:pitch_comput}\\
    \theta_\text{yaw} &= \mathrm{atan2}\left(p_{t,y},p_{t,x}\right) \label{eq:yaw_comput}.
\end{align}

Note that the pitch angle is offset with respect to the pitch that would be defined in the impact hammer's base link frame, since the hammer must strike towards the rock.

Finally, a hard constraint pitch filter is applied with respect to the grizzly base link frame, such that if the angle is lower than a given threshold, $\theta_\text{hard}$, the breaking pose is maintained and it becomes eligible for the impact hammer. It is important to clarify that $\theta_\text{soft} \gg \theta_\text{hard}$ (see Eq.~\eqref{eq:soft_angle_constraint}). This hard constraint is applied as follows:
\begin{equation}
    \frac{\pi}{2} - \arccos \left(\frac{\bm{n}_{\text{proj}} \cdot (p_{t,x}, p_{t,y}, 0)^{\top}}{\| \bm{n}_{\text{proj}} \|_2 \| (p_{t,x}, p_{t,y}, 0)^{\top} \|_2}\right) \leq \theta_\text{hard}.
    \label{eq:hard_constrain}
\end{equation}

\begin{figure}[]
        \centering
        \subfloat[]{\includegraphics[width=0.98\linewidth]{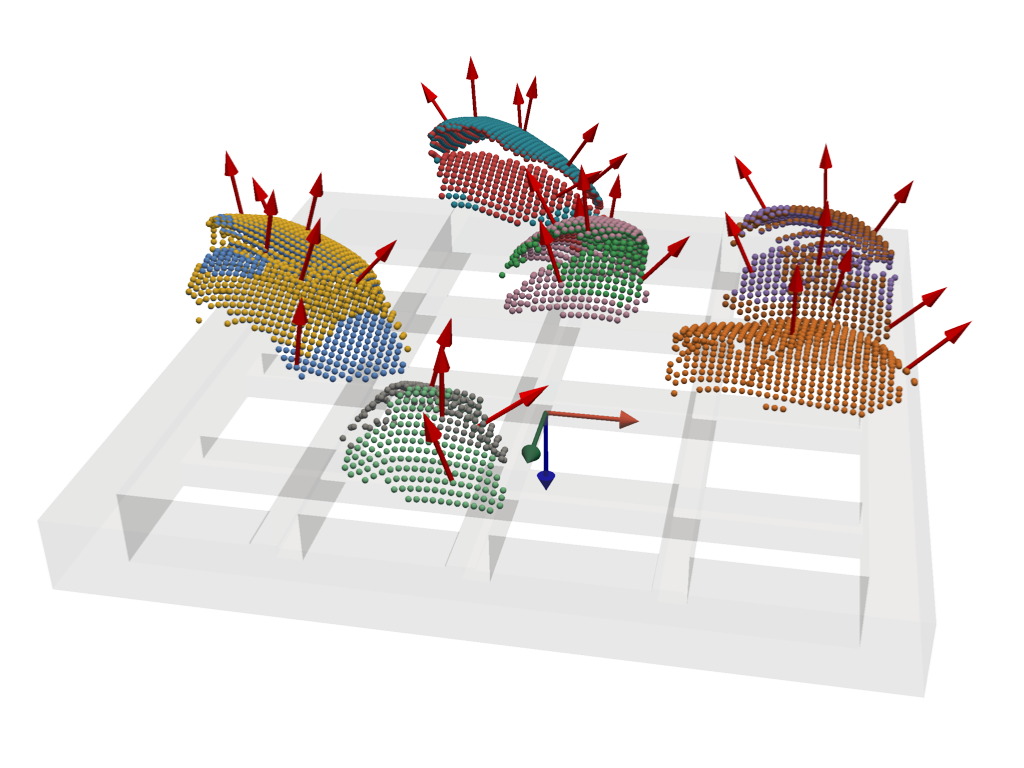}\label{fig:target_poses_generator_1}} \\
        \subfloat[]{\includegraphics[width=0.98\linewidth]{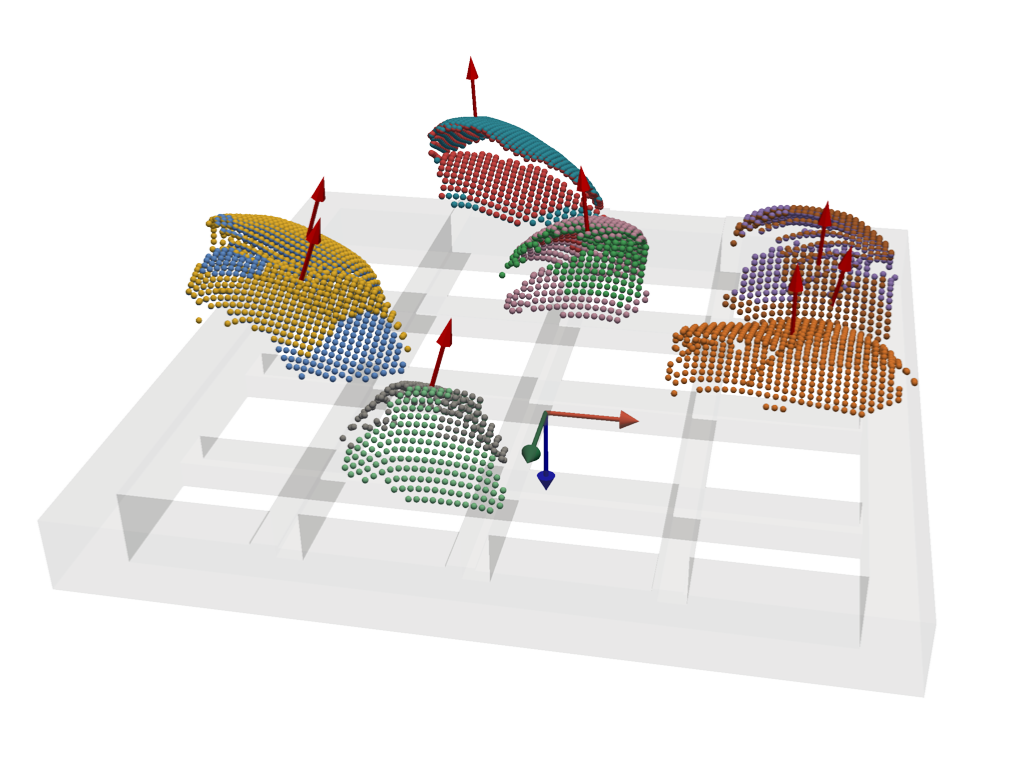}\label{fig:target_poses_generator_2}} 
        \caption{Target rock-breaking poses for point cloud segmented rocks. (a) Raw, unfiltered target poses candidates; (b) remaining poses after applying the soft and hard Euler pitch angle filters.}
    \label{fig:target_poses_generator}
\end{figure}

Fig.~\ref{fig:target_poses_generator} illustrates the resulting rock-breaking poses, where Fig.~\ref{fig:target_poses_generator_1} shows the result of the surface-normal analysis to generate target poses for each cluster $\mathcal{C}_\text{rock}$, and Fig.~\ref{fig:target_poses_generator_2} shows the effect of applying the orientation constraints to the same set of generated poses. 

\subsubsection{Rock-breaking poses processor}
\label{subsubsec:breaking_poses_processor}

Pose hierarchization constitutes a critical step in the pipeline, allowing the hammer to perform efficiently. Since the objective of the system is to generate rock-breaking target poses, the rock segmentations obtained from the \textit{point cloud rocks segmentation} and by the \textit{masked point cloud generation} are not fused. Instead, the rock-breaking poses generated by each method are merged into a single set of candidate poses, to which an initial redundancy filter is applied to remove poses that are closer than a predefined threshold to each other.

A scoring system is employed to assign each pose a numerical value that reflects its quality and priority relative to other candidate poses. As in~\cite{cardenas2022automatic}, the scoring system is designed given several criteria based on operation manuals and feedback provided by experienced operators. In addition, our approach incorporates a temporal component that accounts for the pose's reliability based on its temporal stability and its distance from the most recently reached rock-breaking pose, thus avoiding repeated selection of the same pose in successive rock-breaking attempts. 

The weighting of each criterion in the scoring system is controlled by scalar coefficients, allowing different criteria to be emphasized depending on the situation and thus, adapting the impact hammer's high level behavior. Individual criteria can also be disabled by setting their corresponding weight value to zero. The criteria used for rock-breaking pose prioritization are the following:
\begin{itemize}
    \item Euclidean distance to the end-effector ($d_\text{eef}$).
    \item Rock volume ($r_v$).
    \item Temporal consistency ($t_c$).
    \item Distance to the last rock-breaking pose reached ($d_{\text{eef}^{-}}$).
\end{itemize}

Since the decision-making factors exhibit widely varying numerical ranges, each category is first normalized using the corresponding minimum and maximum values among all the current candidate poses, which results in values in the $[0,1]$ range for all criteria.\footnote{In addition, we always add a small scalar to the current distance between the rock-breaking poses and the end-effector position to avoid undefinitions when computing $d_\text{eef}^{-1}$.} The final score for each pose is computed via
\begin{equation}
    \texttt{score} = \lambda_1 d_{\text{eef}}^{-1} + \lambda_2 t_c + \lambda_3 r_v + \lambda_4 d_{\text{eef}^-},
    \label{eq:breaking_poses_prioritization}
\end{equation}
where $\lambda_i >0$, $i\in \{1, 2, 3, 4\}$ represents the weight associated to each criterion.

\begin{figure}
        \centering
        \subfloat[]{\includegraphics[width=0.98\linewidth]{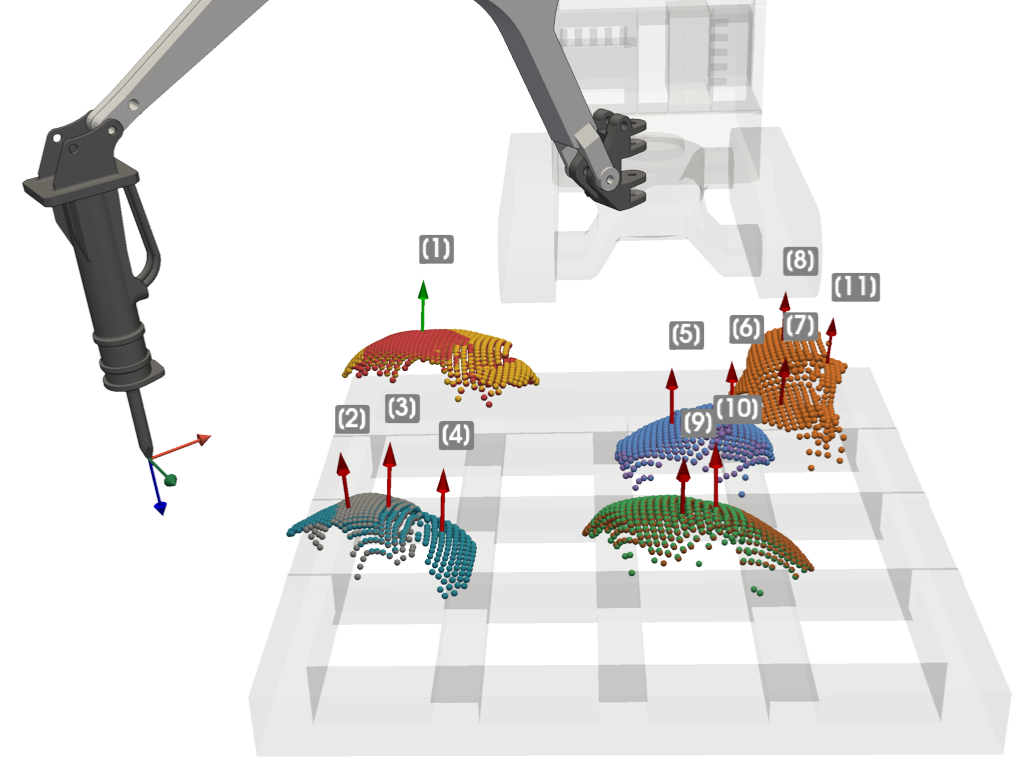}\label{fig:prioritizion_poses_dist_effector_1}} \\
        \subfloat[]{\includegraphics[width=0.98\linewidth]{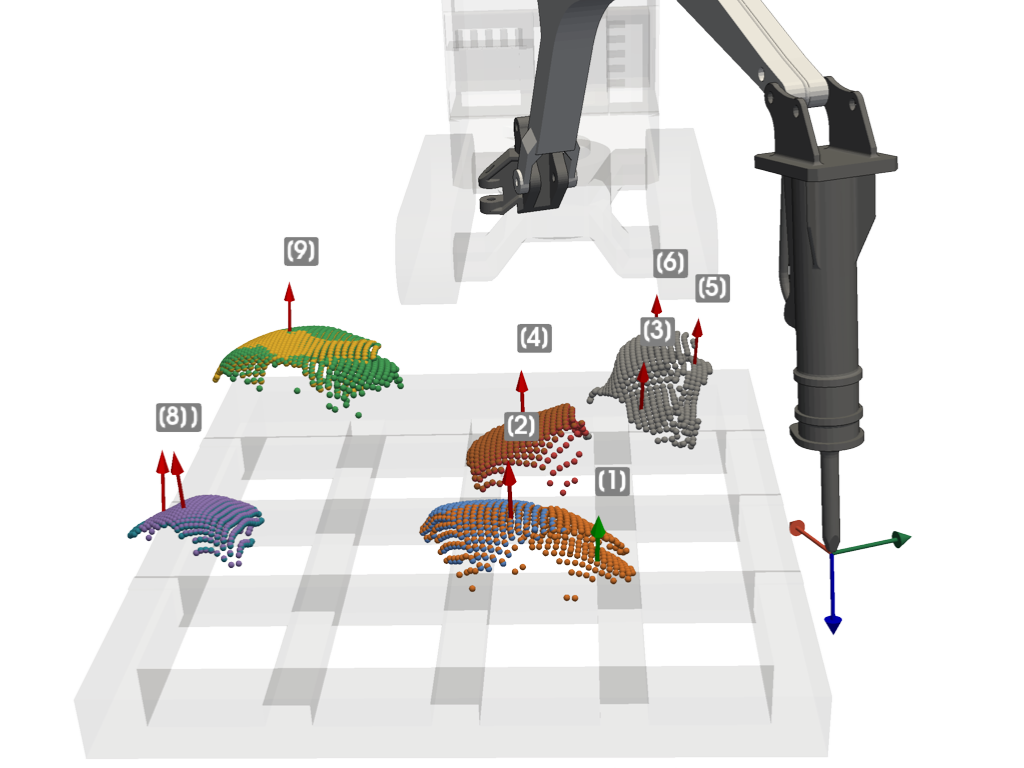}\label{fig:prioritizion_poses_dist_effector_2}} 
        \caption{Prioritization of rock-breaking poses based on their distance to the impact hammer's end-effector. (a) End-effector on the right side of the grizzly. (b) End-effector on the left side of the grizzly.}
    \label{fig:prioritizion_poses_dist_effector}
\end{figure}

The distance between a target pose and the end-effector, $d_\text{eef}$, is used to select target poses that are closer to the end-effector, reducing the time required to reach them between successive breaking attempts. To illustrate the effect of this criterion, Fig.~\ref{fig:prioritizion_poses_dist_effector} shows two grizzly configurations, highlighting how poses are prioritized based on their distance to the impact hammer's end-effector.

The rock volume criterion uses an estimation of the volume of the rock associated to a given target end-effector pose to determine its prioritization, and is based on what was proposed in~\cite{cardenas2022automatic}. Prioritizing breaking the largest rocks helps creating space and avoiding accumulation of material over the steel grate, since larger rocks can impede material flow which may lead to process delays. 

\begin{figure}
        \centering
        \subfloat[]{\includegraphics[width=0.98\linewidth]{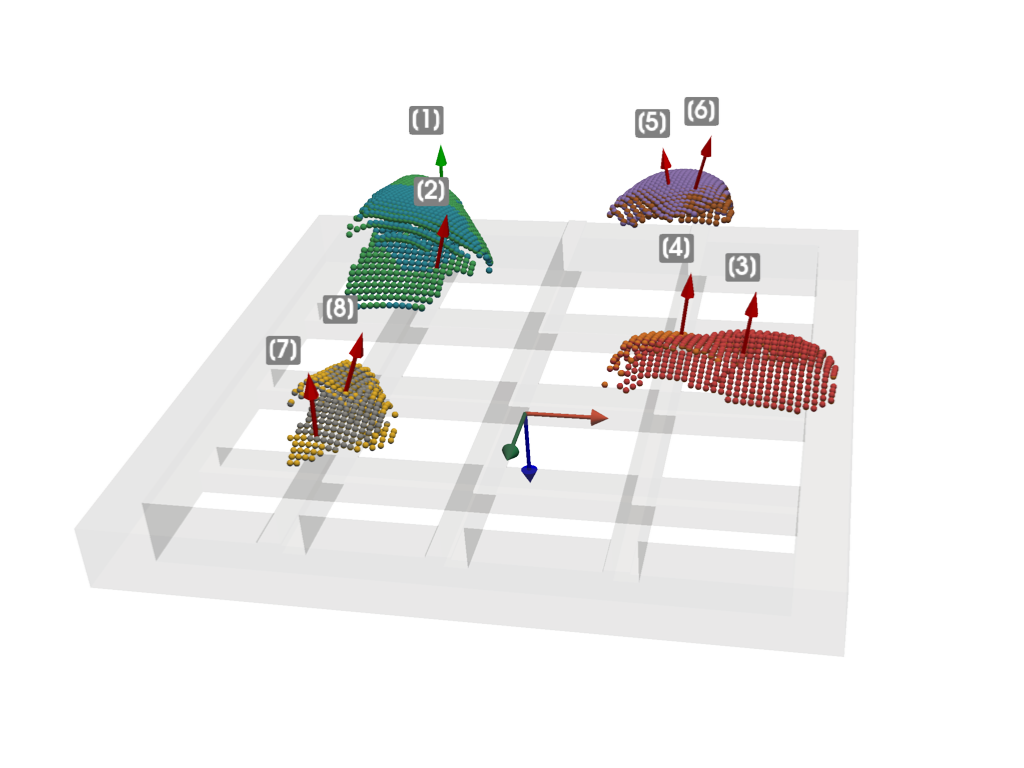}\label{fig:prioritizion_poses_rock_volume_1}} \\
        \subfloat[]{\includegraphics[width=0.98\linewidth]{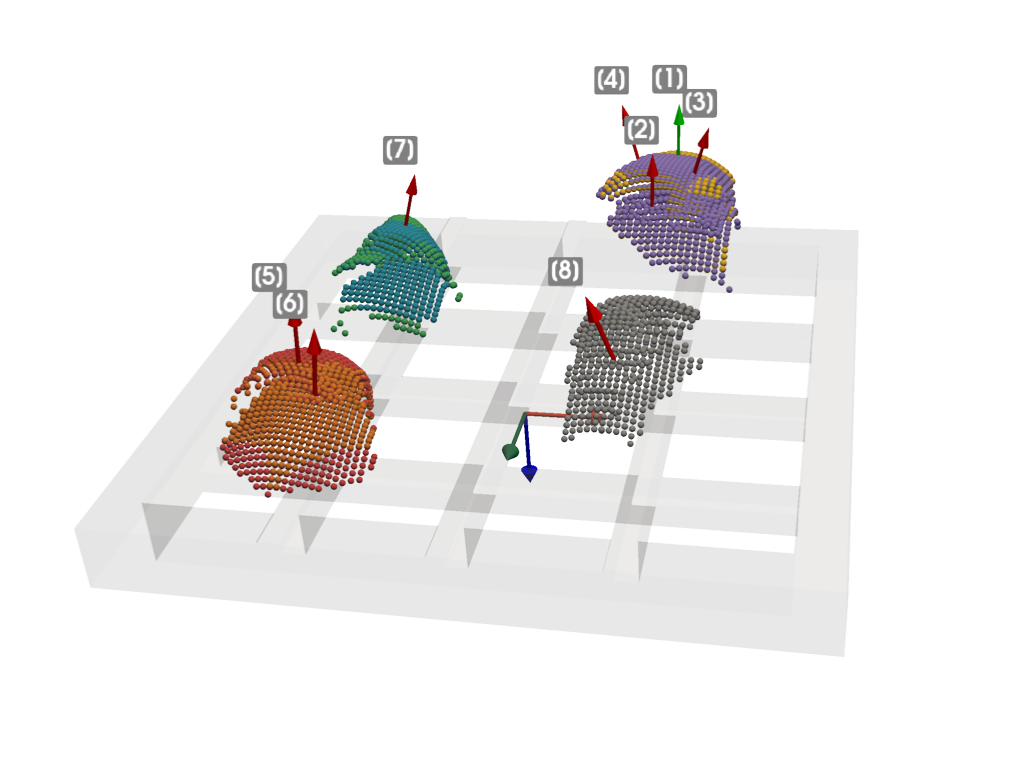}\label{fig:prioritizion_poses_rock_volume_2}} 
        \caption{Rock-breaking poses prioritized according to the size of the rock to which they belong; (a) and (b) show two different configurations for rocks over the steel grate.}
    \label{fig:prioritizion_poses_rock_volume}
\end{figure}

The volume of a given rock, $r_v$, is estimated at the segmentation stage, by computing the oriented bounding box of the points corresponding to the rock itself, and $r_v$ being obtained as (eight times) the product of the bounding box extents. When generating the rock-breaking poses, each pose is associated with the rock to which it belongs, and the corresponding volume value is stored for use in the prioritization process. Fig.~\ref{fig:prioritizion_poses_rock_volume} illustrates the effect of this criterion in practice, showing two rock configurations and how the preferred pose is consistently associated with the largest rock. It is important to note that, since the rocks segmented by the ``\textit{Point cloud rock segmentation}'' and ``\textit{Masked point cloud generation}'' methods are not merged, the same rock may have different sizes under each segmentation approach (assuming both methods successfully segment it). Consequently, poses belonging to the same real rock can be prioritized as if they belong to different rocks.

The temporal consistency criterion, $t_c$, is designed to provide a stability measure across successive frames to the computed poses. If a given rock-breaking pose appears in the set of generated candidates at time $t$, its associated temporal weight is incremented by one; conversely, if it does not appear, its value is decremented by one. This criterion takes values in the $[0, t_\text{max}]$ range, where $0$ indicates that the breaking pose has appeared for the first time, and $t_\text{max}$ represents the highest possible stability level. For our experimental setup, the value of $t_\text{max}$ is set to $15$. This value was selected based on the pose generation frequency of $\approx10$~Hz; thus, a pose that remains as candidate for about $1.5$ seconds is considered stable.

\begin{figure}
        \centering
        \subfloat[]{\includegraphics[width=0.98\linewidth]{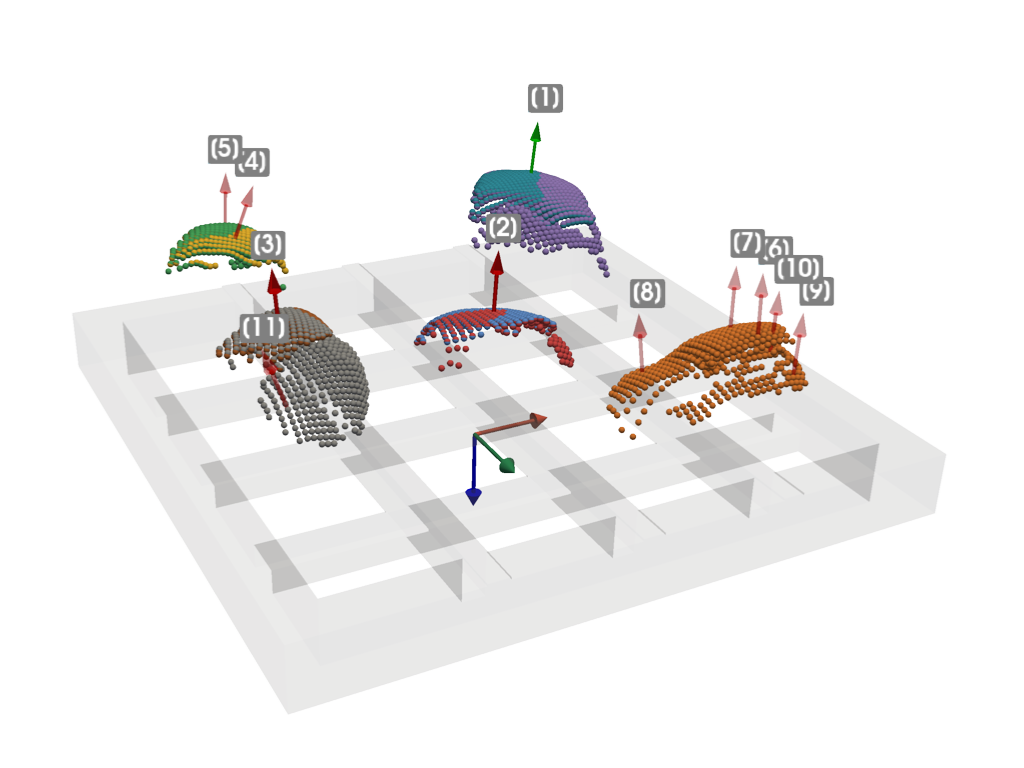}\label{fig:prioritizion_poses_temporal_consistency_1}} \\
        \subfloat[]{\includegraphics[width=0.98\linewidth]{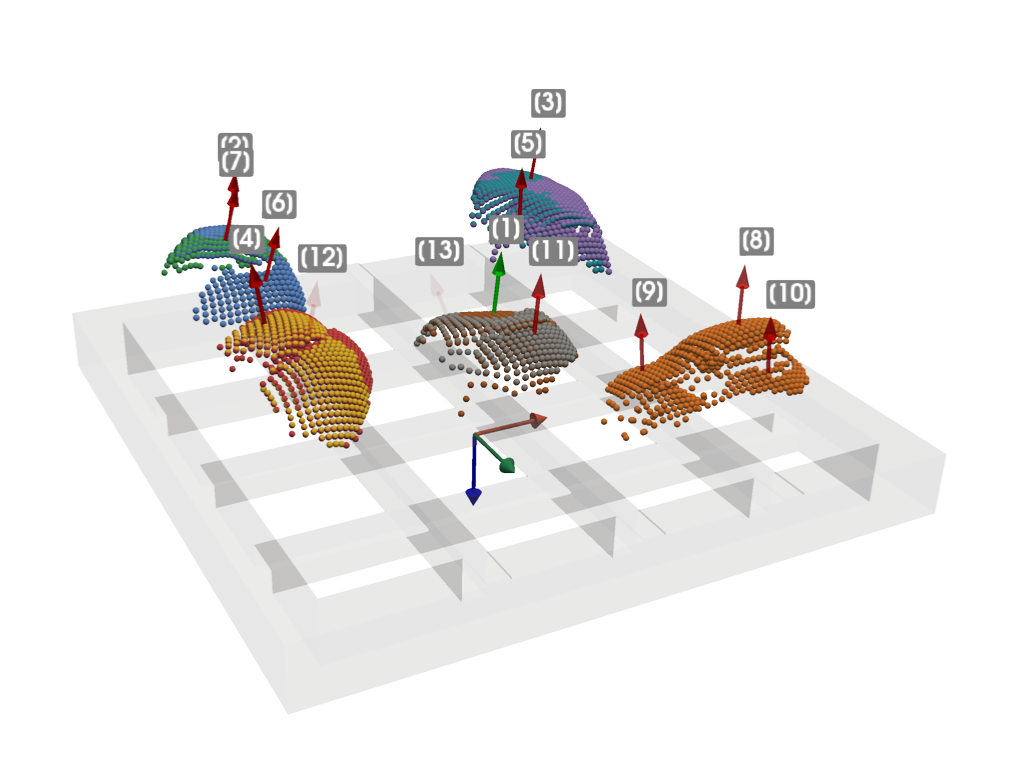}\label{fig:prioritizion_poses_temporal_consistency_2}} 
        \caption{Rock-breaking poses prioritized according to their temporal consistency. (a) Prioritized poses at time instant $t=t_1$. (b) Prioritized poses at time instant $t=t_2$, such that $t_2>t_1$.}
    \label{fig:prioritizion_poses_temporal_consistency}
\end{figure}

Fig.~\ref{fig:prioritizion_poses_temporal_consistency} illustrates the effect of this criterion using shaded arrows: newly appearing poses are depicted with nearly transparent arrows, while more temporally consistent poses are shown with increased opacity as they approach the maximum value. Fig.~\ref{fig:prioritizion_poses_temporal_consistency_1} shows the generated poses at time instant $t=t_1$, when the poses have not yet stabilized. Fig.~\ref{fig:prioritizion_poses_temporal_consistency_2} shows the poses at time instant $t=t_2>t_1$, where they are more stable and most exhibit high temporal consistency.

During hammer operation, multiple cycles of reaching breaking poses and striking rocks are performed. To avoid repeatedly striking the same location on the grizzly, or even the same pose when a rock is not fractured on the first attempt, a final criterion is introduced that penalizes poses based on their Euclidean distance to the last selected breaking pose. This distance is denoted as $d_{\text{eef}^-}$.

\begin{figure}
        \centering
        \subfloat[]{\includegraphics[width=0.98\linewidth]{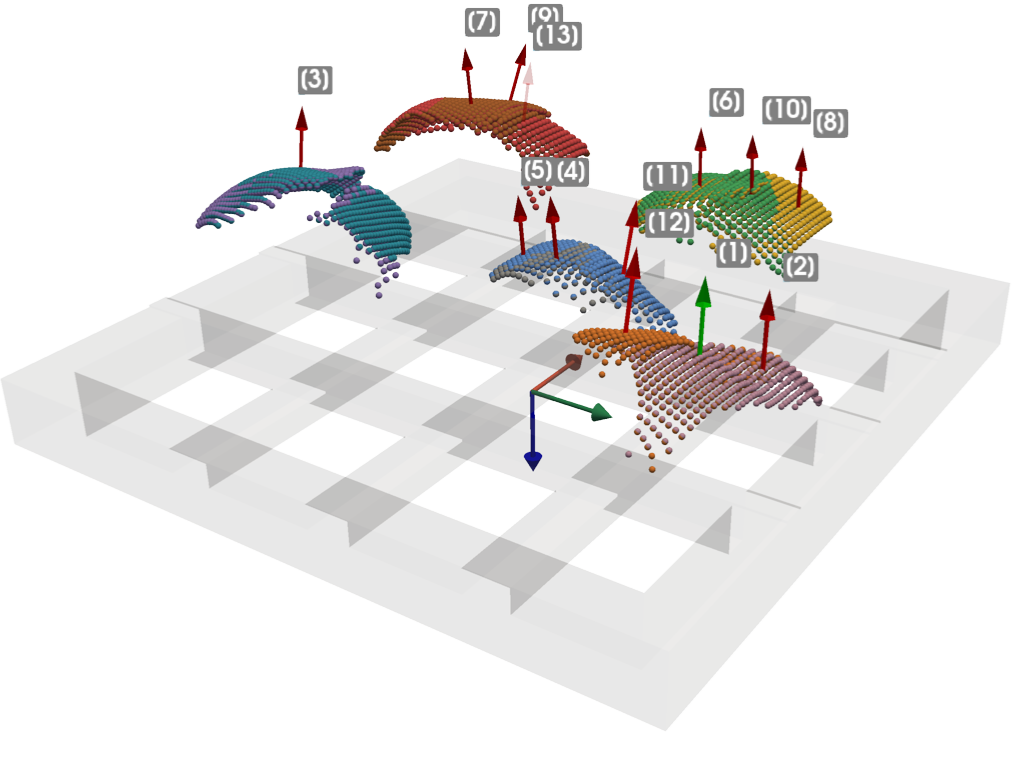}\label{fig:prioritizion_poses_last_pose_1}} \\
        \subfloat[]{\includegraphics[width=0.98\linewidth]{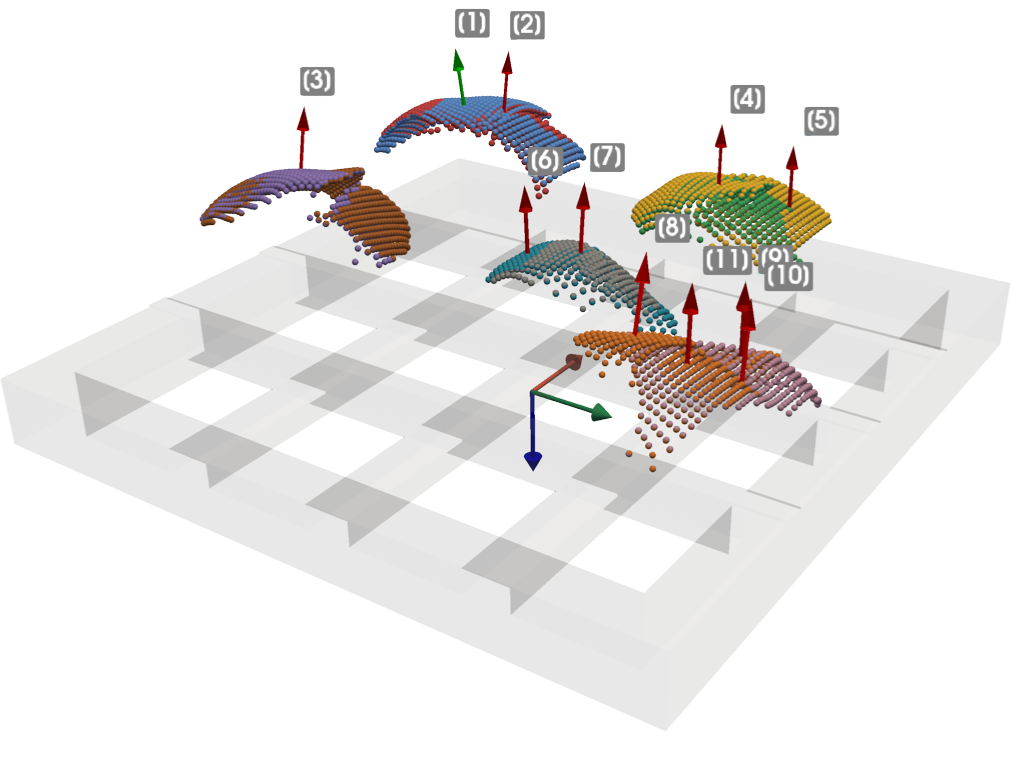}\label{fig:prioritizion_poses_last_pose_2}} 
        \caption{Rock-breaking poses prioritized based on their distance from the last pose struck by the hammer. (a) Prioritization in the absence of a previously selected pose. (b) Prioritization after selection of the first pose in (a).}
    \label{fig:prioritizion_poses_last_pose}
\end{figure}

Fig.~\ref{fig:prioritizion_poses_last_pose} illustrates the effect of this criterion. In Fig.~\ref{fig:prioritizion_poses_last_pose_1}, pose prioritization is shown in the absence of a previously selected pose; consequently, the ordering is effectively arbitrary and reflects only the generation sequence. In Fig.~\ref{fig:prioritizion_poses_last_pose_2}, after one operation cycle has been completed, poses located farther from the last selected pose are preferentially ranked in the subsequent cycle.

When all criteria are applied jointly, the distance-based penalization promotes significant spatial variation in the selected poses, thereby improving coverage of the working area. The values assigned to all parameters introduced in this section and used in the experiments of Sect.~\ref{sec:experiments} are listed in App.~\ref{app:breaking_poses}.

\section{Experimental results}
\label{sec:experiments}

\subsection{Experimental setup and implementation details}
\label{subsec:implementation_details}

The system is validated entirely using the experimental setup shown in Fig.~\ref{fig:experimental_setup}, without relying on simulations. The processing pipeline runs on a Jetson AGX Orin equipped with a capture card that enables the use of two ZED~X stereo cameras; their initialization and configuration parameters can be seen in App.~\ref{app:sensor_data_acquisition}. At the implementation level, the system relies primarily on JAX~\citep{jax2018github} for high-performance numerical computing, on Open3D~\citep{zhou2018open3d} for handling images and 3D data, on PyTorch~\citep{paszke2019pytorch} for running artificial neural network models, and on the ROS framework~\citep{quigley2009ros} to facilitate data exchange across all stages of the pipeline. App.~\ref{app:software_architecture} shows a detailed overview of the proposed pipeline's software architecture. Regarding the RTMDet model used for rock segmentation (see Sect.~\ref{subsubsubsec:image_rocks_segmentation}), App.~\ref{app:rtmdet_settings} summarizes the training procedure and provides details regarding the dataset and the parameters used for this purpose.

\begin{figure*}
    \centering
    \subfloat[]{\includegraphics[width=0.19\linewidth]{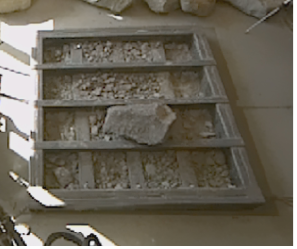}\label{fig:perception_experiments_a}} 
    \subfloat[]{\includegraphics[width=0.19\linewidth]{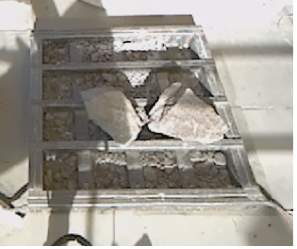}\label{fig:perception_experiments_b}} 
    \subfloat[]{\includegraphics[width=0.19\linewidth]{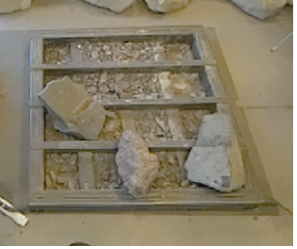}\label{fig:perception_experiments_c}} 
    \subfloat[]{\includegraphics[width=0.19\linewidth]{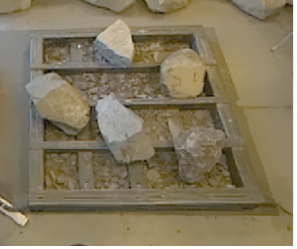}\label{fig:perception_experiments_d}} 
    \subfloat[]{\includegraphics[width=0.19\linewidth]{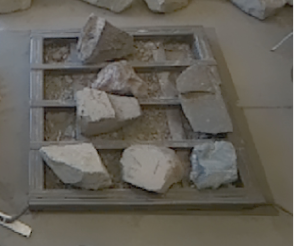}\label{fig:perception_experiments_e}}
    \caption{Examples of experimental configurations used for rock segmentation and rock-breaking pose generation. (a) Single rock; (b) two rocks; (c) three rocks; (d) five rocks; (e) seven rocks.}
    \label{fig:perception_experiments}
\end{figure*}

\subsection{Rock segmentation and rock-breaking pose generation}
\label{subsubsec:rock_segmentation_and_rock_breaking_pose_generation}

To evaluate the performance of the proposed rock-segmentation algorithms and also to characterize the rock-breaking pose generation method, rocks that would not pass through the grizzly are placed at multiple different positions over the steel grate. For all experiments, five rock quantities are utilized: $1$, $2$, $3$, $5$ and $7$ rocks. For each quantity, $30$ distinct configurations are evaluated by running the perception pipeline and saving the segmented rocks and the corresponding generated rock-breaking poses. Each configuration corresponds to a unique spatial arrangement of the rocks, with variations in their positions and orientations to ensure diversity in surface geometry and coverage across the entire steel grate area. Additionally, illumination conditions and the specific rocks used were varied across configurations to further increase variability and assess the robustness of the method. Several examples of configurations for the aforementioned experiments are illustrated in Fig.~\ref{fig:perception_experiments}. 

To evaluate rock segmentation, we consider the following performance metrics:
\begin{itemize}
    \item Avg. Segm. count: average number of segmented rocks.
    \item Segm. MAE: mean absolute error of the number of segmented rocks with respect to ground-truth counts.
    \item Cons. segm. count: percentage of frames across a given trial for which the ground-truth number of rocks over the grizzly is consistent with the number of segmented rocks.
\end{itemize}

For the generation of rock-breaking poses, on the other hand, the following performance metrics are computed:
\begin{itemize}
    \item Avg. pose count: average number of rock-breaking poses generated.
    \item Avg. $|\theta_\text{pitch}|$: average absolute pitch angles of the generated poses, measured with respect to the $\hat{z}$ axis of the impact hammer's base link.
\end{itemize}

\begin{table*}[h]
    \sisetup{separate-uncertainty=true, table-align-uncertainty=true, retain-zero-uncertainty=true}
    \caption{Performance metrics obtained for the rock segmentation and the rock-breaking pose generation stages of the perception pipeline. }
    \label{tab:seg_gen_results}
    \centering
    \begin{tabular*}{\linewidth}{@{\extracolsep{\fill}}c@{\extracolsep{\fill}}l@{\extracolsep{\fill}}S[table-format=2.2(3)]@{\extracolsep{\fill}}S[table-format=2.2(3)] @{\extracolsep{\fill}}c @{\extracolsep{\fill}}S[table-format=2.2(3)]@{\extracolsep{\fill}}S[table-format=2.2(3)]@{\extracolsep{\fill}}}
    \toprule
    \multirow{3}{*}{\makecell{\# Rocks}} & 
    \multirow{3}{*}{\makecell{Input\\ data}} & 
    \multicolumn{3}{c}{Rock segmentation} & \multicolumn{2}{c}{Rock-breaking pose generation} \\ \cmidrule{3-5} \cmidrule{6-7}
     & 
     & 
    {\makecell{Avg. segm. count}} & 
    {\makecell{Avg. segm. MAE}} & 
    \makecell{Cons. segm. count [\%]} &
    {\makecell{Avg. pose count}}   &  
    {\makecell{Avg. $|\theta_\text{pitch}|$ [deg]}}  \\
    \midrule
    \multirow{3}{*}{\makecell[l]{1}}   
         & \texttt{pcl} & 1.01 \pm 0.09 & 0.01 \pm 0.09 & 99.33  & 1.70 \pm 0.80 & 9.01 \pm 5.89 \\
         & \texttt{img} & 1.00 \pm 0.00 & 0.00 \pm 0.00 & 100.00 & 2.62 \pm 1.34 & 9.67 \pm 5.93 \\ 
         & \texttt{pcl+img} & {--} & {--} & {--} & 3.16  \pm 1.68 & 9.56 \pm 5.93 \\ \midrule
    \multirow{3}{*}{\makecell[l]{2}}   
         & \texttt{pcl} & 1.73 \pm 0.44 & 0.27 \pm 0.44 & 73.15  & 2.44 \pm 1.23 & 8.44 \pm 5.53  \\
         & \texttt{img} & 1.76 \pm 0.42 & 0.24 \pm 0.42 & 76.40  & 3.55 \pm 1.78 & 9.65 \pm 5.85  \\
         & \texttt{pcl+img} & {--} & {--} & {--} & 4.96  \pm 2.18 & 9.22 \pm 5.69 \\ \midrule
    \multirow{3}{*}{\makecell[l]{3}}  
         & \texttt{pcl} & 2.78 \pm 0.55 & 0.28 \pm 0.52 & 74.93 & 4.04 \pm 1.29 & 9.64 \pm 5.79 \\
         & \texttt{img} & 2.83 \pm 0.54 & 0.23 \pm 0.51 & 80.80 & 5.68 \pm 2.42 & 9.86 \pm 5.92 \\
         & \texttt{pcl+img} & {--} & {--} & {--} & 8.52  \pm 3.27 & 9.77 \pm 5.88 \\ \midrule
    \multirow{3}{*}{\makecell[l]{5}}  
         & \texttt{pcl} & 4.48 \pm 0.82 & 0.52 \pm 0.82 & 65.67 & 6.74 \pm 1.95 & 9.09 \pm 5.60 \\
         & \texttt{img} & 4.97 \pm 0.38 & 0.10 \pm 0.37 & 92.55 &  9.40 \pm 2.84 & 9.38 \pm 5.58 \\
         & \texttt{pcl+img} & {--} & {--} & {--} & 14.30 \pm 4.29 & 9.58 \pm 5.61 \\ \midrule
    \multirow{3}{*}{\makecell[l]{7}}  
         & \texttt{pcl} & 5.31 \pm 1.40 & 1.70 \pm 1.41 & 15.32 &  9.34  \pm 2.33  & 9.24 \pm 5.68 \\
         & \texttt{img} & 6.81 \pm 0.61 & 0.27 \pm 0.58 & 80.32 & 12.39 \pm 4.07  & 9.65 \pm 5.88 \\
         & \texttt{pcl+img} & {--} & {--} & {--} & 18.20 \pm 5.33 & 9.98 \pm 6.07 \\ \bottomrule
    \end{tabular*}
    
\end{table*}

As previously stated, we positioned rocks over the grizzly of the scaled real environment. For each configuration ($30$ per rock quantity), $200$ perception pipeline steps were evaluated (corresponding to approximately $20$ seconds of execution time per configuration). The obtained results are presented in Table~\ref{tab:seg_gen_results} for the rock segmentation and the rock-breaking poses generation evaluation. The results are reported as a function of the number of rocks in the steel grate and the input data that is processed, where \texttt{pcl} stands for point cloud data, \texttt{img} for image data, and \texttt{pcl+img} to both. Note that the latter input data category only matters for the target rock-breaking pose generation stage, where poses obtained using rock segmentations produced using both images and point clouds (independently) are merged according to the procedure described in Sect.~\ref{subsubsec:breaking_poses_processor}.

The rock segmentation results show that the image-based method produces the same number of rock segmentation instances as the ground-truth rock counts more often than the point cloud–based segmentation approach. This trend is particularly evident in the seven-rocks experiments, where the performance gap between the two methods is most pronounced. This suggests that the image-based segmentation method more reliably individualizes rocks on the grizzly.

The lower performance of the point cloud–based method can be attributed to its reliance on geometric segmentation. When rocks are in close proximity, the algorithm tends to merge adjacent objects into a single segmentation instance, leading to performance degradation as scene complexity increases. In fact, as the number of rocks grows, the metric measuring consistent rock segmentation counts (Cons. segm. count) steadily decreases for the point-cloud based rock segmentation. 

We must recall, however, that one of the primary objectives of the proposed perception pipeline is to generate feasible rock-breaking poses, and the generation of these poses relies primarily in surface normal analysis over rocks, and not on their proper individualization. From this perspective, the point cloud-based method provides sufficient information to distinguish rock material from non-rock material over the grizzly; therefore, DBSCAN remains a suitable algorithm for rock segmentation in 3D space within the context of this application, regardless of being unable to always properly distinguish between overlapping rocks.

The rock-breaking pose generation results, on the other hand, present a clear and consistent trend: as the number of rocks on the grizzly increases, the number of generated poses also increases (approximately linearly). On average, the system produces between $2$ and $3$ poses per rock, indicating that the pipeline scales candidate rock-breaking poses proportionally with scene complexity. Regarding pose orientation, the absolute pitch angle with respect to the $\hat{z}$ axis remains highly consistent across all configurations, with mean values close to $9.5$~deg. and similar standard deviations. This suggests that the pose generation strategy enforces a maximum pitch angle, independent of the number of rocks in the scene. This property is largely conditioned by the thresholds imposed during the pose generation process. Specifically, constraints on allowable pitch angles bound the solution space, leading to low variability across different scenarios, which explains the limited dispersion observed in the orientation metrics.

We must note that, unlike some previous approaches, which rely on accurately identifying individual rock instances to place a single target pose at each centroid (e.g.,~\cite{lampinen2021autonomous}), as previously stated, the proposed method generates poses directly from the local surface geometry. Consequently, valid rock-breaking poses can still be produced even when multiple rocks are merged into a single cluster or when the point cloud segmentation is otherwise imperfect.

The main consequence of inaccurate rock segmentation, however, appears during the pose prioritization stage rather than pose generation itself. In particular, segmentation errors may affect the estimation of rock-level attributes, such as volume, which are used to rank the generated poses. Therefore, while poor segmentation can influence the order in which rocks are processed, it has a considerably smaller impact on the validity of the generated rock-breaking poses.

\subsection{Quality assessment of rock-breaking poses}
\label{subsubsec:quality_assessment_of_rock_breaking_poses}

\subsubsection{Testing the poses in the real-world}
\label{subsubsec:real_world_testing}

To realistically assess the quality of the generated rock-breaking poses, the perception pipeline is coupled with a control system (described in~\cite{leiva2026data}) that allows the impact hammer's end-effector to reach a given target pose (i.e., a \emph{reaching} controller), and high-level behaviors (implemented via state-machines) that allow looping over different poses according to their prioritization, and emulate a rock-breaking stage by pressing the end-effector's tip against the rocks associated with the generated poses. 

The experimental evaluation proceeds as follows:

\begin{enumerate}
    \item A fixed pose for the impact hammer's end-effector, placed above the grizzly, is set as a target for the reaching controller. This target pose allows the machine to get to a ``\emph{home}'' configuration that allows the perception system to generate poses without much occlusion from the impact hammer itself.
    \item A set of target rock-breaking poses generated by the perception pipeline is fixed.\footnote{While during normal operation the perception system constantly update poses for all the rocks over the grizzly, in this evaluation a set of generated poses is fixed and the impact hammer has to sequentially execute each pose according to the established (and also fixed) priority ranking. This is done because rocks remain on the steel grate after the simulation of their potential breakage, therefore, without fixing a set of generated poses, not all of them (and their associated rocks) would be tested, limiting the variability across experiments for a given configuration of rocks.}
    \item Then, iteratively, until all poses are covered:
    \begin{enumerate}
        \item  The highest priority pose is set as a target for the reaching controller. If the target pose cannot be reached by the controller, it is removed, the priority of the remaining poses is updated, and the resulting highest priority pose is set as the new target. 
        \item Once a target pose is reached, the impact hammer is actuated (through a constant current setpoint on the impact hammer's boom actuator) so that its end-effector tip presses against the rock to simulate the striking process, which requires exerting constant pressure over rocks to avoid blank firing. If contact between the machine and the rock surface is detected, then this pressure is maintained for $5$ seconds (through another state machine), during which it is recorded whether the hammer's chisel maintains stable contact over the rock's surface, or slips.
        \item The impact hammer returns to its \emph{home} configuration by setting the corresponding target pose to the reaching controller.
        \item The last visited rock-breaking pose is removed and the remaining poses' priorities are updated.
    \end{enumerate}
   
\end{enumerate}

Given the methodology described above, the quality of a given target pose is assessed depending on whether the impact hammer's end-effector tip (the chisel) reliably remains in contact with the pose's associated rock while exerting pressure over it. In this stage of the experimental evaluation (the step b) above), three possible outcomes were observed:

\begin{itemize}
    \item The chisel did not make contact with the rock, typically because the selected breaking pose was too close to an edge.
    \item The chisel made contact with the rock but slipped over its surface shortly afterwards.
    \item The chisel successfully established stable contact with the rock.
\end{itemize}

In the above, the chisel-rock contact was determined by comparing the current impact hammer's configuration and a prediction of said configuration by means of a learned dynamical model of the machine (also described in~\cite{leiva2026data}). By detecting a divergence between the prediction of the impact hammer's response to a history of control commands with its actual configuration, we can reliably detect contacts, as the learned dynamical model is unaware of exteroceptive information and collisions with the environment, thus, while the real hammer would remain still when exerting pressure over a rock, the learned dynamical model would predict movement as if no rock would be present.

We must note that chisel-rock contact is detected when the predictions of the dynamic model and the true impact hammer configuration are divergent beyond a predefined threshold. Moreover, the learned dynamical model of the impact hammer takes as input a $1.5$~s window of previous commands and encoder measurements (sampled at $20$~Hz) to make predictions for the current time step. This implies that simply touching the rock is not detected as contact under this setting. Moreover, the state machine responsible for exerting pressure over the rocks after the end-effector reaches a target pose, only acts upon the boom actuator if contact is detected as described above.

Given the aforementioned, we define the following performance metrics:
\begin{itemize}

    \item Success rate: quantifies the proportion of poses in which stable contact was successfully maintained and a simulated strike (via sustained pressure exertion) was executed.
    \item Post-contact slip: represents the proportion of poses in which the hammer exerted pressure over the rock but slipped while doing so. This is quantified by monitoring whether the hammer end-effector moved outside a sphere of radius of $20$~cm, centered at the target pose, while pressing the rock.
    \item No contact detection rate: represents the proportion of poses for which pressure was not exerted over a rock because contact between the chisel and the rock's surface was not detected.
\end{itemize}

Qualitatively, the ``no contact detection rate'' can be decomposed into two different rates:
\begin{itemize}
    \item True no contact: represents the proportion of experiments in which no noticeable/meaningful contact between the chisel and the rock's surface was observed.
    \item Pre-contact slip: represents the proportion of attempts in which the chisel touched the rock, but failed to remain in contact with its surface long enough so contact would be detected. This often happened due to the rock on the grizzly being in an unstable pose, and the chisel moving said rock while pushing it, allowing for free end-effector movement.
\end{itemize}

Note that it is not possible to distinguish between ``true no contact'' and ``pre-contact slip'' quantitatively given our experimental setup, as the state machine only detects the absence of stable contact between the chisel and the rock surface, without identifying the underlying cause. Thus, both ``true no contact'' and ``pre-contact slip'', as qualitative metrics, can only be determined based on operator observations recorded during the experiments. 

The evaluation of our proposed pose generation method is conducted across $11$ different configurations, each of them with $6$ rocks over the steel grate. The above resulted in a total of $117$ rock-breaking poses that were set as targets to the impact hammer (after filtering those that were deemed invalid due to rock's shifting positions, and the set of rock-breaking target poses for a given configuration being fixed). In addition, and for all experiments, a $10$~cm offset was applied along the steel grate's $z$-axis to each pose's position, so as to place the impact hammer's chisel (via the reaching controller) above the rock surface prior to the exerting pressure over it. Moreover, to consider the reaching task as successfully achieved by the controller, we set a distance threshold of $0.08$~m between the end-effector's tip and the target pose's position, and a threshold of $0.08$~rad to compare their pitch angles. Under this criterion, $6$ out of the $117$ poses were not successfully reached when evaluating our pose generation method. These instances were excluded from the evaluation metrics, as the control system operates independently of the perception pipeline; consequently, the results were computed using the remaining $111$ rock-breaking poses. Finally, besides the quantitative metrics recorded leveraging the state machine during the conduction of experiments, a human observer also manually annotated the result of each trial to independently construct qualitative metrics via a success rate, a post- and pre-contact slip rate, and a true no contact rate (note that the no contact rate in this case is computed using the sum of true no contacts and pre-contact slips).

The results for these experiments are presented in Table~\ref{tab:quality_assessment_of_rock_breaking_poses_results}. In general, we observe that the quantitative and qualitative assessments of the experiments conducted produce similar results, with discrepancies arising only in unsuccessful trials, where there are disagreements only regarding contact detection. The success rate metric indicates that in approximately $72\%$ of cases, the impact hammer is able to constantly press against a rock while remaining in a fixed configuration. While this does not imply that actuating the impact hammer's end-effector in these situations would result in rock fractures (as these are not characterized in this work), it is an indicator of such striking attempts being viable and potentially not ending abruptly due to post-contact slips. Indeed, the low post-contact slip metric observed indicates that whenever contact is detected (via a learned dynamical model of the machine, as described above), it is highly unlikely that the impact hammer will slip afterwards if only pressing against the rock. We attribute this result to poses being generated considering rock surface geometry explicitly.

Regarding trials where no contact was detected, we observe two different cases that, as explained previously, can only be explained qualitatively given our experimental settings. True no contact is observed mostly when the target rock-breaking pose is near  the edge of a rock. Since there are Euclidean distance and angular difference thresholds to consider a given reaching trial as successful, there is no strict requirement for placing the end-effector tip exactly where the pose is generated. While this is partially alleviated by the fact that pose position and orientation depend on clustering/averaging operations, whenever poses are generated near edges, there is a chance of placing the end-effector in a pose such that when actuating the machine's boom afterwards, it misses the rock surface. We argue that this can be improved by relying on a more precise reaching controller.

Finally, regarding the pre-contact slip metric, we must note that the perception system we have cannot fully characterize the rocks' geometry, as it only sees them from the ceiling. In some instances, rocks can be placed in unstable configurations on top of the steel grate. We observed that many pre-contact slips occurred due to poses being generated in places where exerting pressure over the rocks would force them to move with the end-effector due to their configuration being unstable, which resulted in no contact detection.

\begin{table}[]
    \caption{Quality assessment of rock-breaking poses.}
    \label{tab:quality_assessment_of_rock_breaking_poses_results}
    \centering
    \begin{tabular*}{\linewidth}{@{\extracolsep{\fill}}l@{\extracolsep{\fill}}S[table-format=2.2]@{\extracolsep{\fill}}S[table-format=2.2]@{\extracolsep{\fill}}}
    \toprule
         \multirow{2}{*}{Performance metrics [\%]} & \multicolumn{2}{c}{{Assessment}}  \\ \cmidrule{2-3}
         & {Quantitative} & {Qualitative} \\ \midrule 
         Success rate &   72.07 & 72.07 \\
         Post-contact slip &  0.00 & 3.60  \\
         No contact detected &  27.93 & 24.32 \\
         \quad $\hookrightarrow$ True no contact &{--} & 10.81 \\
         \quad $\hookrightarrow$ Pre-contact slip &{--} & 13.51  \\
         \bottomrule
    \end{tabular*}
\end{table}

\begin{figure*}
    \centering
    \small
    \begin{tabular*}{\linewidth}{@{\extracolsep{\fill}}c@{\hspace{-10pt}}c@{\extracolsep{\fill}}c@{\extracolsep{\fill}}c@{\extracolsep{\fill}}c@{\extracolsep{\fill}}}
        \vspace{0.1cm}
         \rotatebox{90}{\makecell{Rocks}} & 
         \raisebox{-0.3\height}{\includegraphics[width=0.111\linewidth]{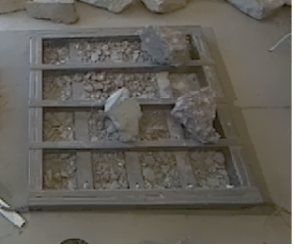}} \raisebox{-0.3\height}{\includegraphics[width=0.111\linewidth]{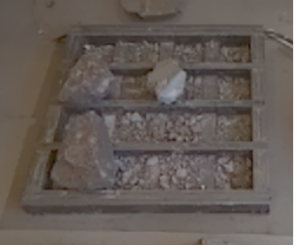}} & 
         \raisebox{-0.3\height}{\includegraphics[width=0.111\linewidth]{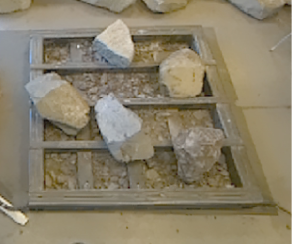}}
         \raisebox{-0.3\height}{\includegraphics[width=0.111\linewidth]{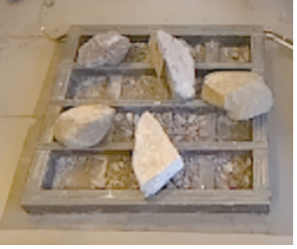}}& 
          \raisebox{-0.3\height}{\includegraphics[width=0.111\linewidth]{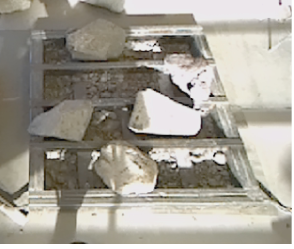}}
         \raisebox{-0.3\height}{\includegraphics[width=0.111\linewidth]{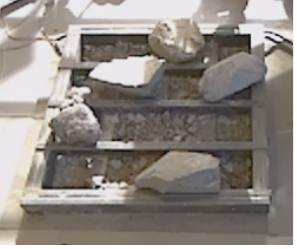}}& 
         \raisebox{-0.3\height}{\includegraphics[width=0.111\linewidth]{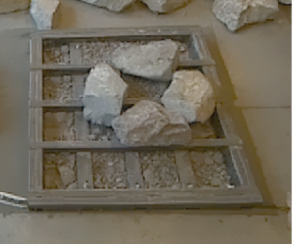}}
         \raisebox{-0.3\height}{\includegraphics[width=0.111\linewidth]{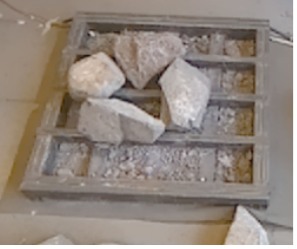}} \\ \vspace{0.1cm}
         \rotatebox{90}{\makecell{Centroid \\ (\texttt{img})}} & 
         \raisebox{-0.3\height}{\includegraphics[width=0.25\linewidth]{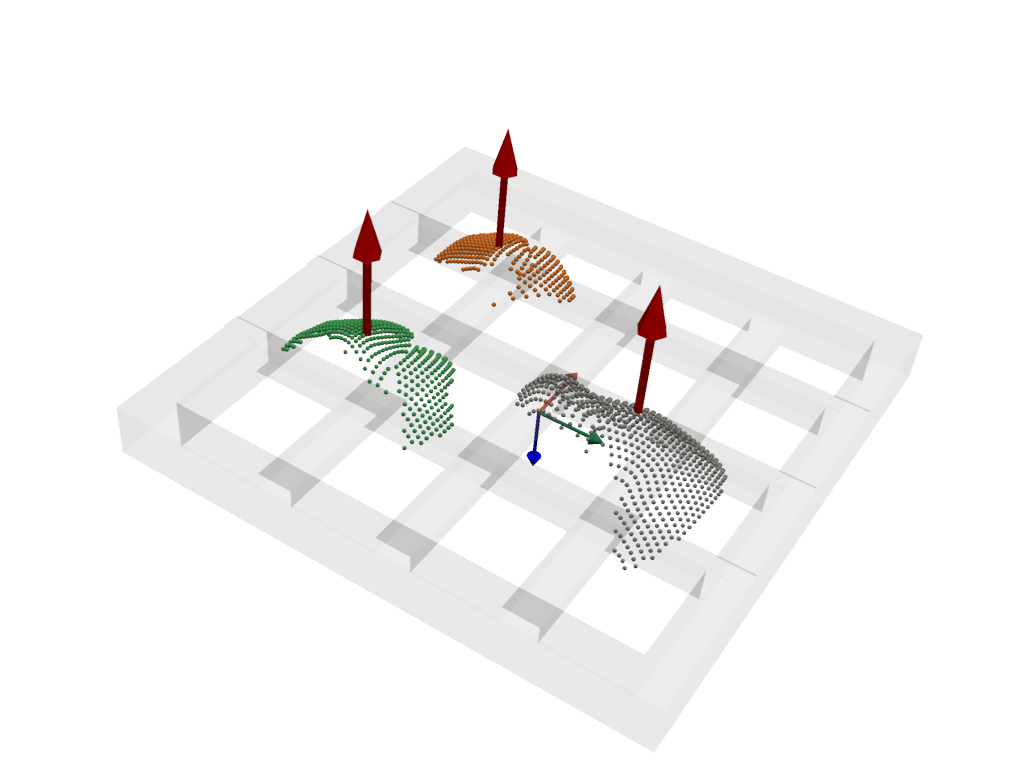}} & 
         \raisebox{-0.3\height}{\includegraphics[width=0.25\linewidth]{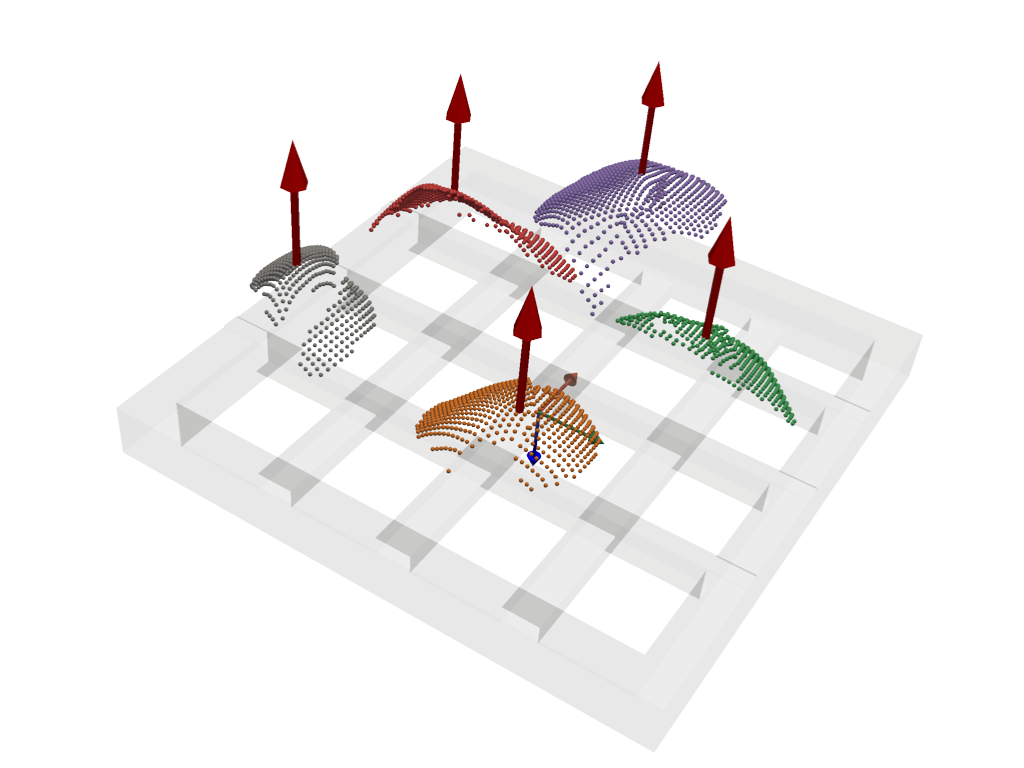}} & 
         \raisebox{-0.3\height}{\includegraphics[width=0.25\linewidth]{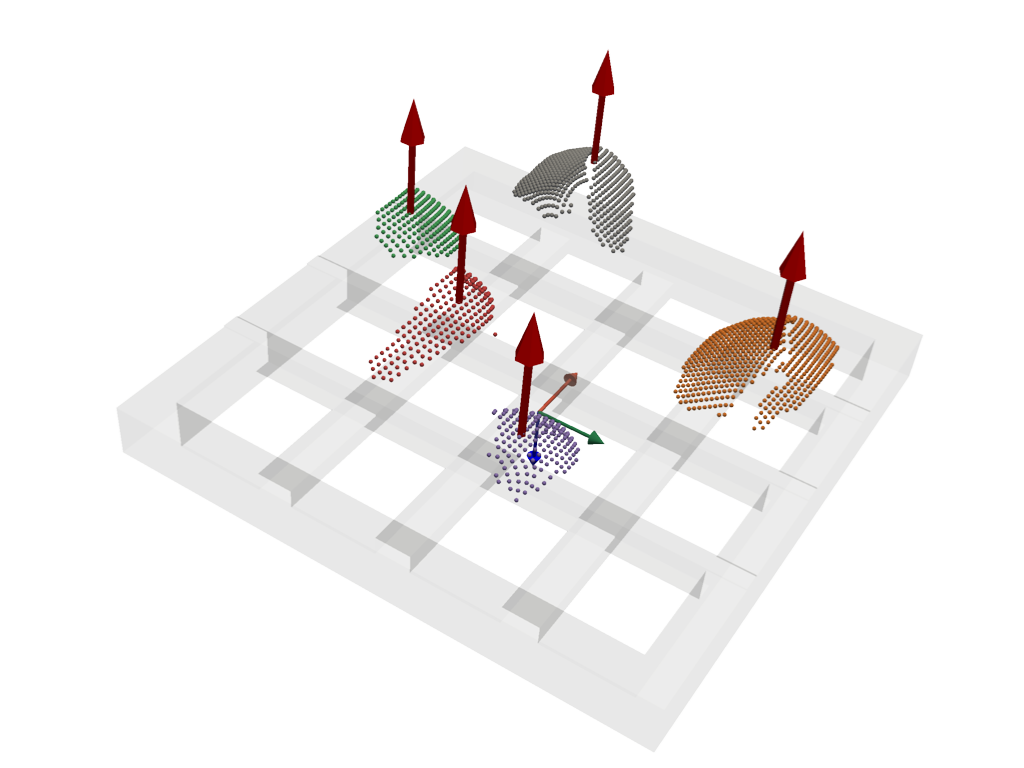}} & 
         \raisebox{-0.3\height}{\includegraphics[width=0.25\linewidth]{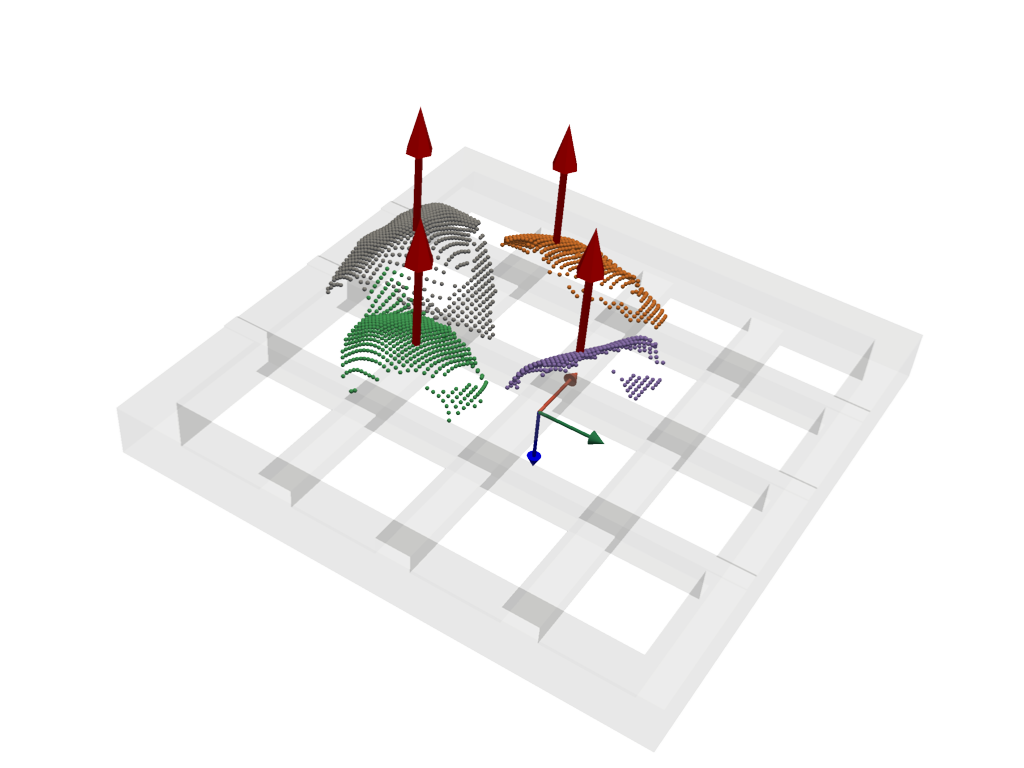}} \\ \vspace{0.1cm}
         \rotatebox{90}{\makecell{Centroid \\ (\texttt{img+pcl})}} & 
         \raisebox{-0.2\height}{\includegraphics[width=0.25\linewidth]{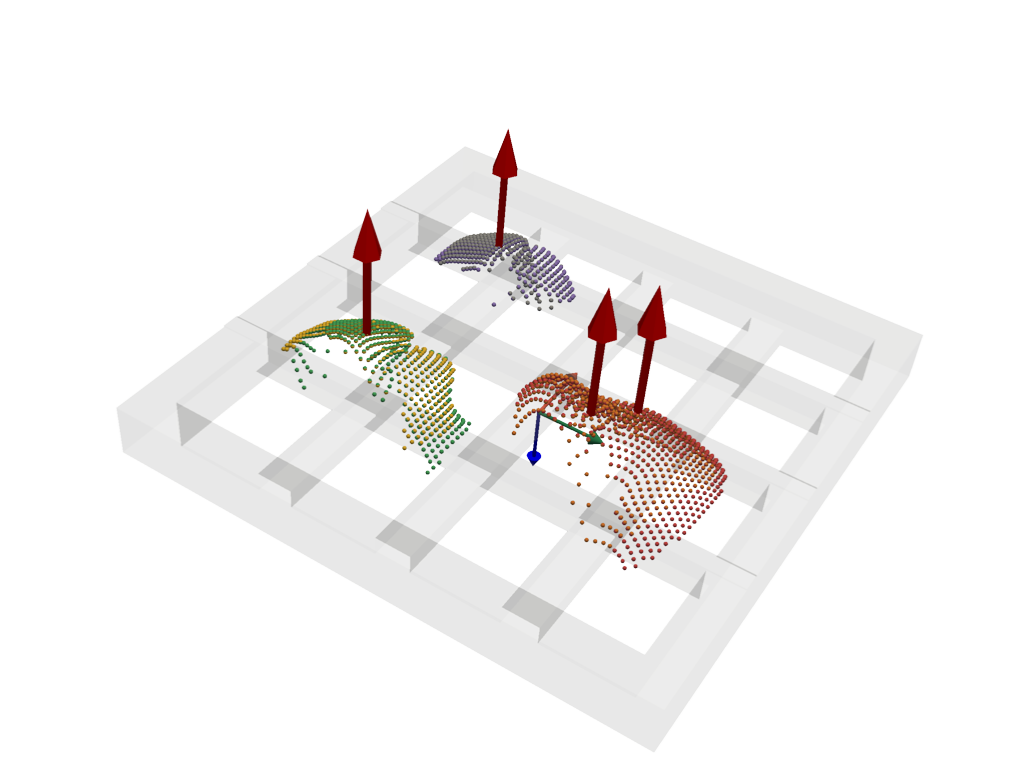}} & 
         \raisebox{-0.2\height}{\includegraphics[width=0.25\linewidth]{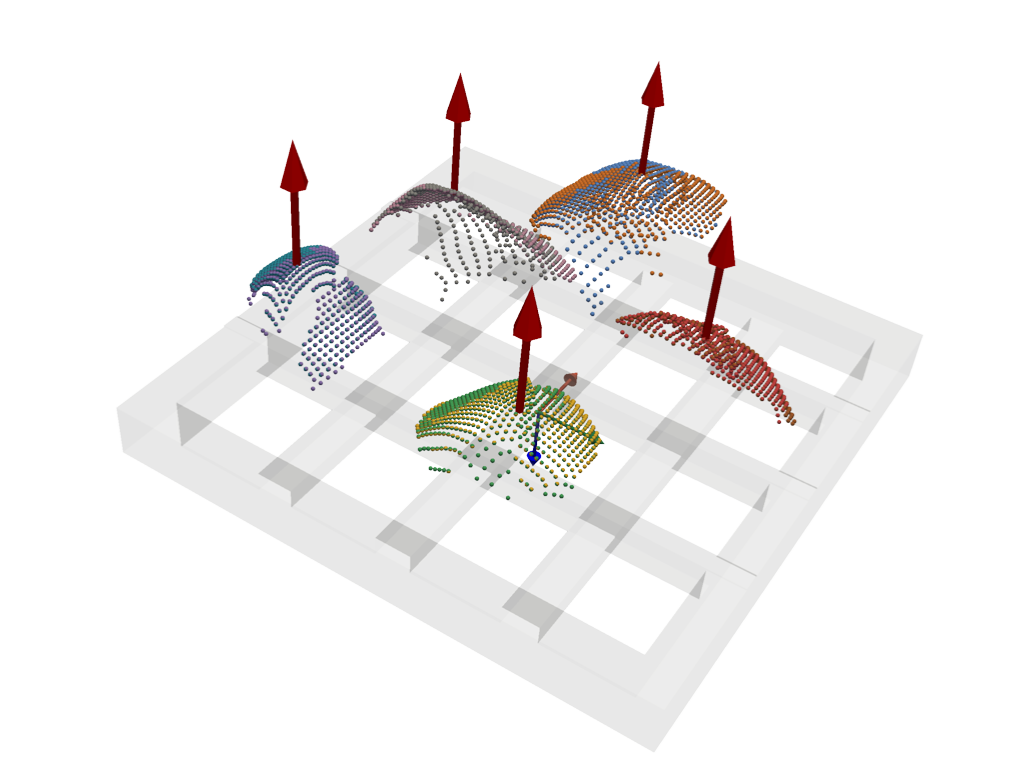}} & 
         \raisebox{-0.2\height}{\includegraphics[width=0.25\linewidth]{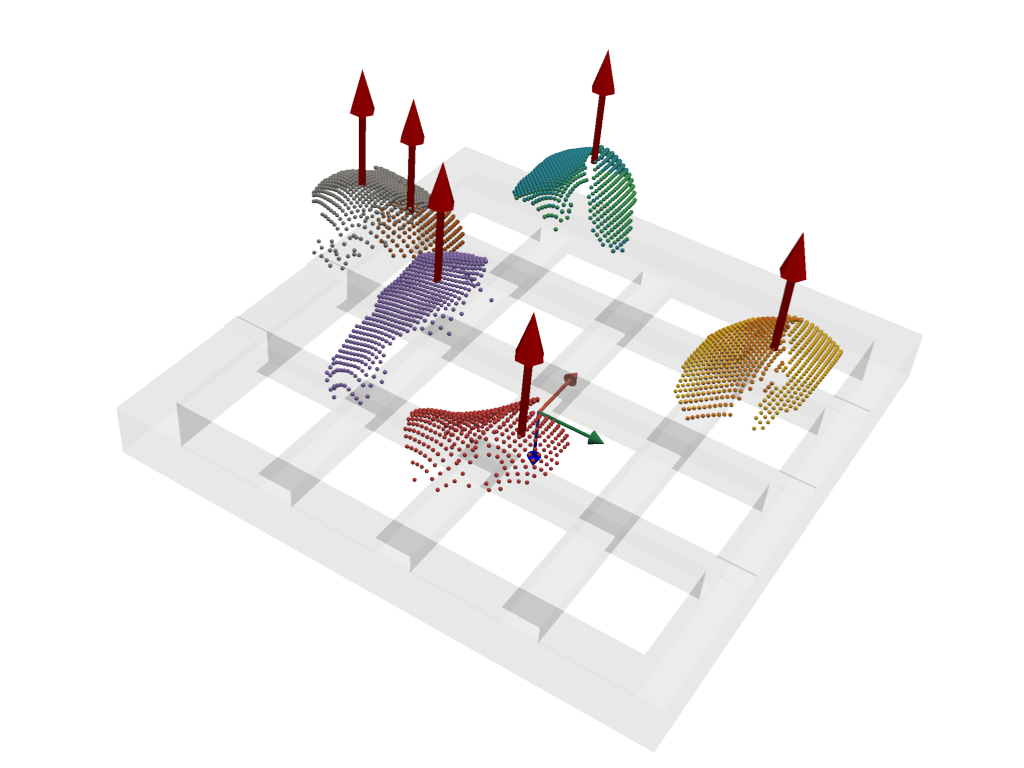}} & 
         \raisebox{-0.2\height}{\includegraphics[width=0.25\linewidth]{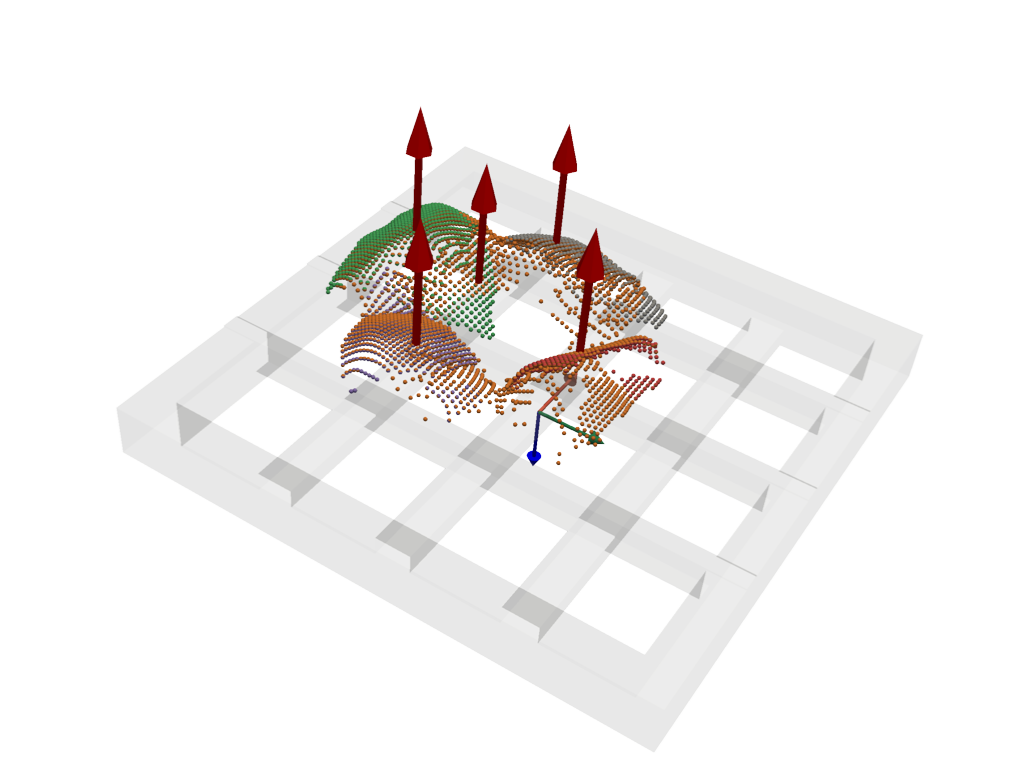}} \\ \vspace{0.1cm}
         \rotatebox{90}{Ours} & 
         \raisebox{-0.4\height}{\includegraphics[width=0.25\linewidth]{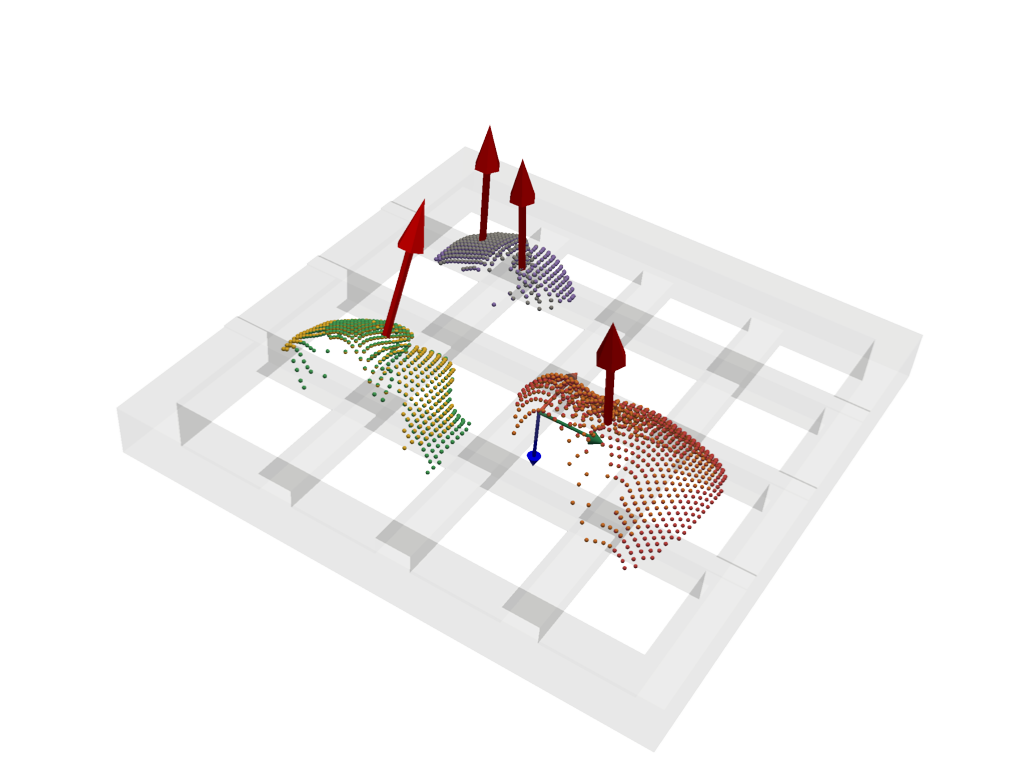}} & 
         \raisebox{-0.4\height}{\includegraphics[width=0.25\linewidth]{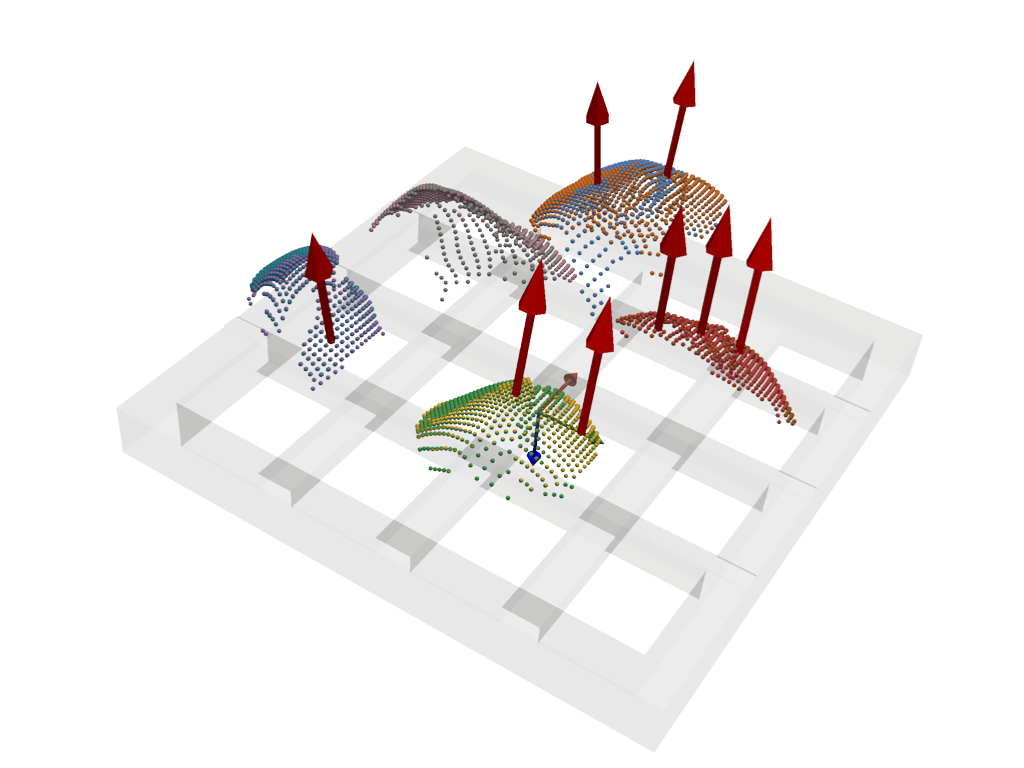}} & 
         \raisebox{-0.4\height}{\includegraphics[width=0.25\linewidth]{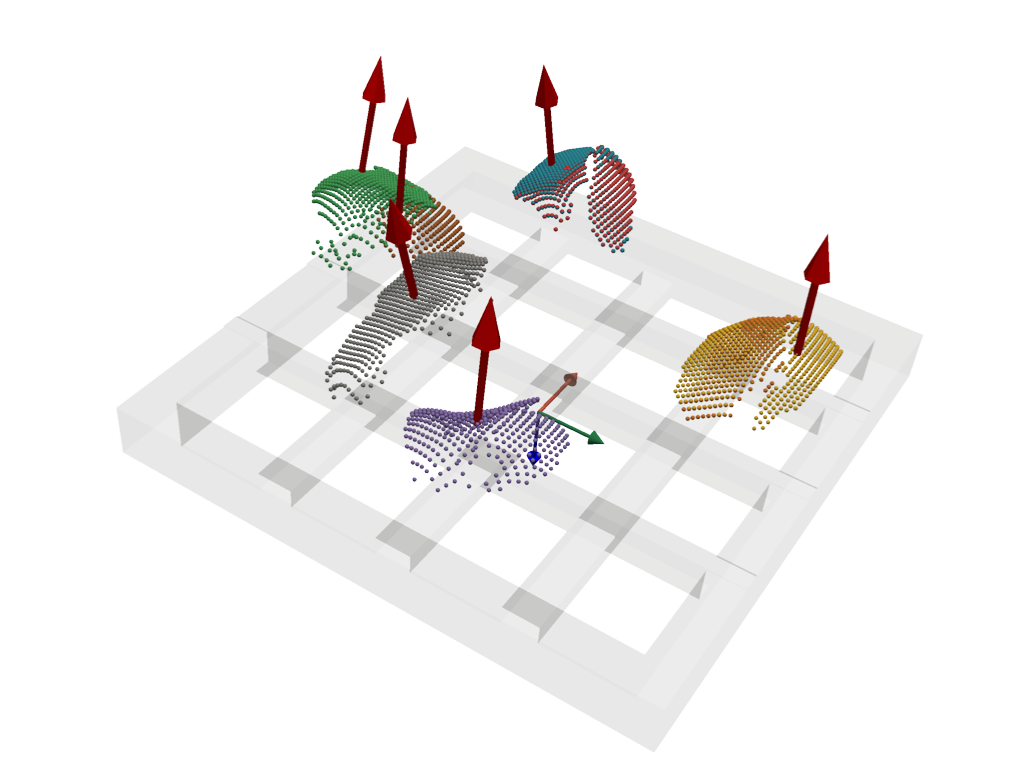}} & 
         \raisebox{-0.4\height}{\includegraphics[width=0.25\linewidth]{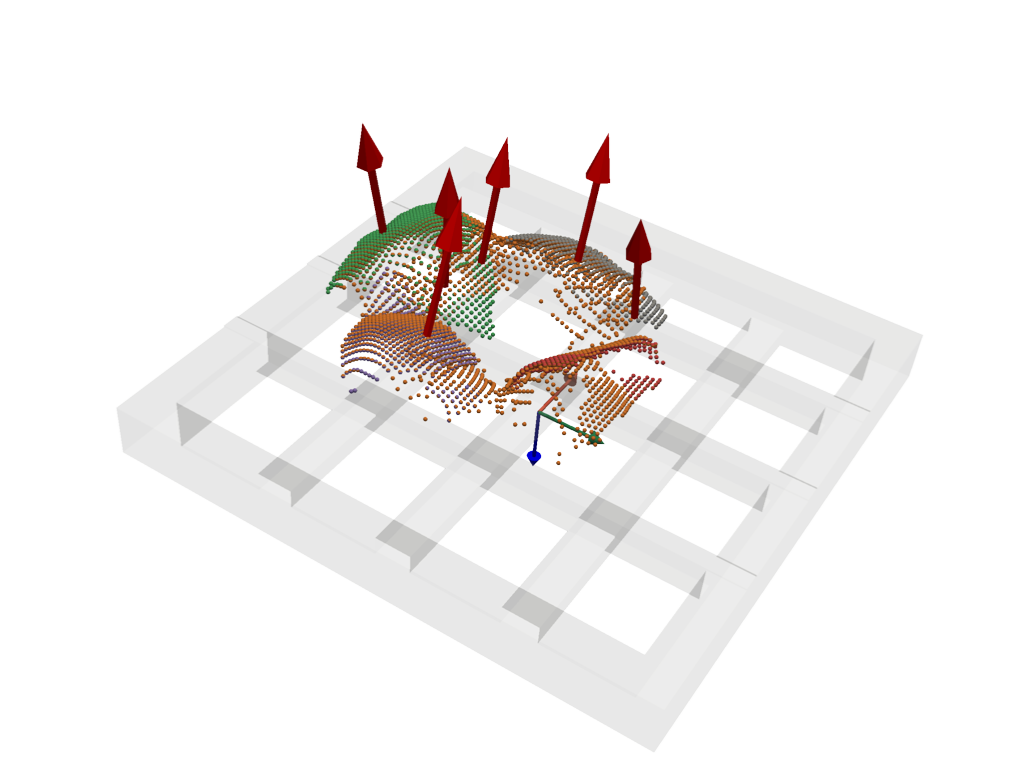}} \\ 
          & \footnotesize{(a)} &  \footnotesize{(b)}&   \footnotesize{(c)} & \footnotesize{(d)}
    \end{tabular*}
    \caption{Rock-breaking pose generation comparison. Vertically aligned images correspond to the same configuration of rocks in the grizzly. The first row of horizontally aligned images shows (RGB) regions of interest provided by the stereo cameras. The second and third rows show the rock-breaking poses generated by the Centroid method using only image-based rock segmentations (\texttt{img}) and image and point cloud-based segmentations (\texttt{img+pcl}). The fourth row shows the rock-breaking poses generated by the method proposed in this work.}
    \label{fig:qualitative_comparison_baseline}
\end{figure*}

\subsubsection{Analysis and comparison of the rock-breaking pose generation method}

We qualitatively compare our approach to generate rock-breaking poses with that proposed in~\cite{lampinen2021autonomous}, which was also tested in the real world, with a real impact hammer, within a full automation system. The implemented method can be roughly described as follows: 

\begin{enumerate}
    \item Given a segmented rock (i.e., a point cloud), its centroid is computed.
    \item A rectangular region whose center is the aforementioned centroid and whose extents align with the segmented rocks' oriented bounding box is defined.
    \item The highest point on the segmented rock's surface that is within the defined rectangular region (projected onto the rocks surface) is selected as a target pose's position.
    \item The pose's orientation is set to be orthogonal to the steel grate.
\end{enumerate}

While the original method described in~\cite{lampinen2021autonomous} only uses image-based rock segmentations, here we also include a variant that takes point cloud-based segmentations as inputs, and fuses and prioritizes the poses generated by both sources of information according to the same procedure followed by our pose generation method. We will refer to both of these baselines as ``\emph{Centroid}'', distinguishing them depending on whether they take as inputs only visual segmentations (\texttt{img}) or both visual and point cloud-based segmentations (\texttt{img+pcl}). 

Since these baselines produce target rock-breaking poses that depend on the segmented rocks' centroids, whereas our method generates candidate poses by analyzing the rock's surface geometry, a quantitative comparison following the methodology described in Sect.~\ref{subsubsec:real_world_testing} is not straightforward. The Centroid baselines are inherently designed to produce at most one rock-breaking pose per segmented rock, while our method may produce multiple or no pose at all for certain surface geometries. This is a critical issue since the evaluation metrics used in this experiment depend on the number of generated poses, as it defines the number of pressing trials. Furthermore, during real-world testing, the hammer sometimes moves rocks when pressing against them. To ensure identical rock configurations for all methods for a fair comparison, manual repositioning of rocks in between trials would be necessary. Therefore, we opt for a purely qualitative comparison.

Fig.~\ref{fig:qualitative_comparison_baseline} shows the (unprioritized) rock-breaking poses generated by these baselines and the method proposed in this work, for four different rock configurations. We note that pure image-based rock segmentation often tends to only partially characterize the surface of individual rocks. We attribute this behavior to the RTMDet segmentation model, in some cases, producing masks within the true rock instance (instead of properly delineating it). However, compared to point cloud-based segmentation, the produced segmentations here tend to better differentiate between neighboring rocks. These differences directly affect the baseline results; for the Centroid method, we can see that, by construction, only one pose is generated per rock instance in the \texttt{img} variant, whereas, for the same rock, the \texttt{img+pcl} variant sometimes produces more than one pose per rock. This happens due to the mismatch in the rock segmentation produced by the image-based method and the point cloud-based method, where different results may imply different centroids, and thus, different poses for the same ground-truth rock instance.

Despite this limitation, we argue that the \texttt{img+pcl} variant is better suited for producing target rock-breaking rocks than the \texttt{img} variant, as fusing rock-breaking poses produced by images and point clouds can partially alleviate cases in which the RTMDet model produces no segmentation masks for a given rock (which may happen, for instance, when a rock has a shape or color outside the training distribution of the segmentation model). However, the \texttt{img+pcl} variant also inherits the limitations of point cloud-based segmentation. We note that, when rocks overlap, the Centroid method using image and point cloud-based segmentations may produce poses far from the true centroid of a rock, as the resulting point-cloud based segmentation would likely fail to distinguish between rock instances. This is exemplified by the image in the third row of Fig.~\ref{fig:qualitative_comparison_baseline}-d, where the pose in the center of the cluster of rocks is generated to correspond with the centroid of a single large rock that is segmented using the point cloud-based method.

Recall that the Centroid variants generate a single rock-breaking pose per rock segmentation, regardless of its size or geometry. As a result, large rocks can be assigned a single breaking pose located near their center. From an operational perspective, this is undesirable, as human operators typically fragment large rocks by initiating the breaking process near their edges. Furthermore, these methods do not account for the orientation of the rock surface; consequently, they may generate rock-breaking poses on highly inclined surfaces, causing the hammer to slip during operation. Such slippage can result in damage to either the machine or the surrounding infrastructure, making it essential to avoid these situations.

As stated in Sect.~\ref{subsubsec:rock_segmentation_and_rock_breaking_pose_generation}, the method we propose circumvents these challenges by relying on analyzing the rocks' surfaces to produce rock-breaking poses, which inherently does not require an accurate rock instance segmentation (however, inaccurate instance segmentation does affect pose prioritization). This key characteristic induces two main qualitative differences when compared with the Centroid method: (i) a single rock segmentation may produce several poses or none at all, and (ii) rock-breaking poses are oriented considering the surface normal of the rock's geometry and the kinematics of the impact hammer. We can see for instance, that images in the fourth row of Fig.~\ref{fig:qualitative_comparison_baseline}-b and Fig.~\ref{fig:qualitative_comparison_baseline}-d show rocks for which no poses are generated due to the rock's surface normal orientation being too extreme, and, particularly in the image shown in Fig.~\ref{fig:qualitative_comparison_baseline}-b, one of the rocks, due to its size and regular shape, produces three adjacent poses. At first glance, the absence of breaking poses on some rocks may appear to be a limitation of the proposed method. However, this behavior is intentional and reflects operational safety constraints. In practice, when a rock presents highly inclined surfaces, operators typically reposition it using the tip of the hammer before attempting to break it. A simpler centroid-based approach could still generate a target in these situations, but such a pose may not be operationally feasible and could increase the risk of hammer slippage.

\subsection{Profiling}
\label{subsubsec:profiling}

A key feature of the proposed system is its ability to operate in real time. To profile the system, we assessed all the pipeline's modules (i.e., its nodes) in terms of their output frequency and latency. Detailed results on these metrics are presented in App.~\ref{app:software_architecture}, which also shows the complete software architecture graphically. For each node, these metrics were estimated measuring processing times over a sliding window of sizes in $\{20, 200\}$. The end-to-end latency was (optimistically) estimated as the cumulative processing time along the critical paths of the pipeline, i.e., the longest processing branches required to generate each output of interest (target rock-breaking poses, and a point cloud representation of the environment that can be consumed by other downstream tasks). It should be noted that these estimates represent a lower bound on the actual system latency, as they do not account for communication overhead between nodes (moreover, when nodes are distributed across multiple machines, additional network and synchronization delays may further increase the end-to-end latency).

Table~\ref{tab:profiling_gnral} summarizes the profiling results for each pipeline output. As can be observed, all outputs are generated at approximately $10$~Hz regardless of the selected window size. Regarding latency, the results show that it takes approximately $670$~ms for a sensor measurement entering the pipeline to produce a corresponding set of rock-breaking poses. Consequently, the generated poses correspond to sensor data acquired roughly $670$~ms earlier. Analogously, in the case of the processed point cloud representing the impact hammer environment, the end-to-end latency is of approximately $380$~ms.

\begin{table}
    \caption{System profiling.}
    \label{tab:profiling_gnral}
    \begin{tabular*}{\linewidth}{@{\extracolsep{\fill}}c@{\extracolsep{\fill}}l@{\extracolsep{\fill}}S[table-format=2.3]@{\extracolsep{\fill}}S[table-format=3.2]@{\extracolsep{\fill}}}
    \toprule
    Window size & Output & {Frequency [Hz]} & {Latency [ms]}\\
    \midrule
    \multirow{2}{*}{\makecell[l]{20}} 
  
    & Rock-breaking poses & 9.875   & 672.83\\
    & Env. point cloud         & 10.068  & 379.83\\
    \midrule
    \multirow{2}{*}{\makecell[l]{200}} 
    
    & Rock-breaking poses & 10.153 & 675.64\\
    & Env. point cloud         & 10.169 & 384.43\\
    \bottomrule
    \end{tabular*}
\end{table}

\section{Conclusion}
\label{sec:conclusion}

In this work, we presented a perception pipeline as a step towards the automation of hydraulic impact hammers commonly used in mining operations. This pipeline allows generating and prioritizing rock-breaking poses, and a 3D scene representation of the environment for other downstream tasks (e.g., collision-aware control subsystems). Moreover, the proposed system runs in real time (on embedded hardware), and takes into consideration key requirements to make its integration to a full automation stack possible. In particular, the perception pipeline was validated through real-world performance evaluations using a Bobcat~E10 mini-excavator equipped with an impact hammer as end-effector, operating over a steel grate where rocks were placed in multiple different configurations.

The results demonstrate that the proposed pipeline operates at approximately $10$~Hz, providing a high-frequency stream of target poses suitable for responsive closed-loop control. In addition, the system provides a robust 3D scene representation by combining ad-hoc self-filtering and occlusion management methods. Furthermore, real-world validation showed that the proposed surface-normal-based pose generation strategy achieved a $72$\% success rate across impact hammer pressing trials, suggesting its practical applicability for autonomous impact hammer operation.

Despite these promising results, some limitations remain. The current generation of rock-breaking poses relies on the segmented rocks' surface analysis and does not explicitly account for the geometry of the support structure. Future work will focus on incorporating a detailed model of the steel grate into the breaking pose computation. This enhancement is expected to improve the identification of stable regions where rocks are less likely to shift during impact. In addition, the system sometimes does not generate rock-breaking poses for rocks with highly inclined surfaces. This is the result of a design decision, and could be alleviated by adaptively relaxing the constrains of the pose generation method, or by using a more naive approach for these specific rocks. As proposed in~\citet{lampinen2021robust}, another path to address this issue could be conceiving a controller that can reposition rocks (via non-prehensile manipulation, given the capabilities of impact hammers); we leave this research direction for future work. Finally, updating the full software stack will be explored to further optimize the processing pipeline, leveraging newer architectural features to reduce latency.

\section*{Acknowledgments}

The authors acknowledge Sandra Romero for providing an initial implementation of the RTMDet segmentation model wrapped in a ROS node.

\appendices

\section{Sensor data acquisition}
\label{app:sensor_data_acquisition}

For our experimental setup, two ZED X stereo cameras were used. Camera initialization and data acquisition are performed using the SDK provided by Stereolabs.\footnote{\url{https://github.com/stereolabs/zed-sdk}} The camera configuration parameters are listed in Table~\ref{tab:zedx_parameters}.

Note that rather than streaming point clouds directly, each camera published RGB-D data and their corresponding intrinsic parameters. Point clouds were subsequently generated using the procedure described in Sect.~\ref{subsubsubsec:pointcloud_generation_and_fusion}. When generating point clouds, a stride of $2$ pixels is used, i.e., a pixel in the $x$- and the $y$-axis of the image coordinates is skipped when computing projections. This reduces computational cost further down the perception pipeline, and, given our setup, preserves sufficient point density for critical tasks, such as DBSCAN clustering and surface normal estimation. 

\begin{table}
    \centering
    \caption{ZED X stereo camera parameters.}
    \label{tab:zedx_parameters}
    \begin{tabular*}{\linewidth}{l@{\extracolsep{\fill}}l}
    \toprule
    \textbf{Parameter} &  \textbf{Value} \\
    \midrule
    Camera resolution  & SVGA ($960\times600$ px.) \\
    Depth mode &  \texttt{Neural}  \\
    Depth stabilization & $40$ \\
    \bottomrule
    \end{tabular*}
\end{table}

\section{Bilateral filter} 
\label{app:bilateral_filter}

The bilateral filter is an edge-preserving smoothing technique widely used in image processing applications, particularly for image denoising. It operates by replacing the value of each pixel with a weighted average of the values within its neighborhood, where the weights depend on both spatial proximity and intensity similarity. This property enables the filter to reduce noise while preserving significant image structures and edges~\citep{tomasi1998bilateral}.

Due to specific constraints in the deployment platform (see App.~\ref{app:software_architecture}), two separate implementations of the bilateral filter were required (since the filter is used by nodes that rely on both Open3D and JAX). In the case of Open3D, the library provides a built-in implementation of the filter. For JAX, however, a custom implementation was necessary. To this end, a vectorized implementation of the bilateral filter found online\footnote{\url{https://github.com/avivelka/Bilateral-Filter}} was adapted and modified to use JAX. Both the Open3D and JAX-based bilateral filters were configured to run using the same set of parameters; a kernel size of $7$, and an std. for image pixel positions of $5$. Given these parameters and the ``smoothness'' of our scene, the std. for image content was less relevant and set to $20$, effectively making the filter behave approximately like Gaussian blur. We note, however, that proper tuning of this parameter can indeed be useful for more complex and larger-scale environments.

\section{Software architecture}
\label{app:software_architecture}

The perception pipeline runs on an Jetson AGX Orin. A critical constraint of this platform is that, given our software stack (running JetPack~5.1.2), it lacks some interoperability features between JAX, Open3D, and PyTorch. This limitation had to be considered during the design of both the individual nodes and the overall processing pipeline in order to avoid computational overhead associated with transferring data structures between these libraries.

\begin{figure*}
        \centering
        \includegraphics[width=\linewidth]{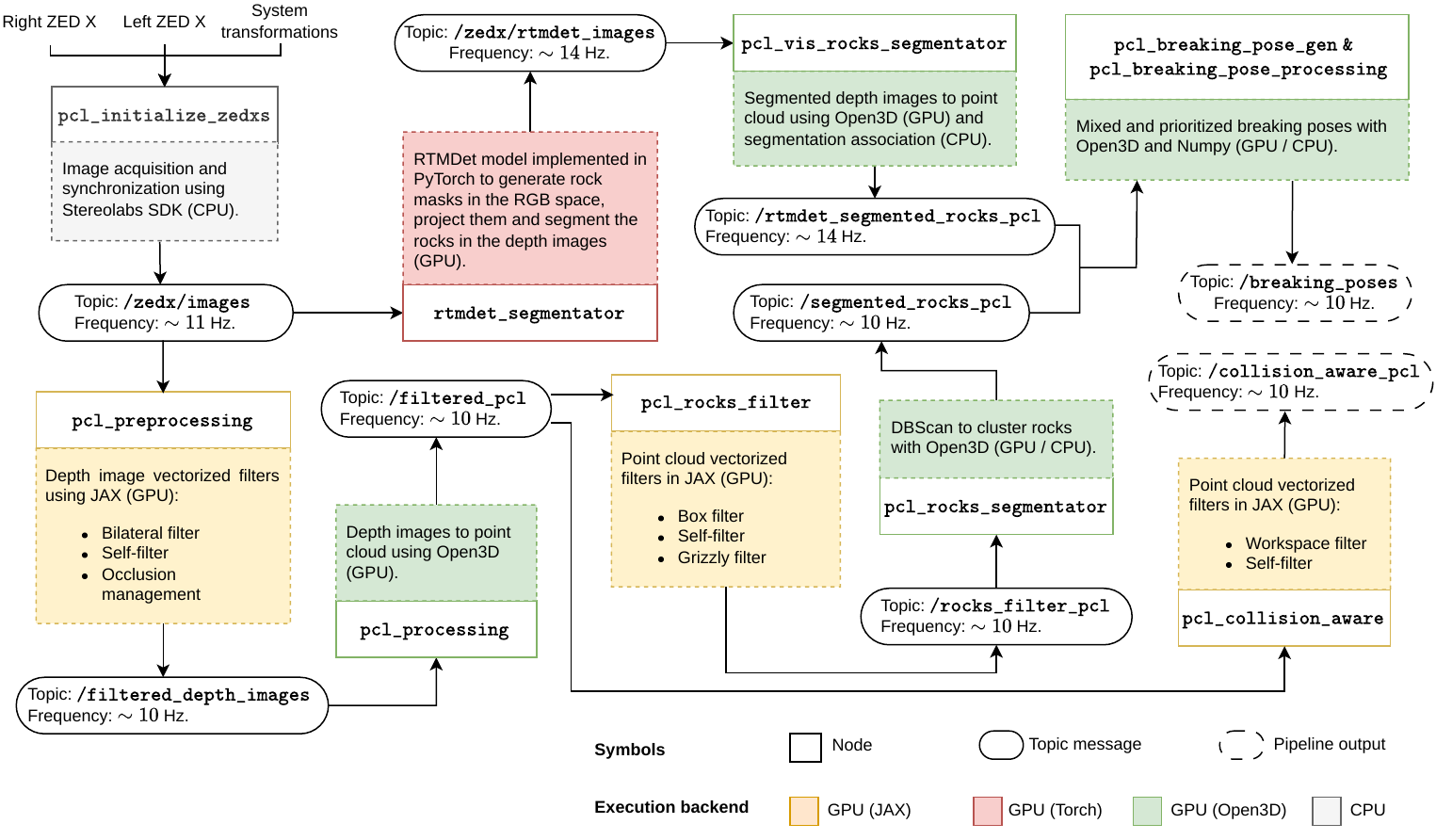}
        \caption{Software architecture diagram. Similarly to Fig.~\ref{fig:streamlined_perception_pipeline}, different colors highlight the primary execution backend for each node.}
    \label{fig:software_architecture}
\end{figure*}

A detailed view of the software architecture can be found in Fig.~\ref{fig:software_architecture}, which specifies for each node their inputs, outputs, functionality and operating frequency. Table~\ref{tab:profiling} shows a detailed profiling of every node.

\begin{table*}
    \centering
    \caption{Detailed system profiling.}
    \label{tab:profiling}
    \begin{tabular*}{\linewidth}{cl@{\extracolsep{\fill}}S[table-format=2.3]@{\extracolsep{\fill}}S[table-format=2.3]@{\extracolsep{\fill}}S[table-format=2.3]@{\extracolsep{\fill}}S[table-format=2.3]@{\extracolsep{\fill}}S[table-format=2.3]}
    \toprule
    Window size & Node & {Frequency [Hz]} & {Stdv. [s]} & {Min. [s]} & {Max. [s]} & {Latency [ms]}\\
    \midrule
    \multirow{12}{*}{\makecell[l]{20}} 
    & \texttt{pcl\_initialize\_zedx}            & 12.087 & 0.011 & 0.064 & 0.104 & 82.73\\
    & \texttt{pcl\_preprocessing}               & 10.150 & 0.010 & 0.083 & 0.118 & 98.52\\
    & \texttt{pcl\_processing}                  & 10.084 & 0.015 & 0.076 & 0.134 & 99.16\\
    & \texttt{pcl\_collision\_aware}            & 10.068 & 0.018 & 0.064 & 0.129 & 99.32\\
    & \texttt{pcl\_rocks\_filter}               & 10.126 & 0.020 & 0.067 & 0.136 & 98.75\\
    & \texttt{pcl\_rocks\_segmentator}          & 10.083 & 0.029 & 0.066 & 0.176 & 99.17\\
    & \texttt{pcl\_breaking\_pose\_gen}        & 9.875  & 0.032 & 0.062 & 0.194 & 101.26\\
    & \texttt{pcl\_breaking\_pose\_processing} & 10.724 & 0.029 & 0.001 & 0.112 & 93.24\\
    & \texttt{rtmdet\_segmentator}$^\dagger$              & 15.154 & 0.021 & 0.030 & 0.093 & 65.98\\
    & \texttt{pcl\_vis\_rocks\_segmentator}     & 14.909 & 0.018 & 0.040 & 0.102 & 67.07\\
    \cmidrule{2-7}
    & Rock-breaking poses output & 9.875  & {--} & {--} & {--} & 672.83\\
    & Point cloud output         & 10.068 & {--} & {--} & {--} & 379.83\\
    \midrule
    \multirow{12}{*}{\makecell[l]{200}} 
    & \texttt{pcl\_initialize\_zedx}            & 11.166 & 0.013 & 0.034 & 0.146 & 89.55\\
    & \texttt{pcl\_preprocessing}               & 10.176 & 0.009 & 0.070 & 0.131 & 98.27\\
    & \texttt{pcl\_processing}                  & 10.175 & 0.014 & 0.062 & 0.154 & 98.28\\
    & \texttt{pcl\_collision\_aware}            & 10.169 & 0.014 & 0.071 & 0.162 & 98.33\\
    & \texttt{pcl\_rocks\_filter}               & 10.171 & 0.015 & 0.066 & 0.161 & 98.31\\
    & \texttt{pcl\_rocks\_segmentator}          & 10.153 & 0.026 & 0.015 & 0.195 & 98.49\\
    & \texttt{pcl\_breaking\_pose\_gen}        & 10.160 & 0.023 & 0.045 & 0.180 & 98.42\\
    & \texttt{pcl\_breaking\_pose\_processing} & 10.602 & 0.030 & 0.000 & 0.130 & 94.32\\
    & \texttt{rtmdet\_segmentator}$^\dagger$              & 14.379 & 0.022 & 0.014 & 0.121 & 69.54\\
    & \texttt{pcl\_vis\_rocks\_segmentator}     & 14.398 & 0.016 & 0.031 & 0.109 & 69.45\\
    \cmidrule{2-7}
    & Rock-breaking poses output & 10.153 & {--} & {--} & {--} & 675.64\\
    & Point cloud output         & 10.169 & {--} & {--} & {--} & 384.43\\
    \bottomrule
    \end{tabular*}
    \vspace{2pt}

    \footnotesize{\raggedright $\dagger$ This node maintains an artificially higher frequency by repeating some of its outputs (RTMDet model inferences). This also explains the frequency observed for the \texttt{pcl\_vis\_rocks\_segmentator} node, which takes the outputs of \texttt{rtmdet\_segmentator} as inputs (see App.~\ref{app:rtmdet_settings} for further details).\hfill}
\end{table*}

\section{RTMDet settings}
\label{app:rtmdet_settings}

The RTMDet model used for rock segmentation was adapted from the accompanying code for~\citet{lyu2022rtmdet}, and wrapped in a ROS node (we used the RTMDet-tiny model). The procedure for the generation of regions of interest over the grizzly was implemented to achieve a similar result to what is reported in~\citet{ruiz2025system}. To improve inference frequency, a quasi-static environment was assumed; under this assumption, the most recent inference result was reused until an updated mask became available from the segmentation model. This strategy enabled the system module to operate (artificially) at a frequency of approximately $15$~Hz.

In addition, the RTMDet model was fine-tuned on a custom dataset comprising images acquired from the experimental setup both with and without rocks placed on the grizzly. This dataset extends the one utilized in~\citet{ruiz2025system} by incorporating images captured with the ZED~X cameras employed in the current experimental setup, thereby ensuring the availability of representative samples for our sensor configuration. Details of the dataset splits, including the number of images and the proportions assigned to the training, validation, and test subsets for each camera type, are provided in Table~\ref{tab:rtmdet_dataset}. The training hyperparameters used for the RTMDet model are listed in Table~\ref{tab:rtmdet_parameters}. Since the model requires a fixed input resolution, all images were resized and padded while preserving their original aspect ratio.

\begin{table}
    \centering
    \caption{RTMDet dataset characterization.}
    \label{tab:rtmdet_dataset}
    \begin{tabular*}{\linewidth}{l@{\extracolsep{\fill}}lll}
    \toprule
    Camera model & Train & Validation & Test \\
    \midrule
    ZED X            & 270     & 134     & 138  \\ 
    MOXA P06HC       & 364     & 32      & 32   \\ \midrule \midrule
    Total images     & 634     & 166     & 170  \\ 
    \bottomrule
    \end{tabular*}
\end{table}

\begin{table}
    \centering
    \caption{RTMDet parameters.}
    \label{tab:rtmdet_parameters}
    \begin{tabular*}{\linewidth}{l@{\extracolsep{\fill}}l}
    \toprule
    Parameter &  Value \\
    \midrule
    Input shape   & 480$\times$480 \\
    Epochs        & 50 \\
    Optimizer     & Adam \\
    Learning rate & 0.004 \\
    Weight decay  & 0.05 \\
    Batch size    & 8 \\
    Padding       & \texttt{True} \\
    Confidence threshold & 0.75 \\
    \bottomrule
    \end{tabular*}
\end{table}

\section{Point cloud filters characterization}
\label{app:point_cloud_filters}

To characterize the point cloud filters implemented as part of the proposed perception pipeline, the number of points removed by each of them alongside their execution time was measured. These metrics allow assessing the extent to which these filters enable a real-time, dynamic representation of the $3$D scene (e.g., for exteroceptive-aware control tasks), while effectively reducing point cloud size, and thus, processing time further down the pipeline (e.g., for the generation of rock-breaking poses).

To do so, the impact hammer was teleoperated while the point cloud filters were applied in real-time. During operation, filtered point cloud data were recorded over $200$ filtering steps. Across $30$ different teleoperation trials, this resulted in approximately $10$ minutes of data, providing a sufficiently broad sample for subsequent analysis.

Two distinct filter stacks were implemented at different nodes within the processing pipeline:

\begin{itemize}
    \item A box filter, self-filter, and grizzly filter applied in the node responsible for retaining only those points corresponding to potential rocks on the grizzly.
    \item A workspace filter and a self-filter applied in the node responsible for generating the dynamic representation of the 3D scene.
\end{itemize} 

Both filter stacks process the same input point cloud, which has approximately $(107.3 \pm 2.3)\times 10^3$ points for the performed experiments. The results regarding the number of filtered points and execution times are summarized in Table~\ref{tab:filter_stacks_characterization}. Note that each individual filter is represented by a symbol ($b$, $s$, $g$, or $w$), and filter stacks are denoted by their composition.

For the percentage of filtered points, both absolute and relative metrics are reported. The absolute percentage is computed with respect to the original input point cloud, whereas the relative percentage is computed with respect to the output of the preceding filtering stage. Consequently, the absolute metric quantifies the cumulative reduction achieved by a filter, while the relative metric reflects its local effect within the filter stack.

\begin{table}
    \sisetup{separate-uncertainty=true, table-align-uncertainty=true, retain-zero-uncertainty=true}
    \centering
    \caption{Point cloud filters characterization.}
    \label{tab:filter_stacks_characterization}
    \begin{tabular*}{\linewidth}{@{\extracolsep{\fill}}c@{\extracolsep{\fill}}l@{\extracolsep{\fill}}S[table-format=2.2(3)]@{\extracolsep{\fill}}S[table-format=2.2(3)]@{\extracolsep{\fill}}S[table-format=2.2(3)]@{\extracolsep{\fill}}}
    \toprule
       & \multirow{2}{*}{Filter}  & \multicolumn{2}{c}{\makecell{Filtered points}} & {\multirow{2}{*}{Time [ms]}}\\ \cmidrule{3-4}
       &  & {Absolute [\%]} & {Relative [\%]} & \\ \midrule
       \multirow{4}{*}{\rotatebox{90}{\makecell{Rocks \\point cloud}}}
       & Box ($b$)&  87.53 \pm 1.26 & 87.53 \pm 1.26 & 4.21 \pm 1.71 \\
       & 3D Self-filter ($s$)& 0.04  \pm 0.09 & 0.30  \pm 0.78  & 6.34  \pm 4.53\\
       & Grizzly ($g$)& 4.49  \pm 2.09 & 35.12 \pm 16.30 & 34.67 \pm 8.74 \\ \cmidrule{2-5}
       & $g\circ s \circ b$    & 92.05 \pm 1.40  & {--} &  45.23 \pm 8.30 \\  \midrule \midrule
       \multirow{3}{*}{\rotatebox{90}{3D scene}}
       & Workspace ($w$) & 58.16 \pm 1.44 & 58.16 \pm 1.44 & 4.35  \pm 1.58 \\
       & 3D Self-filter ($s$) & 4.87  \pm 0.81 & 11.62 \pm 1.76 & 40.93 \pm 4.53 \\ \cmidrule{2-5}
       & $s \circ w$ &  63.4 \pm 1.12  & {--}& 45.28 \pm 4.76 \\ 
       \bottomrule
    \end{tabular*}
\end{table}

\section{Rock-breaking poses generation}
\label{app:breaking_poses}

An impact hammer is subject to several physical and operational constraints that must be considered to ensure safe operation. One of the most important constraints is that the hammer should strike the rocks as perpendicular as possible to the surface of the grizzly. Since satisfying this condition exactly is not always feasible when targeting a rock, an acceptable pitch range is defined. To enforce this constraint, as described in Sect.~\ref{subsubsec:breaking_poses_generator}, two angular thresholds are defined. Their values, along with the minimum cluster size required for a candidate pose to be accepted, are reported in Table~\ref{tab:pcl_breaking_poses_gen}.

In addition, and as described in Sect.~\ref{subsubsec:breaking_poses_processor}, the order in which the hammer visits the generated rock-breaking poses depends on a score-based prioritization. For our experimental setup, the values of the scalar weights used to prioritize the target poses are shown in Table~\ref{tab:pcl_breaking_poses_gen}.

\begin{table}
     \centering
     \caption{Rock-breaking poses generation parameters.}
     \label{tab:pcl_breaking_poses_gen}
     \begin{tabular*}{\linewidth}{l@{\extracolsep{\fill}}l@{\extracolsep{\fill}}l@{\extracolsep{\fill}}}
     \toprule
     Module & Parameter & Value \\ \midrule

     \multirow{3}{*}{Candidates poses filtering}
     & Min. $|\mathcal{C}_\text{rock}|$ ($P_\text{min}$) & 10\\
     & Soft deviation angle ($\theta_\text{soft}$) [rad]  & 0.5618\\
     & Hard deviation angle ($\theta_\text{hard}$) [rad] & 0.3618\\ \midrule
     \multirow{4}{*}{\makecell{Rock-breaking poses \\prioritization}}
     & Distance weight, $\lambda_1$  & 1\\
     & Temporal weight, $\lambda_2$  & 2\\
     & Rock volume weight, $\lambda_3$  & 3\\
     & Last visited pose distance weight, $\lambda_4$  & 5\\
    \bottomrule

     \end{tabular*}
 \end{table}

\balance
\bibliographystyle{apalike}
\bibliography{references}

\end{document}